\documentclass[10pt,twocolumn,letterpaper]{article}

\usepackage{iccv}
\usepackage{subfig}
\usepackage{times}
\usepackage{epsfig}
\usepackage{graphicx}
\usepackage{amsmath}
\usepackage{amssymb}

\usepackage{url}
\usepackage{booktabs}       
\usepackage{amsfonts}       
\usepackage{nicefrac}       
\usepackage{microtype}      
\usepackage{xcolor}         
\usepackage{enumitem}
\usepackage{bm}
\usepackage{array}
\usepackage{graphbox} 
\usepackage{multirow}

\usepackage{algorithm,algpseudocode}

\newcommand{\pt}{prototypicality}

\makeatletter
\newcommand*{\rom}[1]{\expandafter\@slowromancap\romannumeral #1@}

\DeclareMathOperator{\arcosh}{arcosh}

\newcommand{\tb}[3]{\setlength{\tabcolsep}{#2mm}\begin{tabular}{#1}#3\end{tabular}}

\usepackage[pagebackref=true,breaklinks=true,colorlinks,bookmarks=false]{hyperref}

\iccvfinalcopy 

\def\iccvPaperID{12612}

\ificcvfinal\fi

\begin{document}

\title{
Unsupervised Feature Learning with Emergent Data-Driven Prototypicality 
}

\author{
\tb{@{}cccc@{}}{5}{
Yunhui Guo$^{1}$ & 
Youren Zhang$^{2}$ & 
Yubei Chen$^{3}$ &
Stella X. Yu$^{2}$\\
}\\
\tb{ccc}{5}{
$^{1}$The University of Texas at Dallas &  
$^{2}$University of Michigan & 
$^{3}$New York University
}}

\maketitle

\begin{abstract}
Given an image set without any labels, our goal is to train a model that maps each image to a point in a feature space such that, not only proximity indicates visual similarity, but where it is located directly encodes how prototypical the image is according to the dataset.
%

Our key insight is to perform unsupervised feature learning in hyperbolic instead of Euclidean space, where the distance between points still reflect image similarity, and yet we gain additional capacity for representing prototypicality with the location of the point: The closer it is to the origin, the more prototypical it is.  The latter property is simply emergent from optimizing the usual metric learning objective:  The image similar to many training instances is best placed at the center of corresponding points in Euclidean space, but closer to the origin in hyperbolic space.

We propose an unsupervised feature learning algorithm in \underline{H}yperbolic space with sphere p\underline{ACK}ing. HACK first generates uniformly packed particles in the Poincar\'e ball of hyperbolic space and then assigns each image uniquely to each particle.   Images after congealing are regarded more typical of the dataset it belongs to.  With our feature mapper simply trained to spread out training instances in hyperbolic space, we observe that images move closer to the origin with congealing, validating our idea of unsupervised prototypicality discovery.  We demonstrate that our data-driven prototypicality provides an easy and superior unsupervised instance selection to reduce sample complexity, increase model generalization with atypical instances and robustness with typical ones.
\end{abstract}

\begin{figure}[t]
\begin{center}
\includegraphics[width=\linewidth]{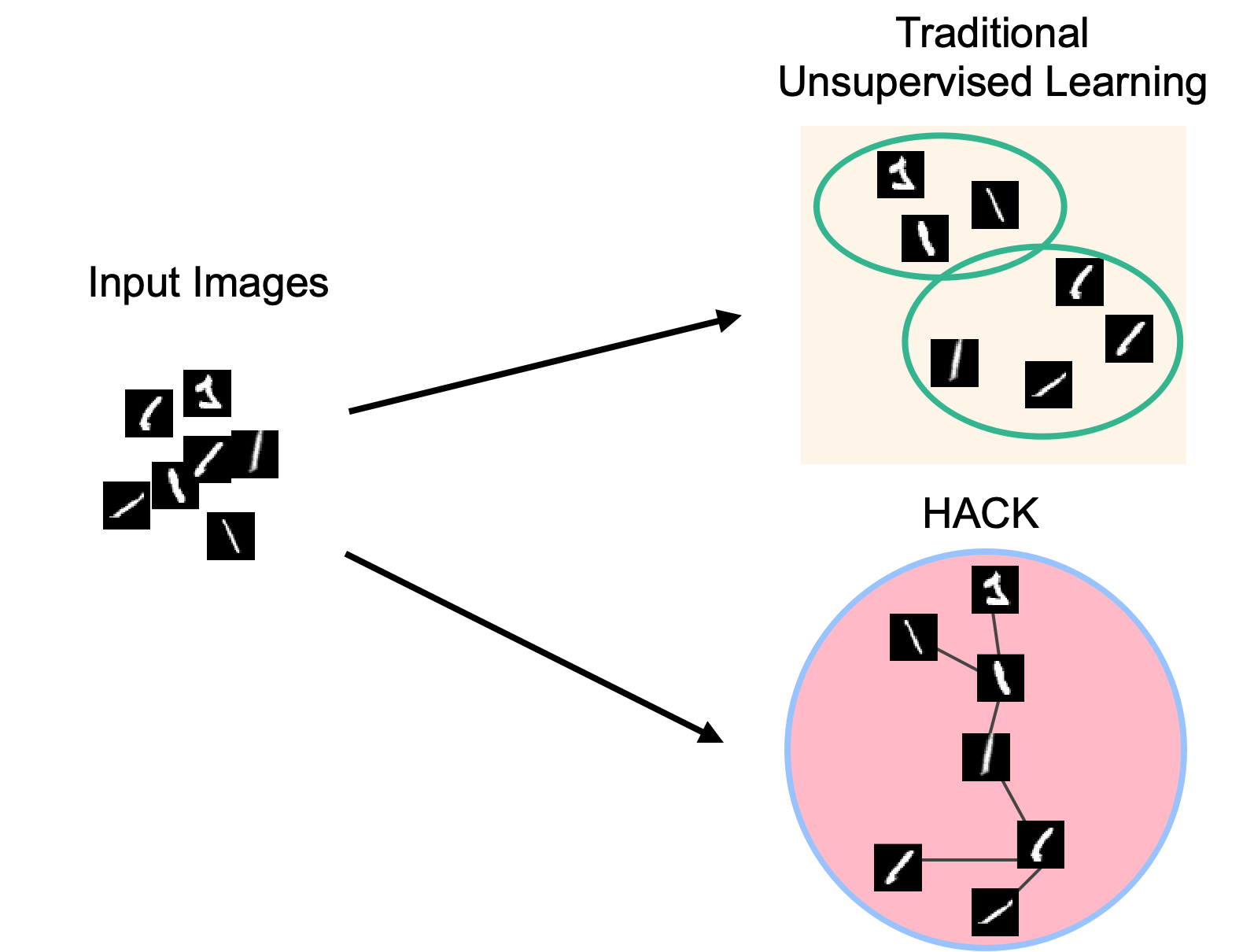}
\end{center}
   \caption{\textbf{Different from the existing unsupervised learning methods which aim to group examples via semantic similarity, HACK organizes images in hyperbolic space in a hierarchical manner.} The typical images are at the center of the Poincar\'e ball and the atypical images are close to the boundary of the Poincar\'e ball. }
\label{fig:teaser}
\end{figure}

\section{Introduction}

Not all instances are created equal. Some instances are more representative of the class and some instances are outliers or anomalies. Representative examples can be viewed as prototypes and used for interpretable machine learning \cite{bien2011prototype}, curriculum learning \cite{bengio2009curriculum}, and learning better decision boundaries \cite{carlini2018prototypical}. With prototypical examples, we can also conduct classification with few or even one example \cite{miller2000learning}. Given an image dataset, thus it is desirable to organize the examples based on \pt.

If the features of the images are given, it is relatively easy to find the prototypes by examining the density peaks of the feature distribution. If the features are not given, discovering prototypical examples without supervision is difficult: There is no universal definition or simple metric to assess the \pt\ of the examples. A naive method to address this problem is to examine the gradient magnitude \cite{carlini2018prototypical}. However, this approach is shown to have a high variance which is resulted from different training setups \cite{carlini2018prototypical}. Some methods address this problem from the perspective of adversarial robustness \cite{stock2018convnets,carlini2018prototypical}: prototypical examples should be more adversarially robust. However, the selection of the prototypical examples highly depends on the adversarial method and the metric used in the adversarial attack. Several other methods exist for this problem but they are either based on heuristics or lack a proper justification \cite{carlini2018prototypical}. 

Naturally, given a feature space, prototypical examples can be identified as density peaks. However, prototypicality undergoes changes as the feature space undergoes changes. In this paper, we propose an unsupervised feature learning algorithm, called HACK, for learning features that reflect prototypicality.
Different from existing unsupervised learning methods, HACK naturally leverages the geometry of \emph{hyperbolic space} for unsupervised learning. Hyperbolic space is non-Euclidean space with constant non-negative curvature \cite{anderson2006hyperbolic}. Different from Euclidean space, hyperbolic space can represent hierarchical relations with low distortion. Poincar\'e ball model is one of the most commonly used models for hyperbolic space \cite{nickel2017poincare}. One notable property of Poincar\'e ball model is that the distance to the origin grows exponentially as we move towards the boundary. Thus, the points located in the center of the ball are close to all the other points while the points located close to the boundary are infinitely far away from other points. With unsupervised learning in hyperbolic space, HACK can learn features which capture both visual similarity and \pt\ (Figure \ref{fig:teaser}).

HACK optimizes the organization of the dataset by assigning the images to a set of uniformly distributed particles in hyperbolic space. The assignment is done by minimizing the total hyperbolic distance between the features and the particles via the Hungarian algorithm. The \pt\ arises naturally based on the distance of the example to the others. Prototypical examples tend to locate in the center of the Poincar\'e ball and atypical examples tend to locate close to the boundary. Hyperbolic space readily facilitates such an organization due to the property of the hyperbolic distance.

Our paper makes the following contributions.
\begin{itemize}[leftmargin=*,topsep=1pt,itemsep=-4pt]

    \item We propose the first unsupervised feature learning method to learn features which capture both visual similarity and \pt. The positions of the features reflect \pt\ of the examples.
    \item The proposed method HACK assigns images to particles that are uniformly packed in hyperbolic space. HACK fully exploits the property of hyperbolic space and \pt\ arises naturally.
    \item We ground the concept of \pt\ based on congealing which conforms to human visual perception. The congealed examples can be used to replace the original examples for constructing datasets with known prototypicality. We validate the effectiveness of the method by using synthetic data with natural and congealed images. We further apply the proposed method to commonly used image datasets to reveal \pt. 
    \item The discovered prototypical and atypical examples are shown to reduce sample complexity and increase the robustness of the model.
\end{itemize}

\section{Related Work}

\noindent \textbf{Prototypicality.} The study of prototypical examples in machine learning has a long history. In \cite{zhang1992selecting}, the authors select typical instances based on the fact that typical instances should be representative of the cluster. In \cite{kim2016examples}, prototypical examples are defined as the examples that have maximum mean discrepancy within the data. Li et al. \cite{li2018deep} propose to discover prototypical examples by architectural modifications: project the dataset onto a low-dimensional manifold and use a prototype layer to minimize the distance between inputs and the prototypes on the manifold. The robustness to adversarial attacks is also used as a criterion for \pt\ \cite{stock2018convnets}. In \cite{carlini2018prototypical}, the authors propose multiple metrics for \pt\ discovery. For example, the features of prototypical examples should be consistent across different training setups. However, these metrics usually depend heavily on the training setups and hyperparameters. The idea of \pt\ is also extensively studied in meta-learning for one-shot or few-shot classification \cite{snell2017prototypical}. No existing works address the \pt\ discovery problem in a data-driven fashion. Our proposed HACK naturally exploits hyperbolic space to organize the images based on \pt.

\noindent \textbf{Unsupervised Learning in Hyperbolic Space.} Learning features in hyperbolic space have shown to be useful for many machine learning problems \cite{nickel2017poincar,ganea2018hyperbolic,guo2022clipped,mettes2023hyperbolic}. One useful property is that hierarchical relations can be embedded in hyperbolic space with low distortion \cite{nickel2017poincar}. Wrapped normal distribution, which is a generalized version of the normal distribution for modeling the distribution of points in hyperbolic space \cite{nagano2019wrapped}, is used as the latent space for constructing hyperbolic variational autoencoders (VAEs) \cite{kingma2013auto}. Poincar\'e VAEs is constructed in \cite{mathieu2019continuous} with a similar idea to \cite{nagano2019wrapped} by replacing the standard normal distribution with hyperbolic normal distribution. Unsupervised 3D segmentation \cite{hsu2020learning} and instance segmentation \cite{weng2021unsupervised} are conducted in hyperbolic space via hierarchical hyperbolic triplet loss. CO-SNE \cite{guo2021co} is recently proposed to visualize high-dimensional hyperbolic features in a two-dimensional hyperbolic space. Although hyperbolic distance facilitates the learning of hierarchical structure, how to leverage hyperbolic space for unsupervised \pt\ discovery is not explored in the current literature. 

\noindent \textbf{Sphere Packing.} Sphere packing aims to pack a set of particles as densely as possible in space \cite{conway2013sphere}. It can be served as a toy model for granular materials and has applications in information theory \cite{shannon2001mathematical} to find error-correcting codes \cite{cohn2016conceptual}. Sphere packing is difficult due to multiple local minima, the curse of high dimensionality, and complicated geometrical configurations. Packing in hyperbolic space is also studied in the literature. It is given in \cite{boroczky1978packing} a universal upper bound for the density of sphere packing in an $n$-dimensional hyperbolic space when $n$ $\ge$ 2. We are interested in generating uniform packing in a two-dimensional hyperbolic space. Uniformity has been shown to be a useful criterion for learning good features on the hypersphere \cite{wang2020understanding}. We opt to find the configuration with an optimization procedure that is easily applicable even with thousands of particles.

\section{Prototypicality as Density Peaks}

Given existing features $\{f(v_i)\}$ obtained by applying a feature extractor for each instance $v_i$, prototypical examples can be found by examining the density peaks via techniques from density estimation. For example, the K-nearest neighbor density (K-NN) estimation \cite{fix1989discriminatory} is defined as,
\begin{equation}
    p_{knn}(v_i, k) = \frac{k}{n} \frac{1}{A_d \cdot D^d(v_i, v_{k(i)})}
\end{equation}
where $d$ is the feature dimension, $A_d = \pi^{d/2} / \Gamma(d/2+1)$, $\Gamma(x)$ is the Gamma function and $k(i)$ is the $k$th nearest neighbor of example $v_i$. The nearest neighbors can be found by computing the distance between the features. Therefore, the process of identifying prototypicality through density estimation can be conceptualized as a two-step procedure involving: 1) feature learning and 2) detecting density peaks. 

In the density estimation approach outlined above, the level of prototypicality depends on the particular features learned. Varying training setups can induce diverse feature spaces, resulting in differing conclusions on prototypicality. Nevertheless, prototypicality is an inherent attribute of the dataset and should remain consistent across various features. The aim of this paper is to extract features that intrinsically showcase the prototypicality of the samples. Specifically, by examining the feature alone within the feature space, we should be able to identify the example's prototypicality.
\begin{figure}[t]
\centering 
   \setlength\tabcolsep{0.1pt}%
   \scalebox{0.9}{
    \begin{tabular}{ccccccc}
    \includegraphics[align=c,width=0.07\textwidth]{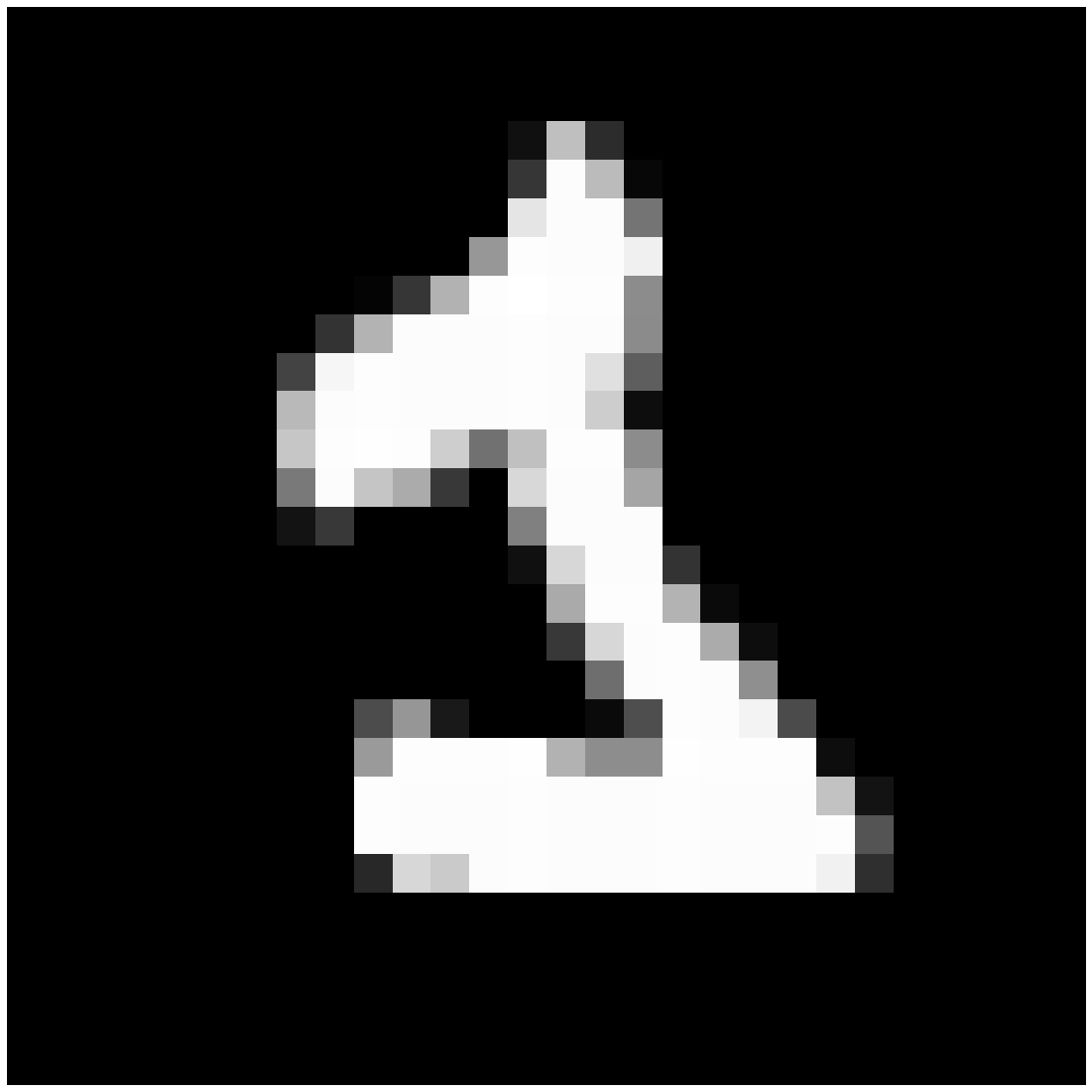} 
   & \includegraphics[align=c,width=0.07\textwidth]{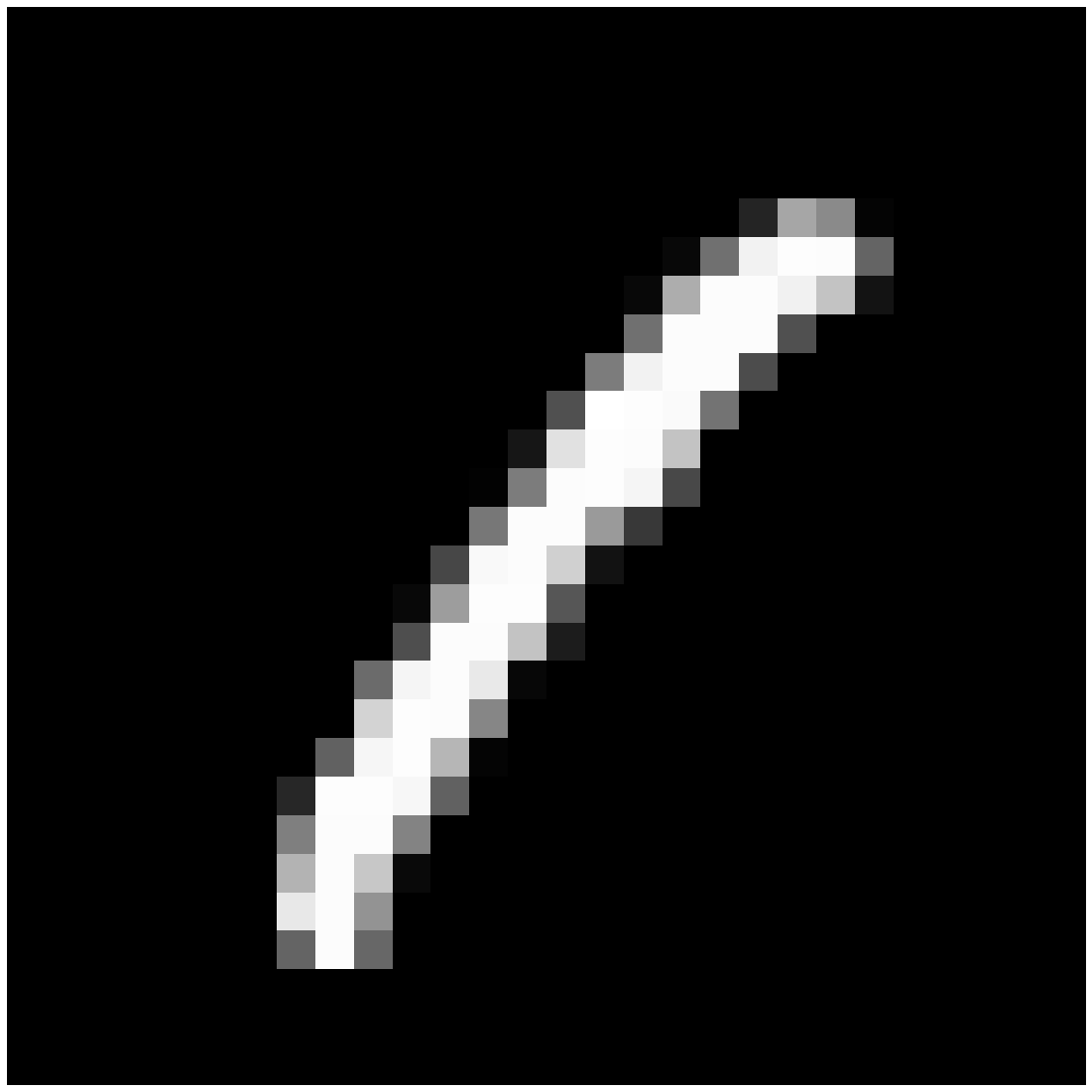} 
    & \includegraphics[align=c,width=0.07\textwidth]{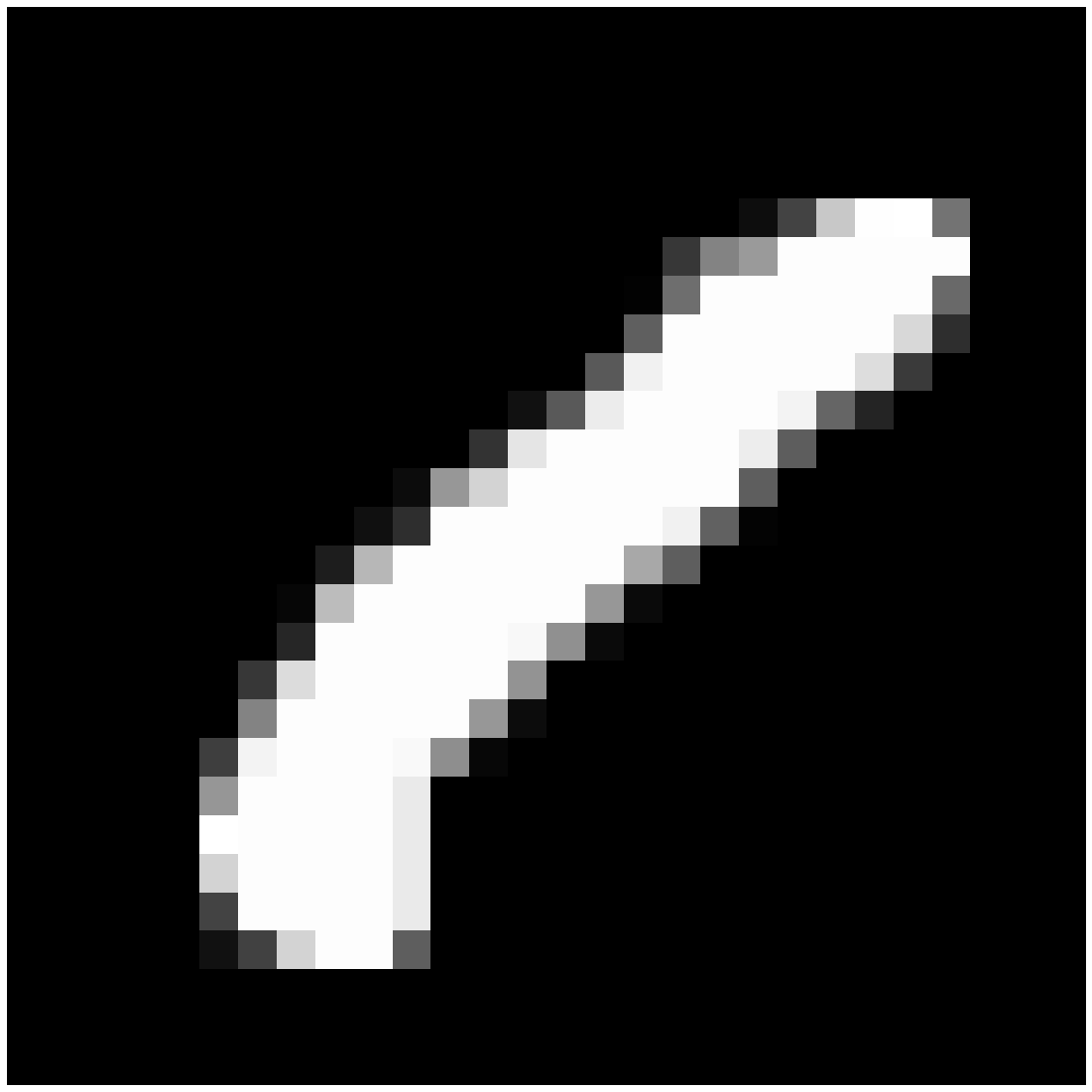} 
     & \includegraphics[align=c,width=0.07\textwidth]{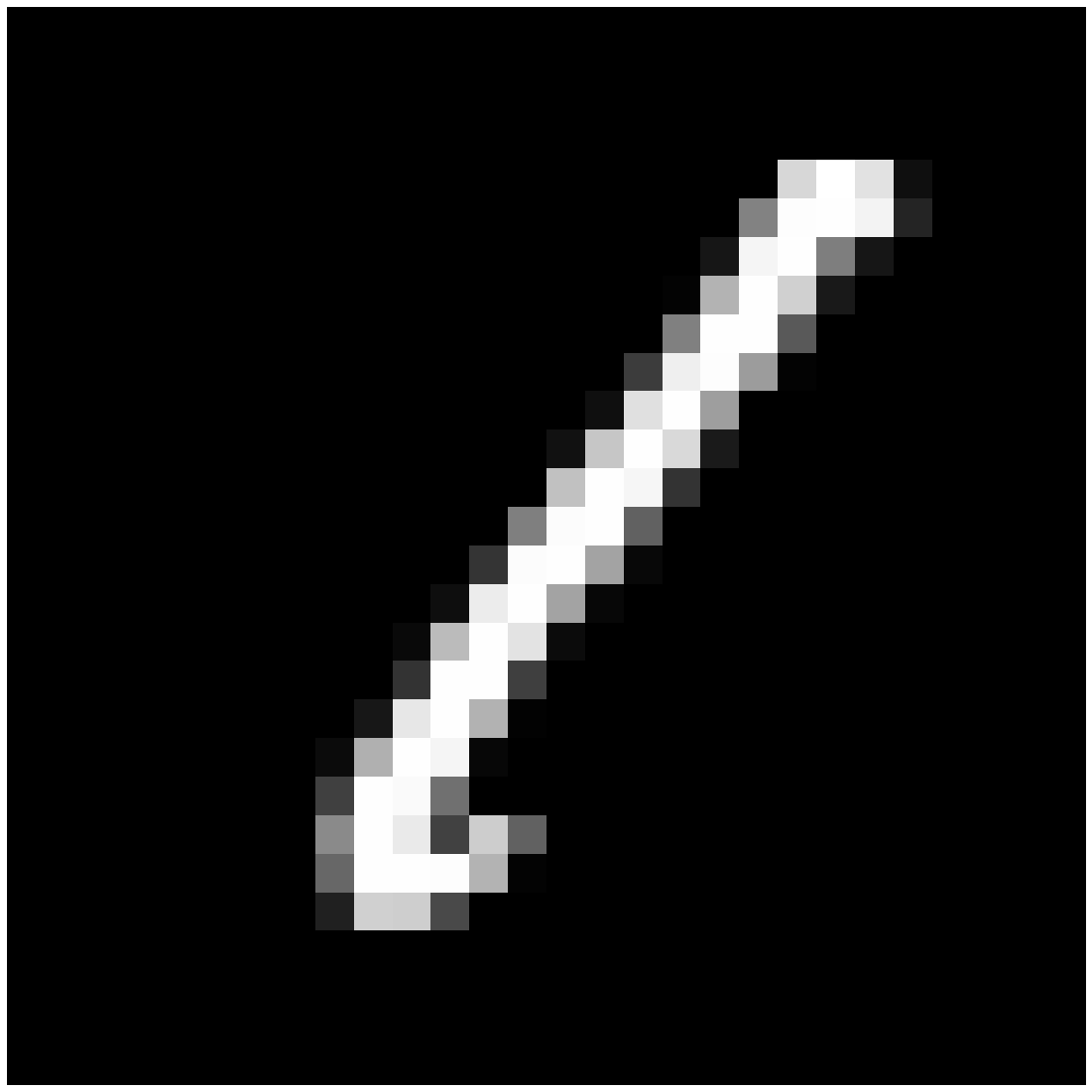} 
     & \includegraphics[align=c,width=0.07\textwidth]{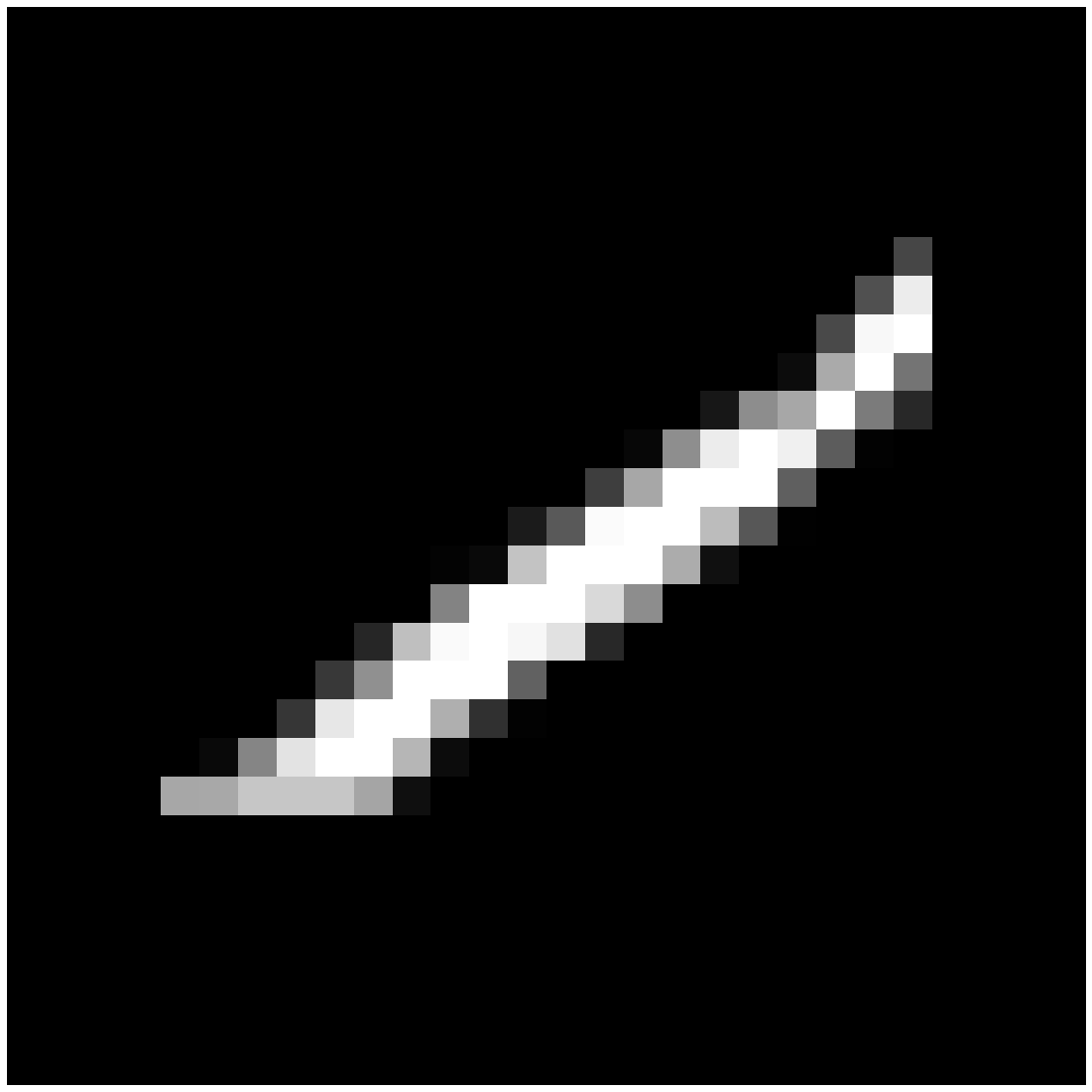} 
      & \includegraphics[align=c,width=0.07\textwidth]{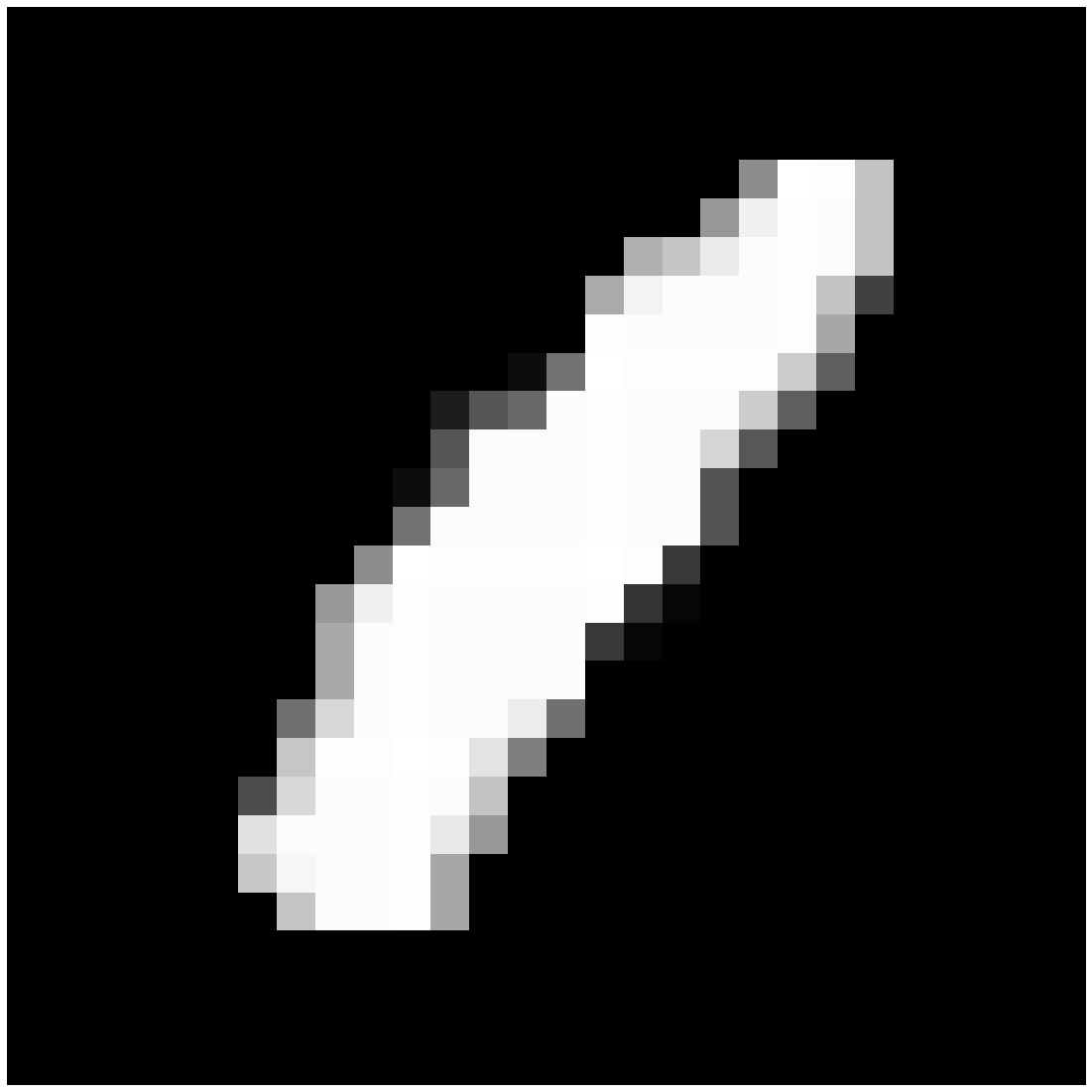} 
          & \includegraphics[align=c,width=0.07\textwidth]{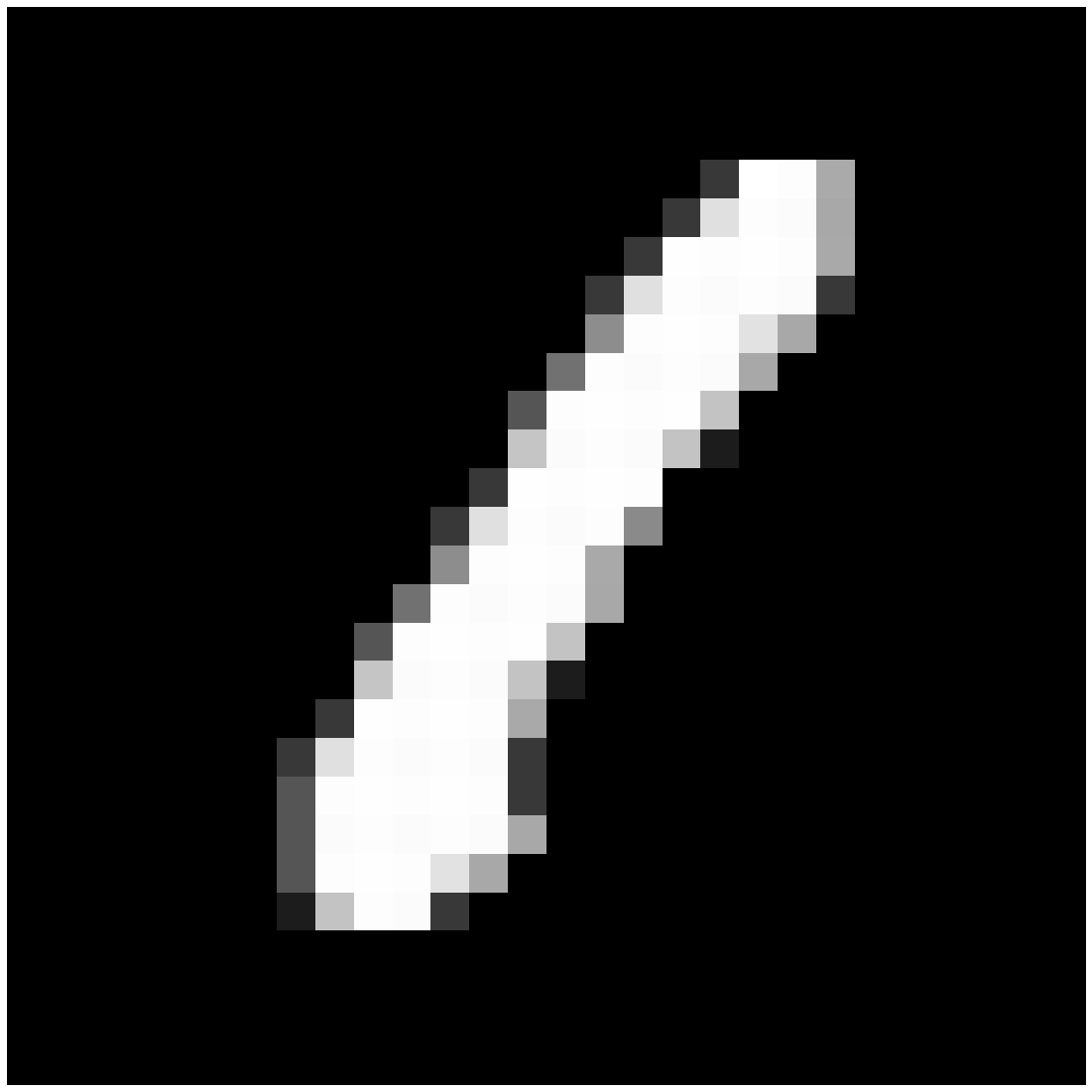} \\
    \includegraphics[align=c,width=0.07\textwidth]{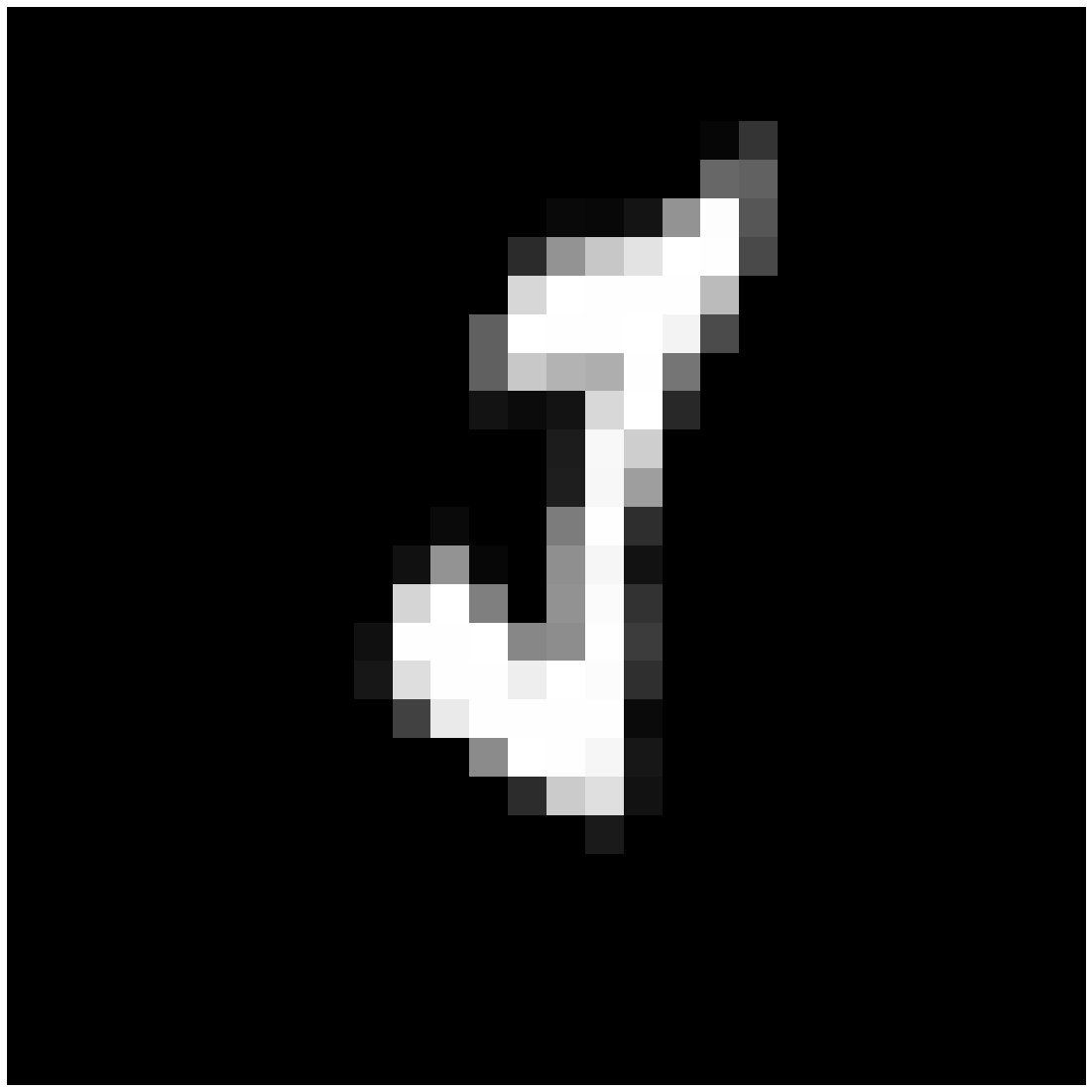} 
   & \includegraphics[align=c,width=0.07\textwidth]{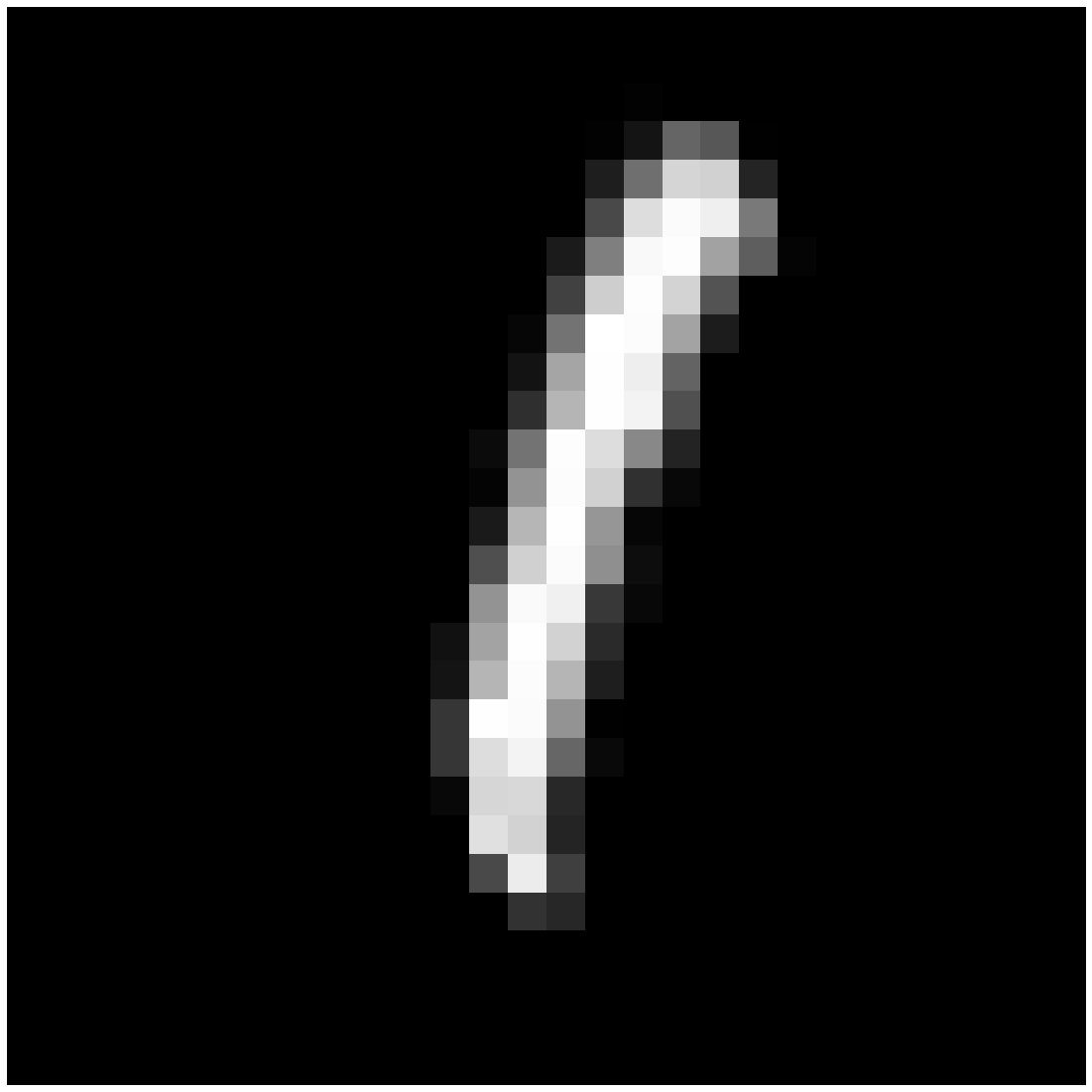} 
    & \includegraphics[align=c,width=0.07\textwidth]{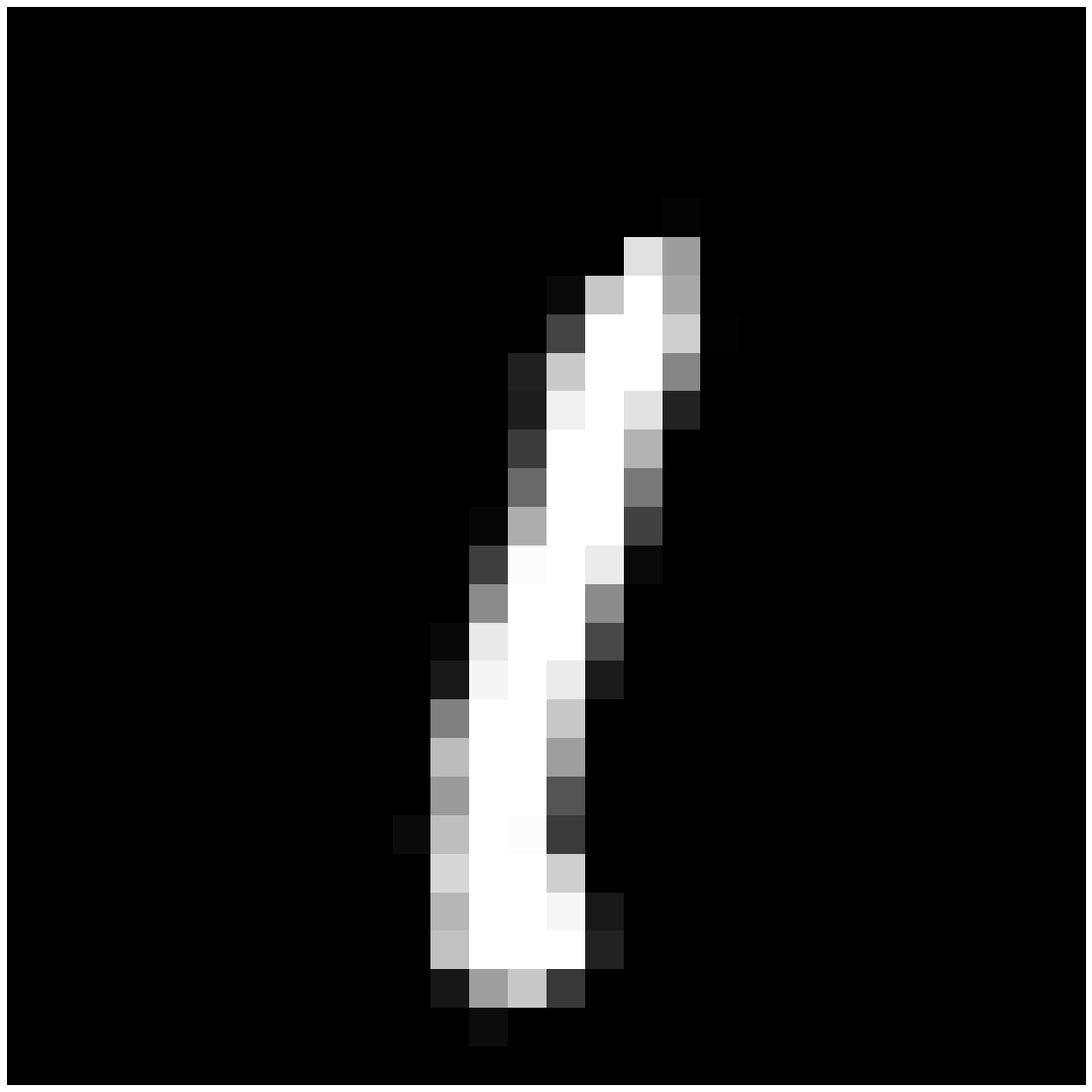} 
     & \includegraphics[align=c,width=0.07\textwidth]{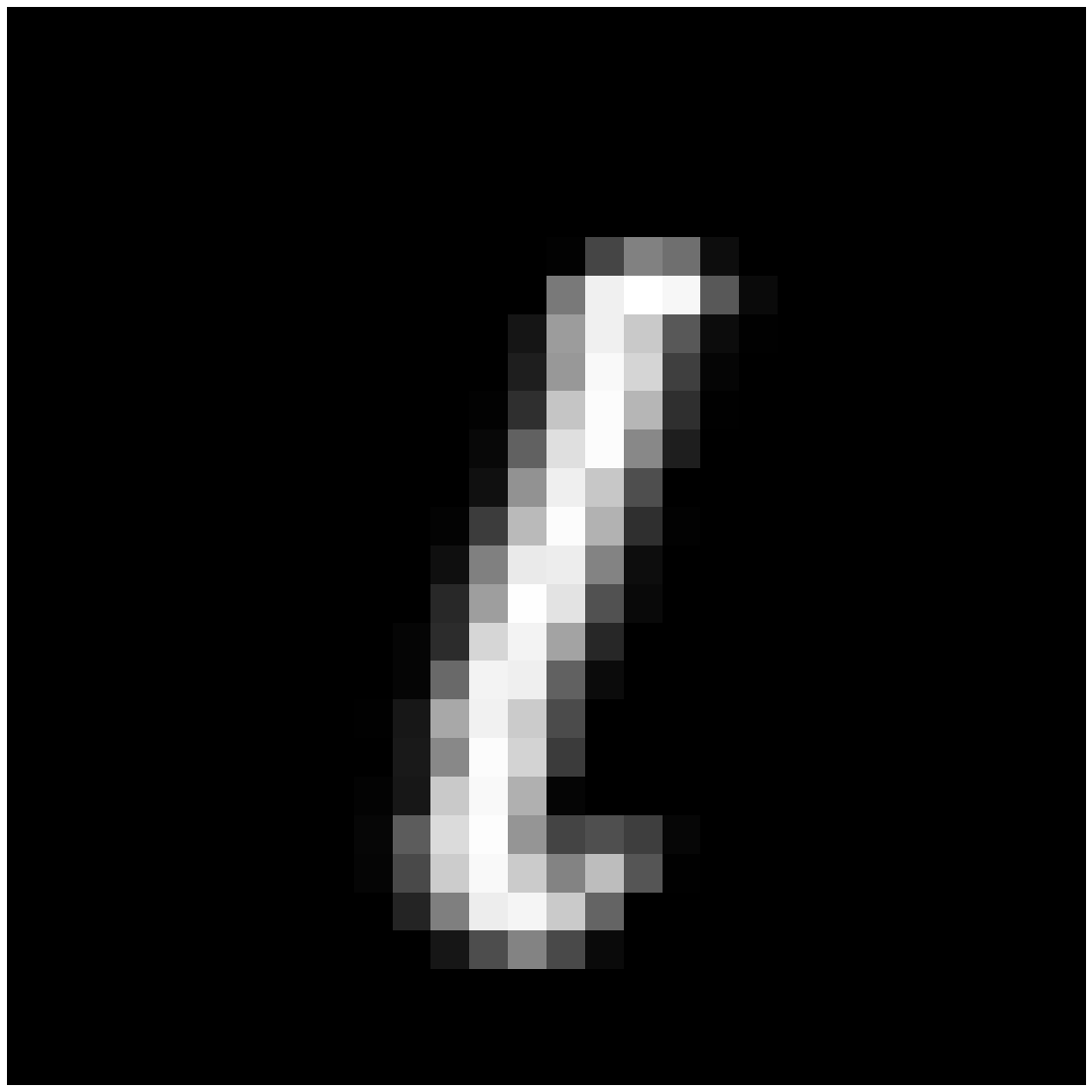} 
     & \includegraphics[align=c,width=0.07\textwidth]{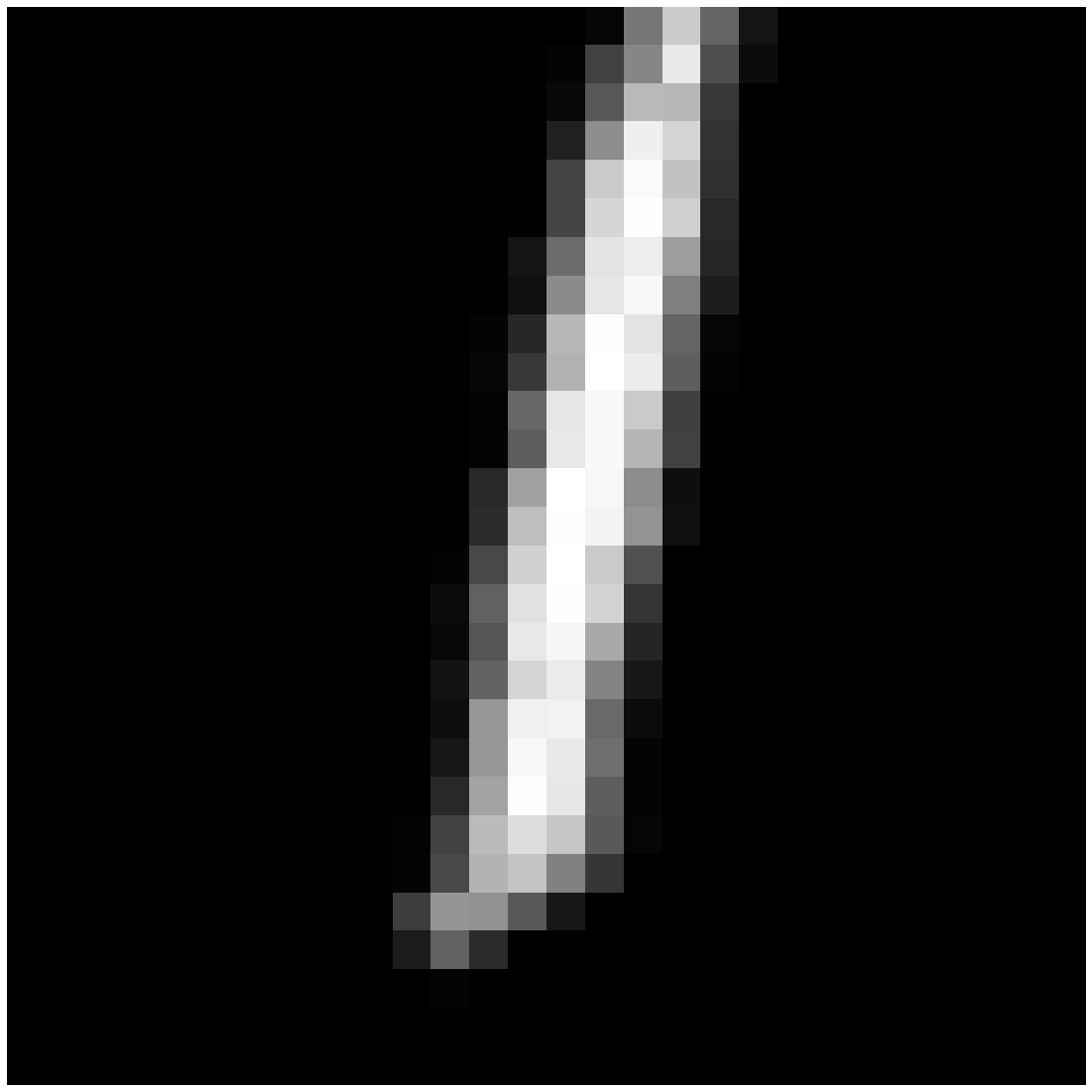} 
      & \includegraphics[align=c,width=0.07\textwidth]{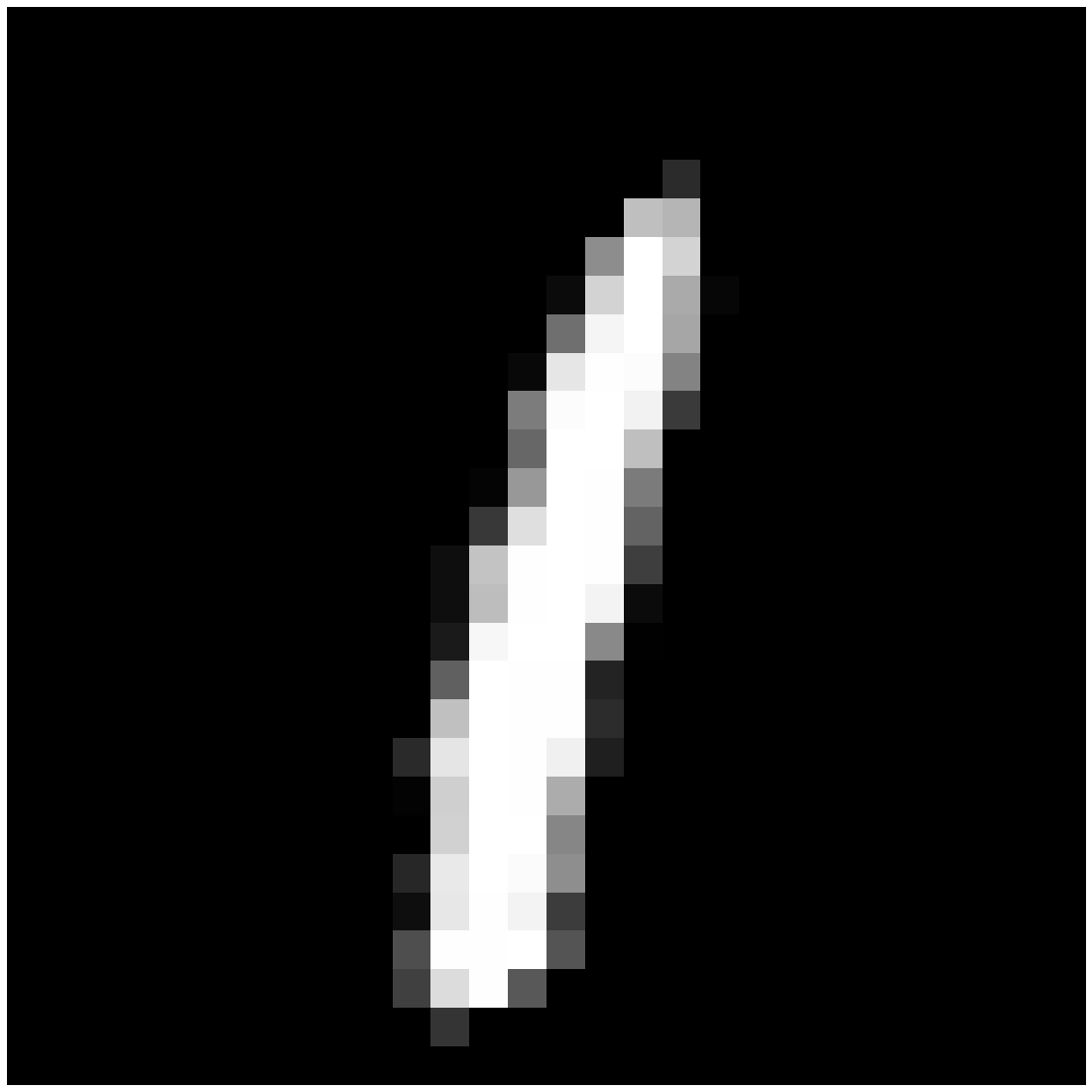} 
          & \includegraphics[align=c,width=0.07\textwidth]{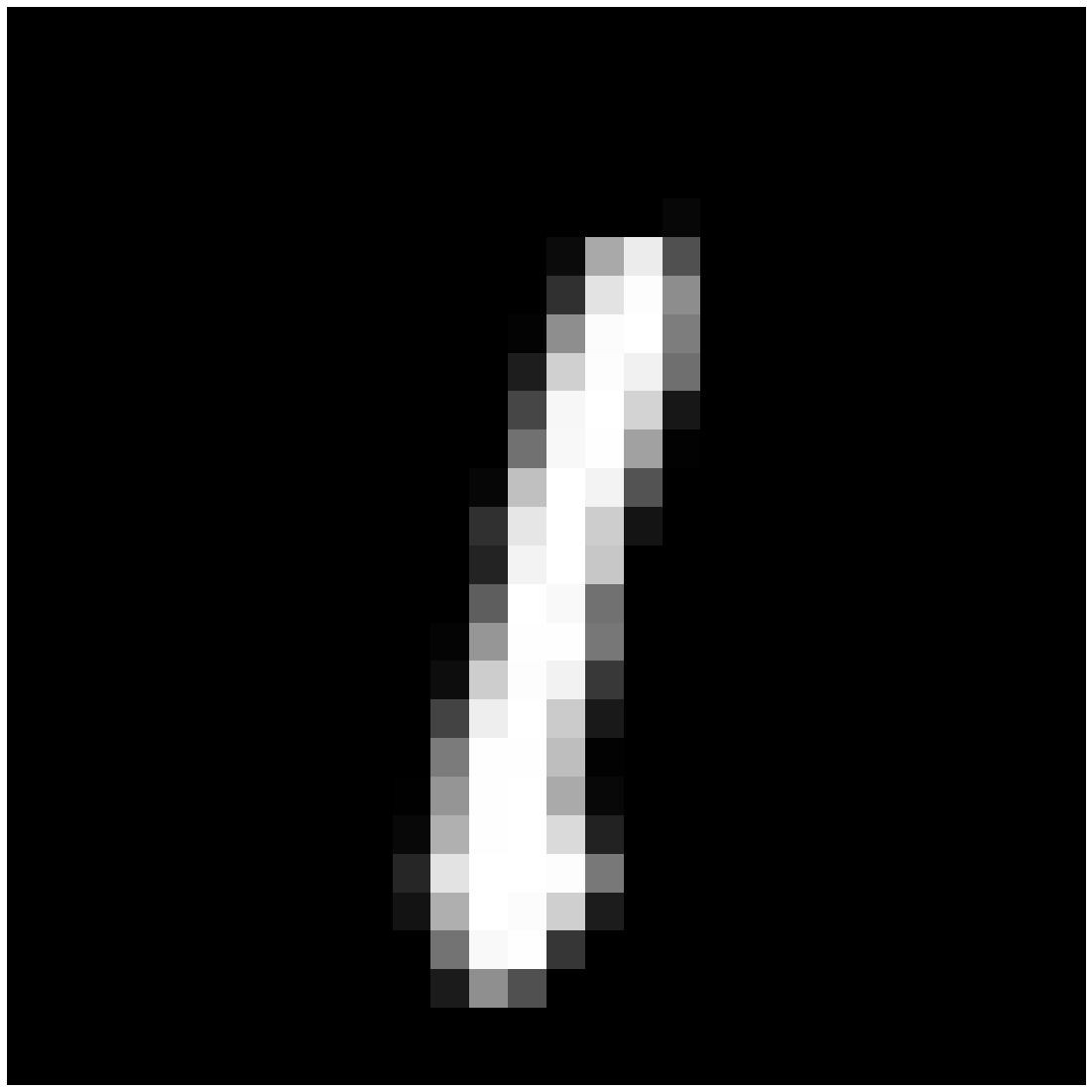} \\
        \end{tabular}}
    \caption{\textbf{Congealed images are more typical than the original images.} First row: sampled original images. Second row: the corresponding congealed images.}
    \label{fig:congealing_discover}
\end{figure}

To determine whether the feature truly captures prototypicality, it is necessary to identify which sample is the prototype. We ground our concept of prototypicality based on congealing \cite{miller2000learning}. In particular, we define prototypical examples in the \emph{pixel space} by examining the distance of the images to the average image in the corresponding class. Our idea is based on a traditional computer vision technique called image alignment \cite{szeliski2007image} that aims to find correspondences across images. During congealing \cite{miller2000learning}, a set of images are transformed to be jointly aligned by minimizing the joint pixel-wise entropies. The congealed images are more prototypical: they are better aligned with the average image. Thus, we have a simple way to transform an atypical example into a typical example (see Figure~\ref{fig:congealing_discover}). This is useful since given an unlabeled image dataset the typicality of the examples is unknown, congealing examples can be naturally served as examples with known typicality and be used as a validation for the effectiveness of our method.

\begin{figure*}
   \centering
    \setlength{\tabcolsep}{0.02\linewidth}
\begin{tabular}{@{\hspace{-3mm}}ccc@{}}
   \includegraphics[height=0.15\textheight]{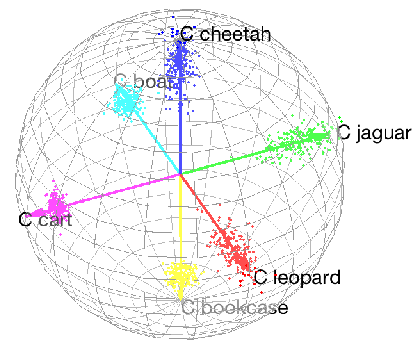} &
   \includegraphics[height=0.15\textheight]{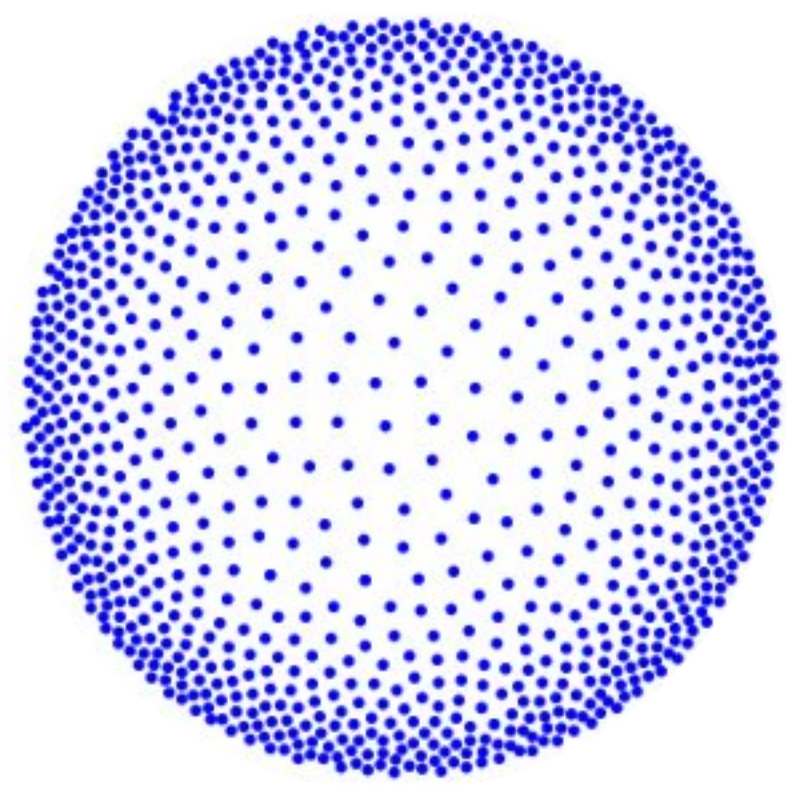}
    &
   \includegraphics[height=0.15\textheight]{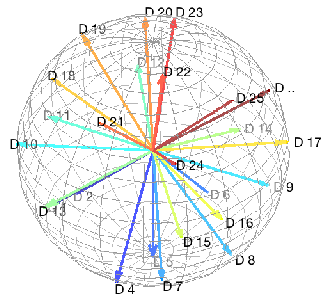} \\
   {\bf a)} Supervised classification &
   {\bf b)} Our unsupervised feature learning &
   {\bf c)} Metric feature learning\\
   with fixed known targets& 
   with fixed but {\it unknown} targets&
   with unknown targets\\
\end{tabular}
    \caption{\textbf{The proposed HACK has a predefined geometrical arrangement and allows the images to be freely assigned to any particle.} a) Standard supervised learning has predefined targets. The image is only allowed to be assigned to the corresponding target. b) HACK packs particles uniformly in hyperbolic space to create initial seeds for organization. The images are assigned to the particles based on their \pt\ and semantic similarities. c) Standard unsupervised learning has no predefined targets and images are clustered based on their semantic similarities.}
    \label{fig: comparison}
\end{figure*}

\begin{figure}[t]
    \centering
    \subfloat[\centering ]{{ \includegraphics[width=0.45\linewidth]{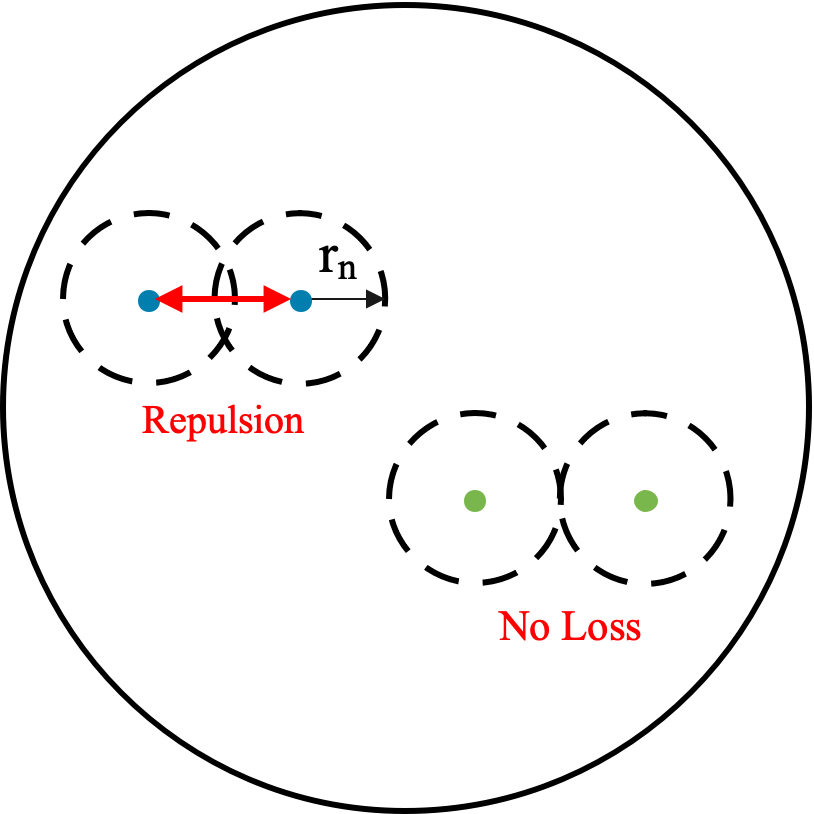}}}
    \subfloat[\centering]{{ \includegraphics[width=0.55\linewidth]{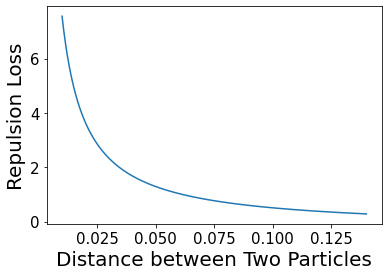} }}
    \caption{\textbf{The proposed repulsion loss is used to generate uniformly packed particles in hyperbolic space. } (a) If the distance between two particles is within $r_{n,r}$, minimizing the repulsion loss would push them away. (b) The repulsion loss is larger when the two particles become closer.}
    \label{fig:repulsion_loss}
\end{figure}

\section{Unsupervised Hyperbolic Feature Learning}
We aim to develop a method that can automatically discover prototypical examples unsupervisedly. In particular, we conduct unsupervised learning in hyperbolic space with sphere packing (Figure \ref{fig:overview}). We specify where the targets should be located ahead of training with uniform packing, which by design are maximally evenly spread out in hyperbolic space. The uniformly distributed particles guide feature learning to achieve maximum instance discrimination \cite{wu2018unsupervised}.

HACK figures out which instance should be mapped to which target through bipartite graph matching as a global optimization procedure. During training HACK minimizes the total hyperbolic distances between the mapped image point (in the feature space) and the target, those that are more typical naturally emerge closer to the origin of Poincar\'e ball. Prototypicality comes for free as a result of self-organization. HACK differs from the existing learning methods in several aspects (Figure \ref{fig: comparison}). Different from supervised learning, HACK allows the image to be assigned to \emph{any} target (particle). This enables the exploration of the natural organization of the data. 
On the other hand, existing unsupervised learning methods often employ maximal instance discrimination as a criterion for feature learning. However, if these approaches are directly applied to learning features in hyperbolic space, they will drive all instances towards the boundary to achieve maximal instance discrimination. Instead, HACK specifies a predefined geometrical organization which encourages the corresponding structure to be emerged from the dataset.



\subsection{Poincar\'e Ball Model for Hyperbolic Space}
\label{sec:p_ball}
\noindent \textbf{Hyperbolic space.} Euclidean space has a curvature of zero and a hyperbolic space is a Riemannian manifold with constant negative curvature.

\noindent \textbf{Poincar\'e Ball Model for Hyperbolic Space.} There are several isometrically equivalent models for visualizing hyperbolic space with Euclidean representation. The Poincar\'e ball model is the commonly used one in hyperbolic representation learning \cite{nickel2017poincare}. The $n$-dimensional Poincar\'e  ball model is defined as $(\mathbb{B}^n, \mathfrak{g}_\mathbf{x})$, where $\mathbb{B}^n$ = $\{\mathbf{x} \in \mathbb{R}^n: \lVert \mathbf{x} \rVert < 1 \}$ and $\mathfrak{g}_\mathbf{x}  = (\gamma_\mathbf{x})^2 I_n $ is the Riemannian metric tensor. $\gamma_\mathbf{x} = \frac{2}{1- \lVert \mathbf{x} \rVert^2}$ is the conformal factor and $I_n$ is the Euclidean metric tensor.\\

\noindent \textbf{Hyperbolic Distance.} Given two points $\bm{u} \in \mathbb{B}^n$ and $\bm{v} \in \mathbb{B}^n$, the hyperbolic distance is defined as,
\begin{equation}d_{\mathbb{B}^n}(\bm{u}, \bm{v}) = \arcosh\left(1 + 2\frac{\lVert\bm{u}-\bm{v}\rVert^2}{(1-\lVert\bm{u}\rVert^2)(1-\lVert\bm{v}\rVert^2)}\right)\label{eq1}\end{equation}
where $\arcosh$ is the inverse hyperbolic cosine function and $\lVert \cdot \rVert$ is the usual Euclidean norm.

Hyperbolic distance has the unique property that it grows exponentially as we move towards the boundary of the Poincar\'e ball. In particular, the points on the circle represent points in infinity. Hyperbolic space is naturally suitable for embedding hierarchical structure \cite{sarkar2011low,nickel2017poincare} and can be regarded as a continuous representation of trees \cite{chami2020trees}. The hyperbolic distance between samples implicitly reflects their hierarchical relation. Thus, by embedding images in hyperbolic space we can naturally organize images based on their semantic similarity and \pt.

\subsection{Sphere Packing in Hyperbolic Space}
\label{sec:sphere_packing}
Given $n$ particles, our goal is to pack the particles into a two-dimensional hyperbolic space as densely as possible. We derive a simple repulsion loss function to encourage the particles to be equally distant from each other. The loss is derived via the following steps. First, we need to determine the radius of the Poincar\'e ball used for packing. We use a curvature of 1.0 so the radius of the Poincar\'e ball is 1.0. The whole Poincar\'e ball cannot be used for packing since the volume is infinite. We use $r < 1$ to denote the actual radius used for packing. Thus, our goal is to pack $n$ particles in a compact subspace of Poincar\'e ball. Then, the Euclidean radius $r$ is further converted into hyperbolic radius $r_{\mathbb{B}}$. Let $s = \frac{1}{\sqrt{c}}$, where $c$ is the curvature. The relation between $r$ and $r_{\mathbb{B}}$ is $ r_{\mathbb{B}} = s \log \frac{s + r}{s - r}$. Next, the total hyperbolic area $A_{\mathbb{B}}$ of a Poincar\'e ball of radius $r_{\mathbb{B}}$ can be computed as $ A_{\mathbb{B}} = 4\pi s^2 \sinh^2(\frac{r_{\mathbb{B}}}{2s})$, where $\sinh$ is the hyperbolic sine function. Finally, the area per point $A_{n}$ can be easily computed as $\frac{ A_{\mathbb{B}}}{n}$, where $n$ is the total number of particles. Given $A_{n}$, the radius per point can be computed as $ r_{n} = 2s \sinh^{-1}(\sqrt{\frac{A_{n}}{ 4 \pi s^2}})$. We use the following loss to generate uniform packing in hyperbolic space. Given two particles $i$ and $j$, the repulsion loss $V$ is defined as,
\begin{equation}
    \small V(i, j) =
        \{  \frac{1}{ [2r_{n} - \max (0, 2r_{n} - d_{\mathbb{B}}(i,j))]^k} - \frac{1}{(2r_{n})^k}  \} \cdot C(k)
    \label{eq:repulsion}
\end{equation}
where $C(k) = \frac{(2r_n)^{k+1}}{k}$ and $k$ is a hyperparameter. Intuitively, if the particle $i$ and the particle $j$ are within $2r_{n}$, the repulsion loss is positive. Minimizing the repulsion loss would push the particles $i$ and $j$ away. If the repulsion is zero, this indicates all the particles are equally distant (Figure \ref{fig:repulsion_loss} a). Figure \ref{fig:repulsion_loss} b) shows that the repulsion loss grows significantly when the two particles become close.

We also adopt the following boundary loss to prevent the particles from escaping the ball,
\begin{equation}
    B(i; r) = \max (0, \textnormal{norm}_{i} - r + \textnormal{margin})
\end{equation}
where $\textnormal{norm}_{i}$ is the $\ell_2$ norm of the representation of the particle $i$. Figure \ref{fig: comparison} b) shows an example of the generated particles that are uniformly packed in hyperbolic space.

\begin{figure}
\begin{center}
\includegraphics[width=\linewidth]{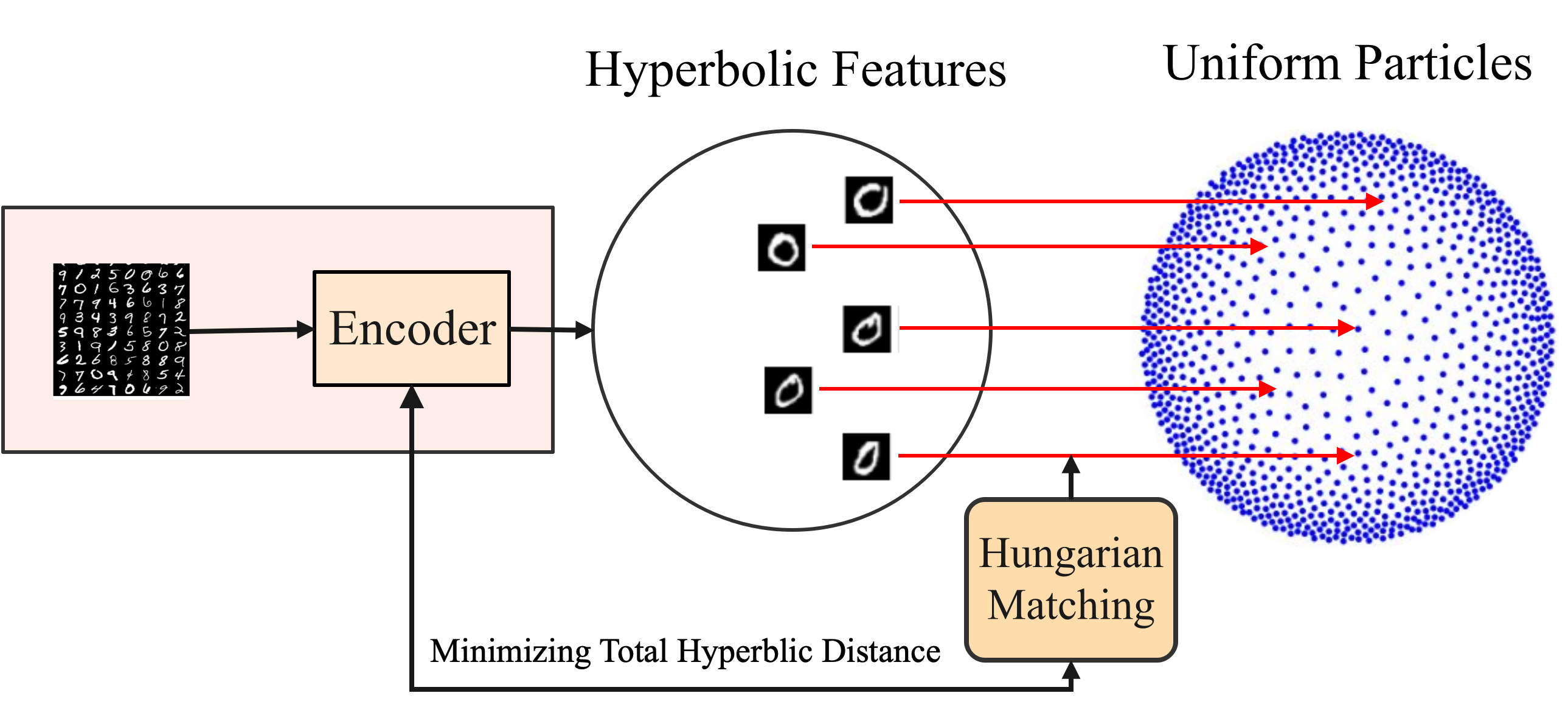}
\end{center}
   \caption{\textbf{HACK conducts unsupervised learning in hyperbolic space with sphere packing.} The images are mapped to particles by minimizing the total hyperbolic distance. HACK learns features that can capture both visual similarities and \pt.}
\label{fig:overview}
\end{figure}

\subsection{Hyperbolic Instance Assignment}
\label{sec:instance_assigment}

HACK learns the features by optimizing the assignments of the images to particles (Figure \ref{fig:overview}). Once we generate a fixed set of uniformly packed particles in a two-dimensional hyperbolic space, our next goal is to assign each image to the corresponding particle. The assignment should be one-to-one, i.e., each image should be assigned to one particle and each particle is allowed to be associated with one image. We cast the instance assignment problem as a bipartite matching problem \cite{gibbons1985algorithmic} and solve it with Hungarian algorithm \cite{munkres1957algorithms}.

Initially, we randomly assign the particles to the images, thus there is a random one-to-one correspondence between the images to the particles (not optimized). Given a batch of samples $\{(\mathbf{x}_1, s_1), (\mathbf{x}_2, s_2), ..., (\mathbf{x}_b, s_b)\}$, where $\mathbf{x}_i$ is an image and $s_i$ is the corresponding particle, and an encoder $f_\theta$, we generate the hyperbolic feature for each image $\mathbf{x}_i$ as $f_\theta(\mathbf{x}_i) \in \mathbb{B}^2$, where $\mathbb{B}^2$ is a two-dimensional Poincar\'e ball. We aim to find the minimum cost bipartite matching of the images to the particles within this batch. It is worth noting that the assignment is done without supervision.

In bipartite matching, the cost is the hyperbolic distance of each image to the particle. Thus, the criterion is to minimize the total hyperbolic distances of the assignment. We achieve this goal with the Hungarian algorithm \cite{munkres1957algorithms} which has a complexity of $\mathcal{O}(b^3)$, where $b$ is the batch size. It is worth noting that the assignment is only limited to the samples in the particular batch, thus the time and memory complexity is tolerable. The one-to-one correspondence between the images and particles is always maintained during training. The details of HACK are shown in Algorithm \ref{alg:hack_algo}.

\begin{algorithm}[t]
\caption{HACK: Unsupervised Learning in Hyperbolic Space. }
\begin{algorithmic}[1]
\Require \# of images: $n \geq 0$. Radius for packing: $r < 1$. An encoder with parameters $\theta$: $f_\theta$ 
\State Generate uniformly distributed particles in hyperbolic space by minimizing the repulsion loss in Equation \ref{eq:repulsion}
\State Given $\{(\mathbf{x}_1, s_1), (\mathbf{x}_2, s_2), ..., (\mathbf{x}_b, s_b)\}$, optimize $f_\theta$ by minimizing the total hyperbolic distance via Hungarian algorithm.
\end{algorithmic}
\label{alg:hack_algo}
\end{algorithm}

Due to the property of hyperbolic distance, the images that are more typical tend to be assigned to the particles located near the origin. Thus, HACK implicitly defines \pt\ as the distance of the sample to the others. The \pt\ of the images can be easily reflected by the location of the assigned particles. Moreover, similar images tend to cluster together due to semantic similarity. In summary, with hyperbolic instance assignment, HACK automatically organizes images based on \pt\ by exploiting the hyperbolicity of the space.

\subsection{Discussion}
\noindent \textbf{Hoes Does HACK Work?} 
Hyperbolic space can embed tree structures with no distortion. In particular, the root of the tree can be embedded in the center of the Poincar\'e ball and the leaves are embedded close to the boundary \cite{nickel2017poincar,ganea2018hyperbolic}. Thus, the root is close to all the other nodes. This agrees with our intuition that typical examples should be close to all other examples. By minimizing the total assignment loss of the images to the particles, we seek to organize the images implicitly in a tree-structure manner. Consider three images $A$, $B$, $C$ for an example. Assume image $A$ is the most typical image. Thus the feature of $A$ is close to both the features of $B$ and $C$. The bipartite matching tends to assign image $A$ to the particle in the center since this naturally reflects the feature distances between the three images.\\

\noindent \textbf{Connection to Existing Methods.} Existing works address the problem of \pt\ discovery with ad-hoc defined metrics \cite{carlini2018prototypical}. These metrics usually have high variances due to different training setups or hyperparameters. In this paper, we take a different perspective by exploiting the natural organization of the data by optimizing hyperbolic instance assignments. The property of hyperbolic space facilitates the discovery of \pt. Also, popular contrastive learning based unsupervised learning methods such as SimCLR \cite{chen2020simple} and MoCo \cite{he2020momentum} cannot achieve this goal since the predefined structure is not specified.

\begin{figure}[t]
    \centering
    \begin{tabular}{cc}
       \includegraphics[height=0.22\textwidth]{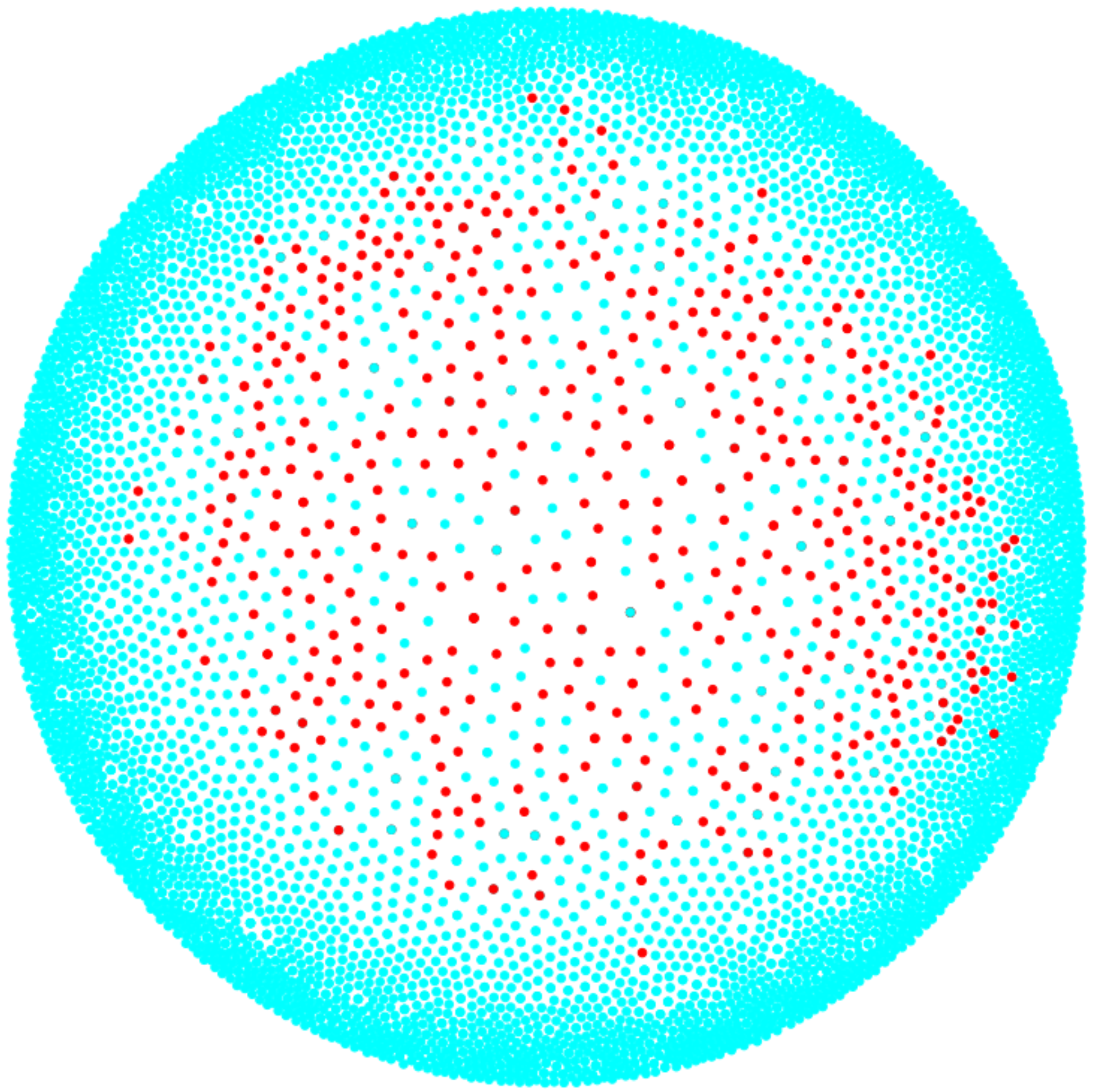} &
       \includegraphics[height=0.22\textwidth]{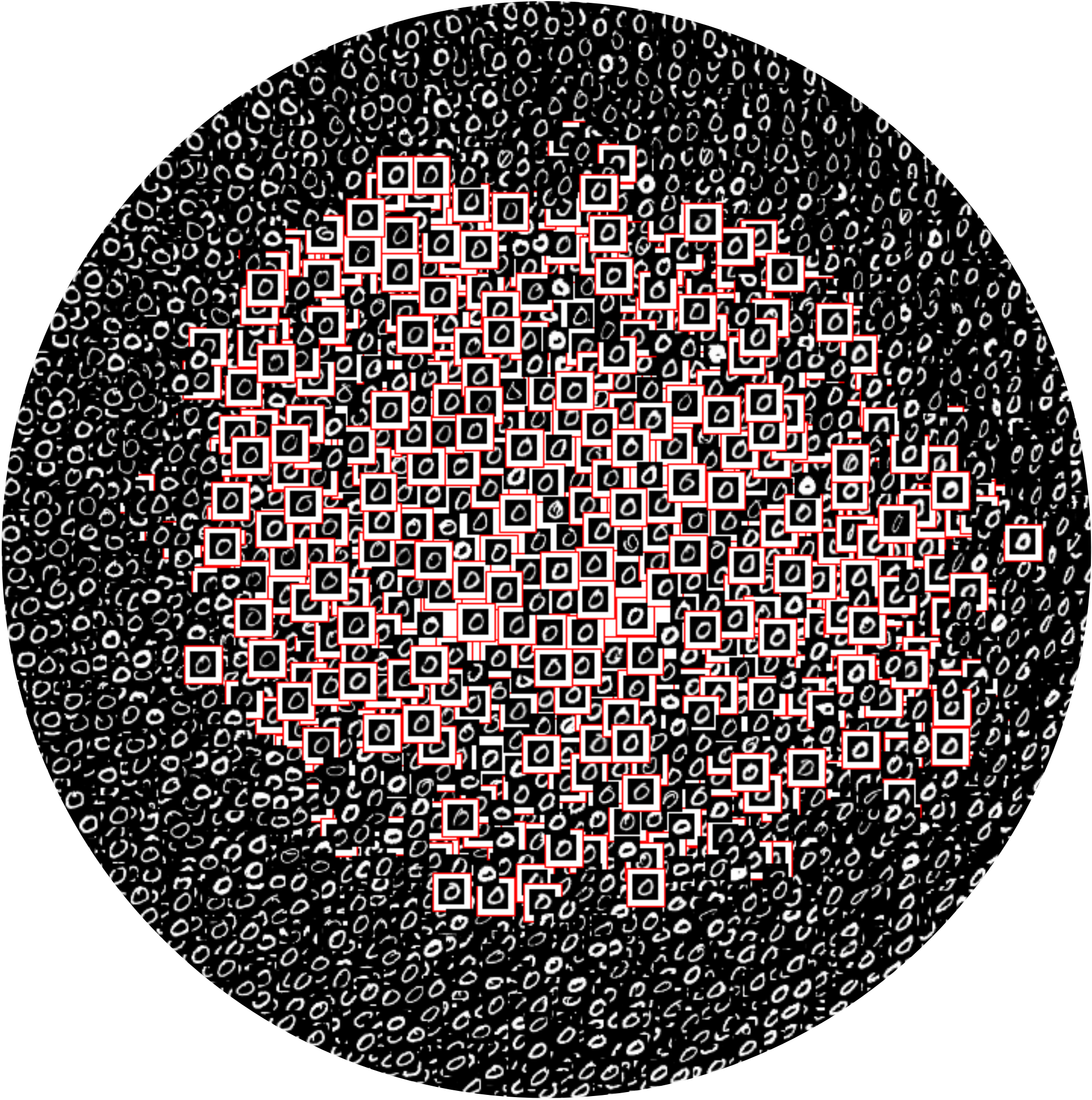} \\
       a) & b) 
    \end{tabular}
    \caption{\textbf{Congealed images are located in the center of the Poincar\'e ball.} a) \textcolor{red}{Red} dots denote congealed images and \textcolor{cyan}{cyan} dots denote original images. b) Typical images are in the center and atypical images are close to the boundary. Images are also clustered together based on visual similarity. Congealed images are shown in \textcolor{red}{red} boxes.  }
        \label{fig:congealing_embeddings}
\end{figure}

\section{Experiments}
We design several experiments to show the effectiveness of HACK for the semantic and prototypical organization. First, we first construct a dataset with known \pt\ using the congealing algorithm \cite{miller2000learning}. Then, we apply HACK to datasets with unknown \pt\ to organize the samples based on the semantic and prototypical structure. Finally, we show that the prototypical structure can be used to reduce sample complexity and increase model robustness.\\

\noindent \textbf{Datasets.} We first construct a dataset called \emph{Congealed MNIST.} To verify the efficacy of HACK for unsupervised \pt\ discovery, we need a benchmark with known prototypical examples. However, currently there is no standard benchmark for this purpose. To construct the benchmark, we use the congealing algorithm from \cite{miller2000learning} to align the images in each class of MNIST \cite{lecun1998mnist}. The congealing algorithm is initially used for one-shot classification. During congealing, the images are brought into correspondence with each other jointly. The congealed images are more prototypical: they are better aligned with the average image. In Figure \ref{fig:congealing_discover}, we show the original images and the images after congealing. The original images are transformed via affine transformation to better align with each other. The synthetic data is generated by replacing 500 original images with the corresponding congealed images. In Section \ref{sec:gradual} of the Appendix, we show the results of changing the number of replaced original images. We expect HACK to discover the congealed images and place them in the center of the Poincar\'e ball. We also aim to discover the prototypical examples from each class of the standard MNIST dataset \cite{lecun1998mnist} and CIFAR10 \cite{krizhevsky2009learning}. CIFAR10 consists of 60000 from 10 object categories ranging from airplane to truck. CIFAR10 is more challenging than MNIST since it has larger intra-class variations.\\

\begin{figure}
    \centering
    \begin{tabular}{cc}
       \includegraphics[height=0.22\textwidth]{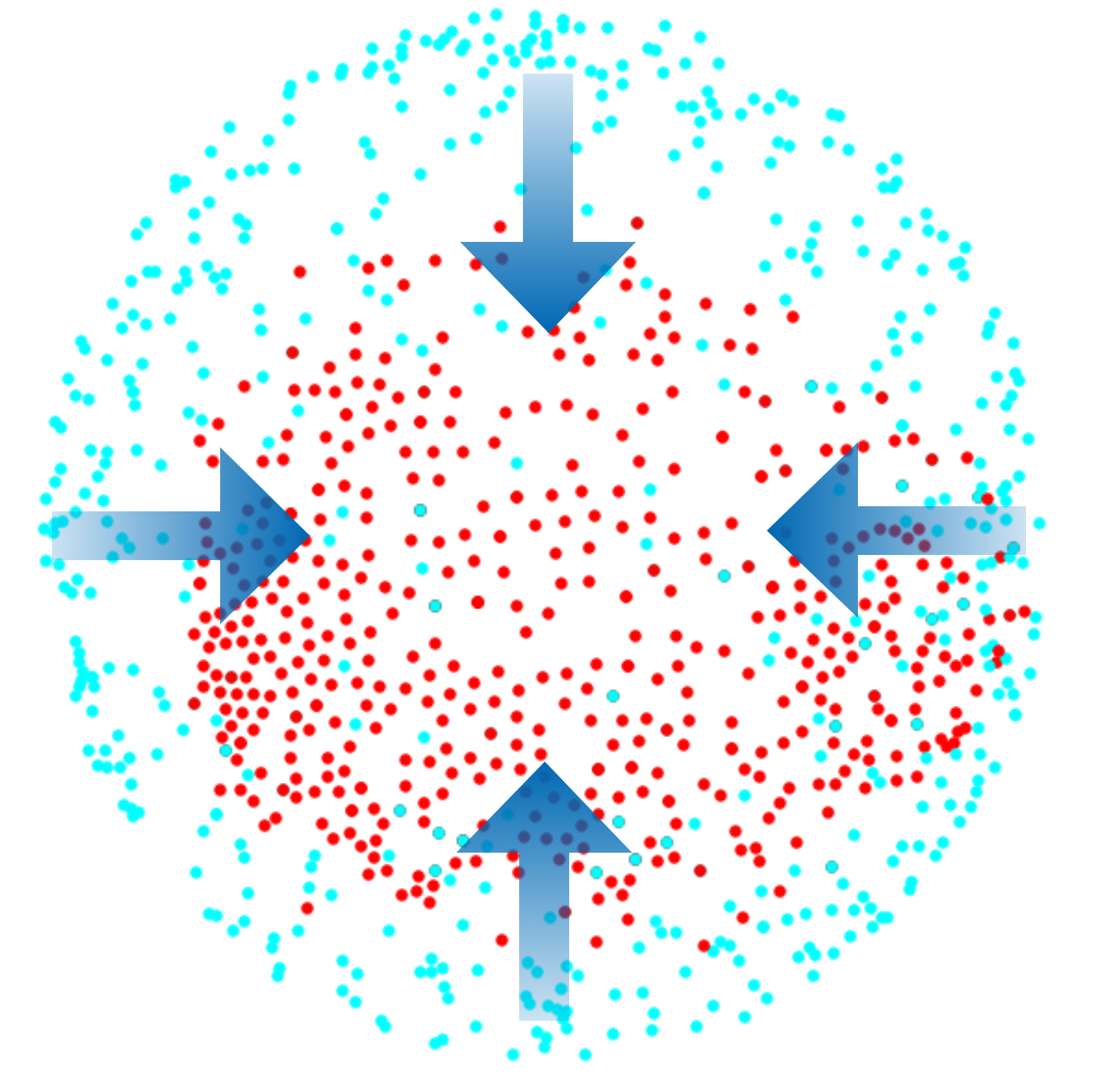} &
       \includegraphics[height=0.22\textwidth]{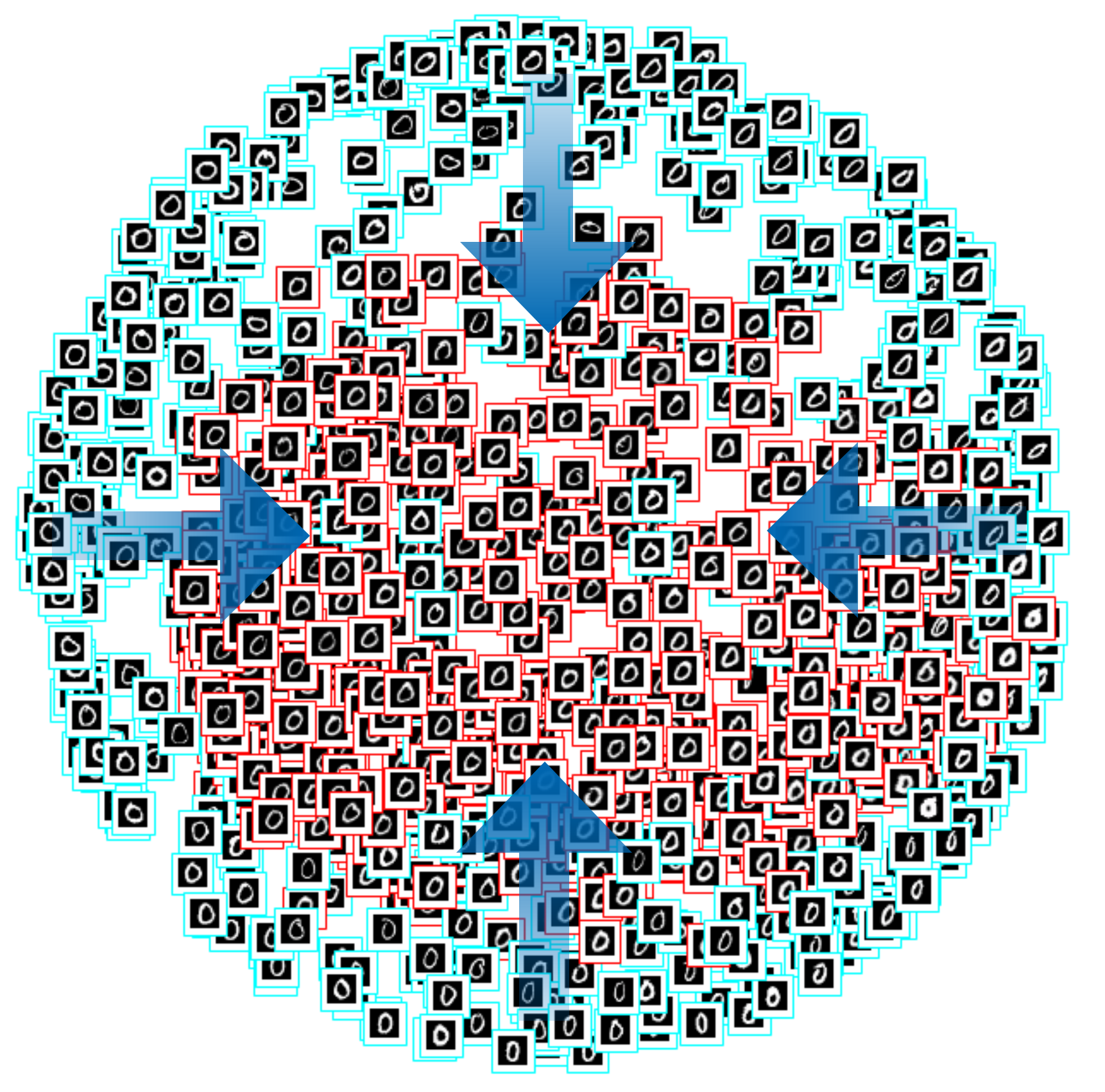} \\
       a) & b) 
    \end{tabular}
    \caption{\textbf{Original images are pushed to the center of the ball after congealing.} We train the first model with original images and train the second model by replacing a subset of original images (marked with \textcolor{cyan}{cyan}) with the corresponding congealed images. The features of the congealed images (marked with \textcolor{red}{red}) become closer to the center of the ball.}
        \label{fig:congealing_move}
\end{figure}

\noindent \textbf{Baselines.} We consider several existing metrics proposed in \cite{carlini2018prototypical} for \pt\ discovery, the details can be found in Section \ref{sec:baseline} of the Appendix.

\begin{itemize}[leftmargin=*,topsep=1pt,itemsep=-4pt]

    \item Holdout Retraining \cite{carlini2018prototypical}: We consider the Holdout Retraining proposed in \cite{carlini2018prototypical}. The idea is that the distance of features of prototypical examples obtained from models trained on different datasets should be close.
    \item Model Confidence: Intuitively, the model should be confident in prototypical examples. Thus, it is natural to use the confidence of the model prediction as the criterion for \pt.

\end{itemize}



\noindent \textbf{Implementation Details.} We implement HACK in PyTorch and the code will be made public. To generate uniform particles, we first randomly initialize the particles and then run the training for 1000 epochs to minimize the repulsion loss and boundary loss. The learning rate is 0.01. The curvature of the Poincar\'e ball is 1.0 and the $r$ is 0.76 which is used to alleviate the numerical issues \cite{guo2021free}. The hyperparameter $k$ is 1.55 which is shown to generate uniform particles well. For the assignment, we use a LeNet \cite{lecun1998gradient} for MNIST and a ResNet20 \cite{he2016deep} for CIFAR10 as the encoder. We apply HACK to each class separately. We attach a fully connected layer to project the feature into a two-dimensional Euclidean space. The image features are further projected onto hyperbolic space via an exponential map. We run the training for 200 epochs using a cosine learning rate scheduler \cite{loshchilov2016sgdr} with an initial learning rate of 0.1. We optimize the assignment \emph{every other} epoch. All the experiments are run on a NVIDIA TITAN RTX GPU.

\begin{figure}[t]
    \centering
    \includegraphics[width=\linewidth]{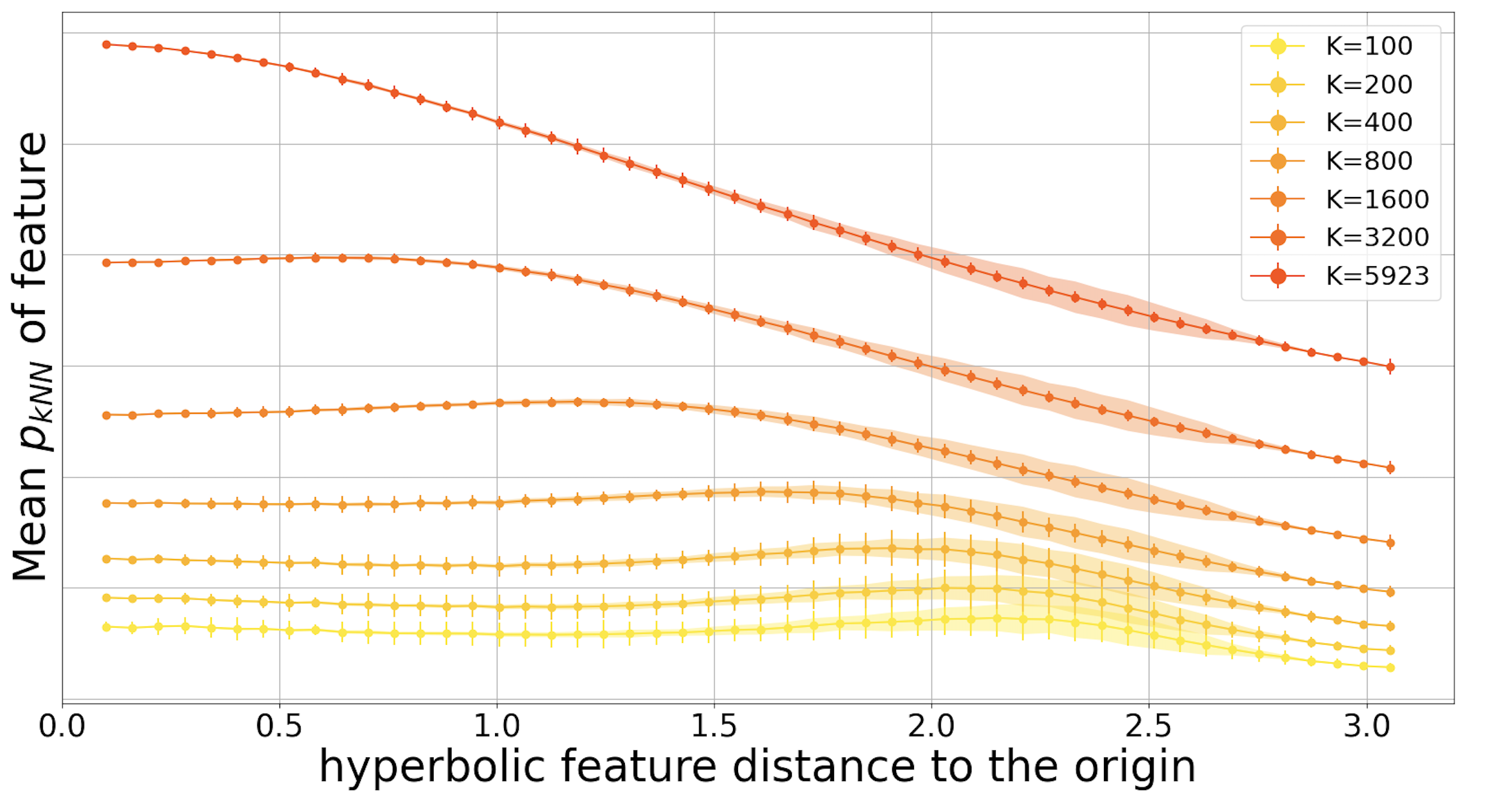}
    \caption{\textbf{Hyperbolic space can capture the \pt\ inherently.} The error bar of each point is given by the variance of density within the corresponding portion, and the width of the shaded band indicates the number of features within the portion.}
    \label{fig:norm}
\end{figure}

\subsection{Prototypicality in the Hyperbolic Feature Norm}

We explicitly show that the hyperbolic space can capture \pt\ by analyzing the relation between hyperbolic norms and the K-NN density estimation. Taken the learned hyperbolic features, we first divide the range of norms of hyperbolic features into numerous portions with equal length ($50$ portions for this plot). The mean K-NN density is calculated by averaging the density estimation of features within each portion. Figure \ref{fig:norm} shows that the mean density drops as the norm increases, which shows that the \pt\ emerges automatically within the norms, the inherent characteristic of hyperbolic space. This validates that prototypicality is reflected in the hyperbolic feature norm.

\subsection{Visual Prototypicality: Congealed MNIST}
We further apply HACK for visual feature learning on congealed MNIST. Figure \ref{fig:congealing_embeddings} shows that HACK can discover the congealed images from all images. In Figure \ref{fig:congealing_embeddings} a), the \textcolor{red}{red} particles denote the congealed images and \textcolor{cyan}{cyan} particles denote the original images. We can observe that the congealed images are assigned to the particles located in the center of the Poincar\'e ball. This verifies that HACK can \emph{indeed} discover prototypical examples from the original dataset. Section \ref{sec:with_c_and_o} in the Appendix shows that the features of atypical examples gradually move to the boundary of the Poincar\'e ball during training. In Figure \ref{fig:congealing_embeddings} b), we show the actual images that are embedded in the two-dimensional hyperbolic space. We can observe that the images in the center of Poincar\'e ball are more prototypical and images close to the boundary are more atypical. Also, the images are naturally organized by their semantic similarity. Figure \ref{fig:congealing_move} shows that the features of the original images become closer to the center of Poincar\'e ball after congealing. In summary, HACK can discover \pt\ and also organizes the images based on their semantics. To the best of our knowledge, this is the first unsupervised learning method that can be used to discover prototypical examples in a data-driven fashion.

\begin{figure}[!t]
    \centering
    \setlength{\tabcolsep}{5pt}
    \begin{tabular}{m{0.7cm}<{\centering}m{3.5cm}<{\centering}m{3.5cm}<{\centering}}
                   & Typical Images & Atypical Images \\
        HACK      & \includegraphics[width=0.2\textwidth]{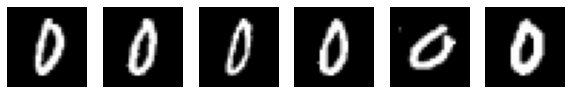} & \includegraphics[width=0.18\textwidth]{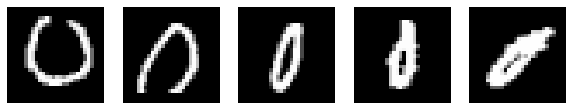} \\
        HR         & \includegraphics[width=0.2\textwidth]{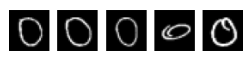} & \includegraphics[width=0.18\textwidth]{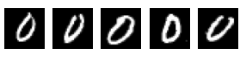} \\
        MC         & \includegraphics[width=0.2\textwidth]{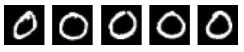} & \includegraphics[width=0.185\textwidth]{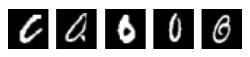} \\
        \begin{tabular}[c]{@{}c@{}}Mean \\ Image\end{tabular} & \multicolumn{1}{m{0.35cm}<{\centering}}{{\includegraphics[width=0.08\textwidth]{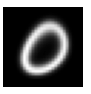}}}           
        \end{tabular}
    \caption{\textbf{HACK can discover both typical and atypical examples.} First column: typical images discovered by different methods. Second column: atypical images discovered by different methods.}
    \label{fig:comparison_metrics}
\end{figure}

\subsection{Prototypicality for Instance Selection}

\begin{figure}[!t]
   \centering
    \setlength{\tabcolsep}{0pt}
\begin{tabular}{cc}
   \includegraphics[width=0.25\textwidth]{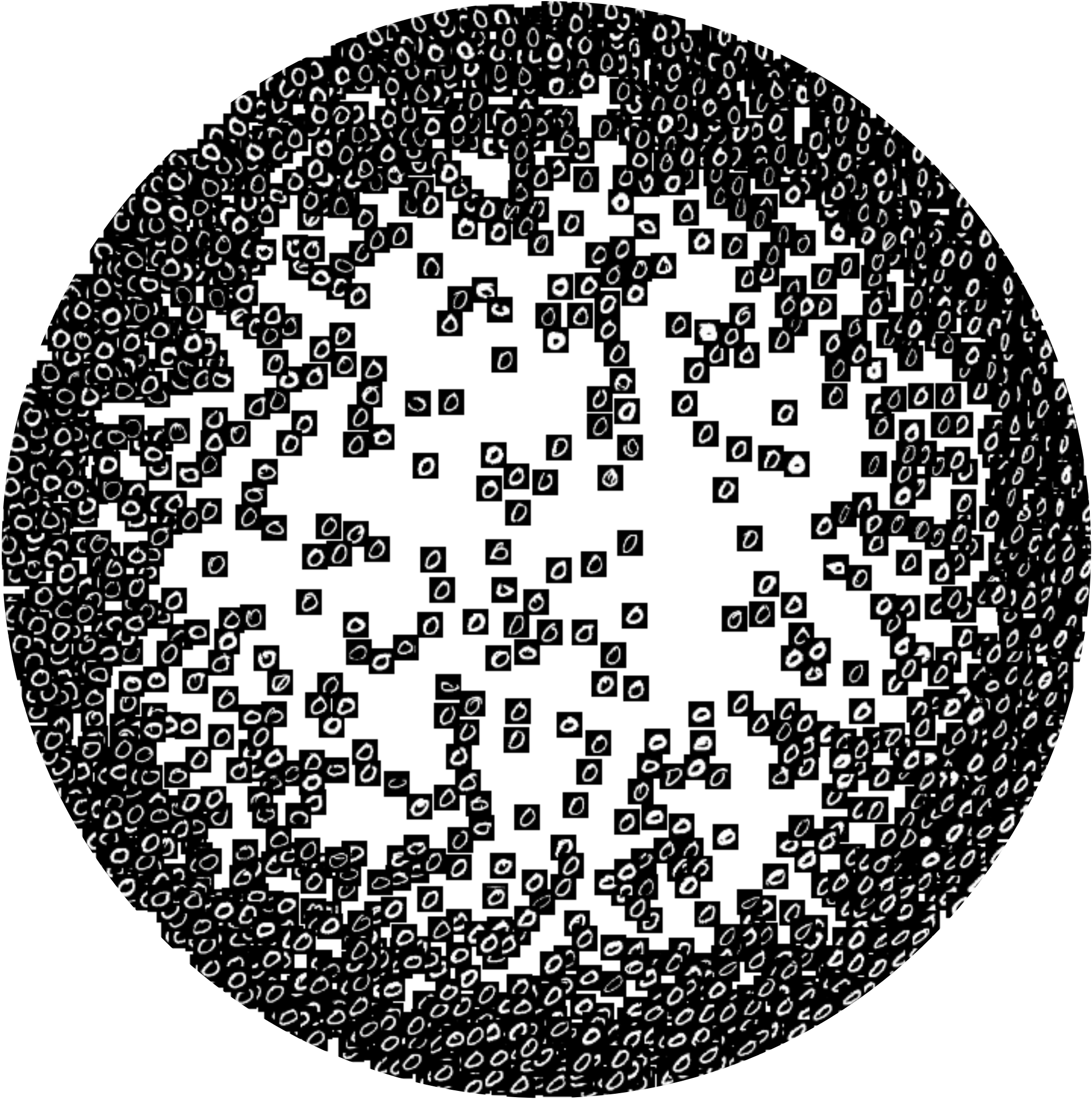} & \includegraphics[width=0.25\textwidth]{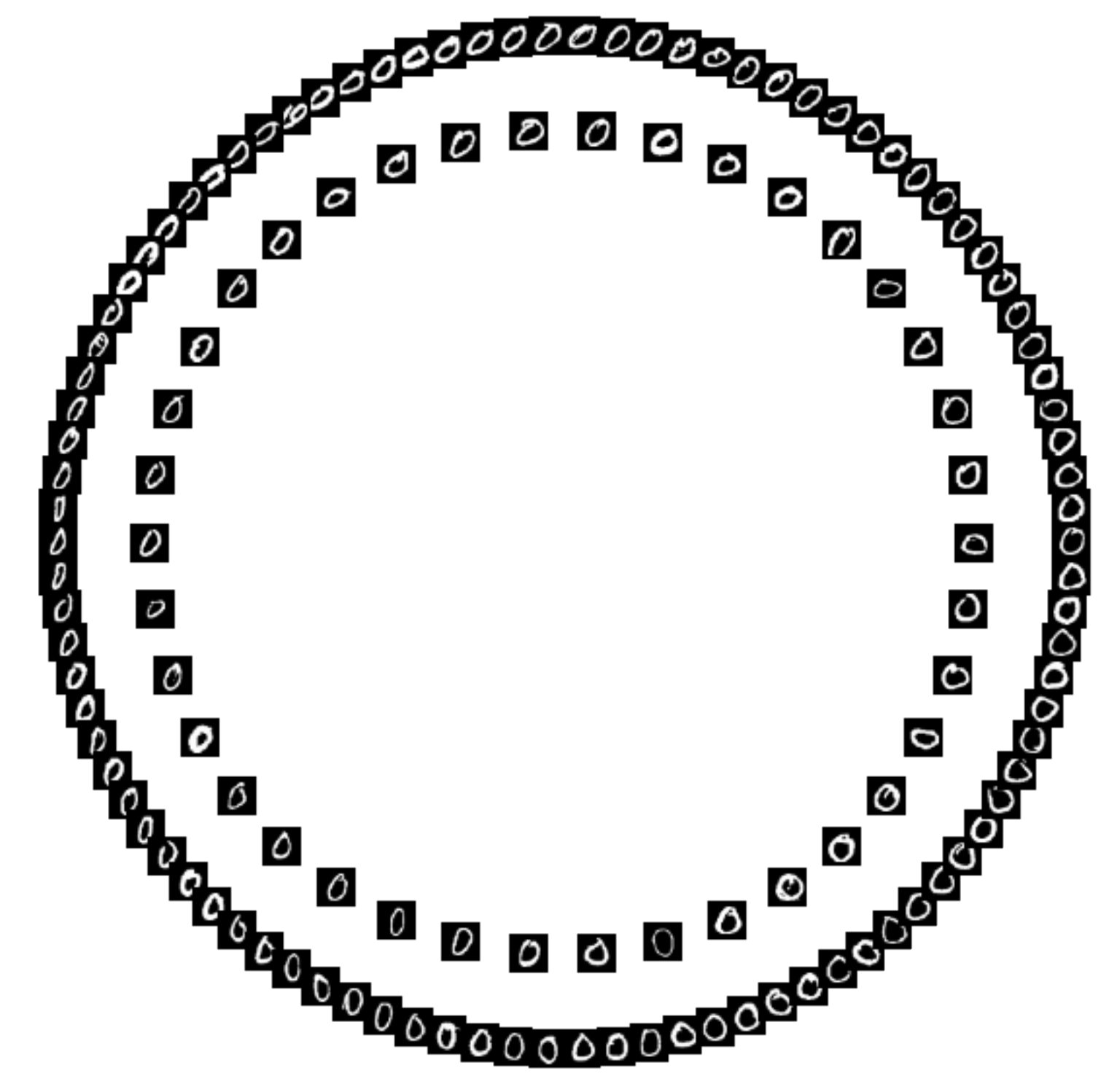} \\
   a) & b) \\
   \includegraphics[width=0.25\textwidth]{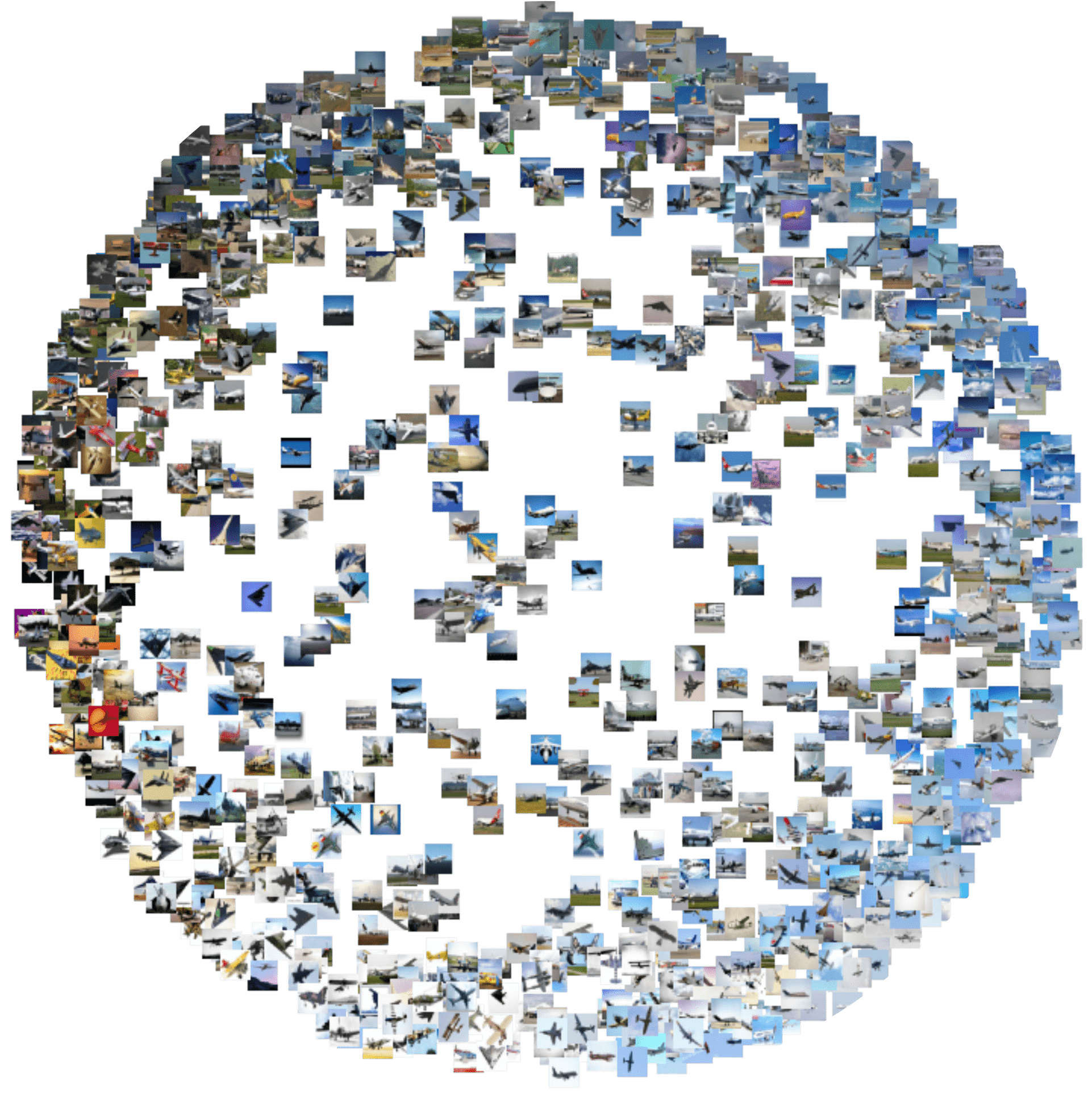} & \includegraphics[width=0.25\textwidth]{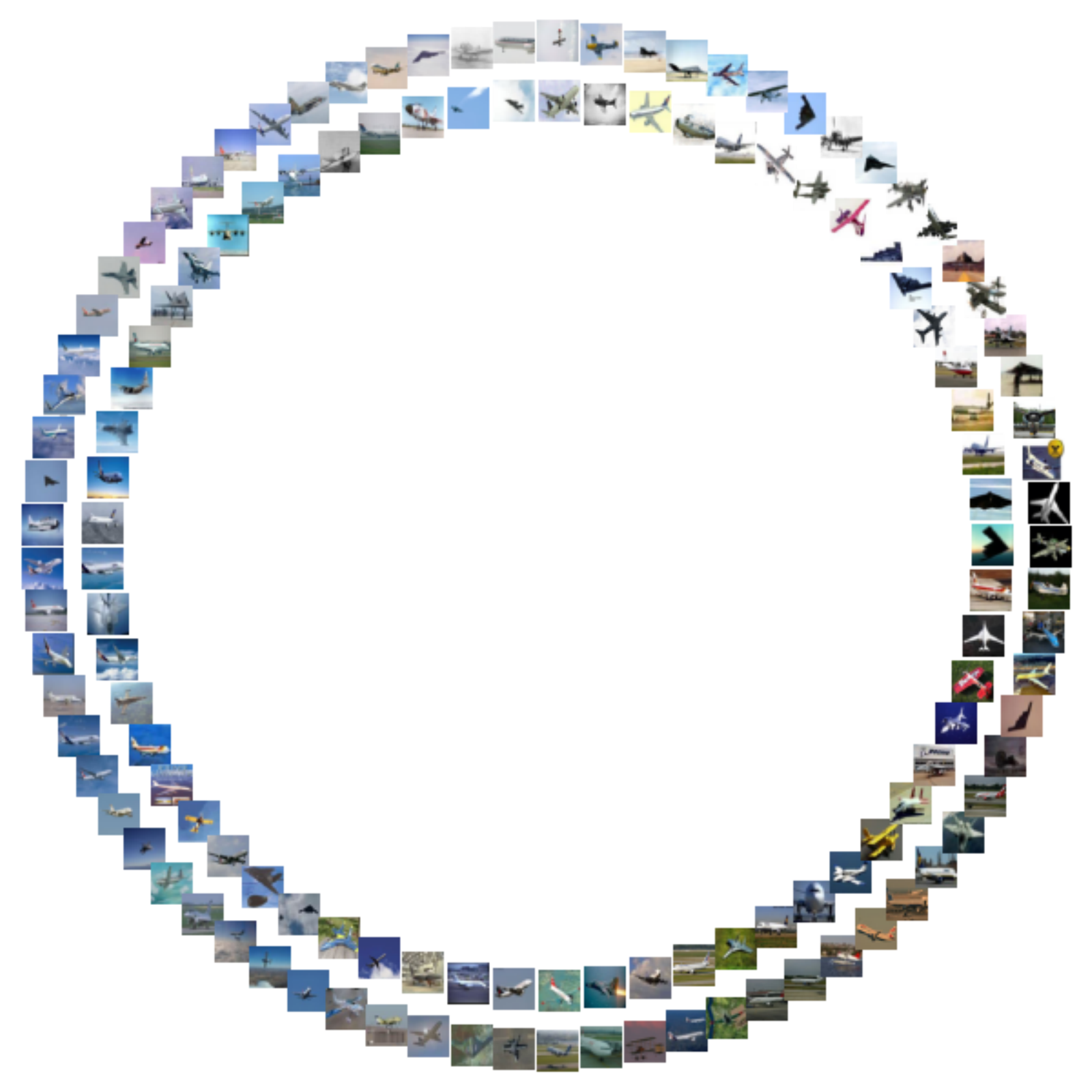} \\
    c) & d) 
\end{tabular}
    \caption{\textbf{Our unsupervised learning methods conforms to our visual perception}. a) Samples of 2000 images from MNIST. b) Images of MNIST arranged angularly. c) Samples of 2000 images from CIFAR10. d) Images of CIFAR10 arranged angularly. Images are organized based on \pt\ and visual similarity. }
        \label{fig:all_embeddings}
\end{figure}

Figure \ref{fig:all_embeddings} shows the embedding of class 0 from MNIST and class ``airplane" from CIFAR10 in the hyperbolic space. We sample 2000 images from MNIST and CIFAR10 for better visualization. We also show the arrangement of the images angularly with different angles. Radially, we can observe that images are arranged based on \pt. The prototypical images tend to locate in the center of the Poincar\'e ball. Especially for CIFAR10, the images become blurry and even unrecognizable as we move toward the boundary of the ball. Angularly, the images are arranged based on visual similarity. The visual similarity of images has a smooth transition as we move around angularly. Please see Section \ref{sec:more_cifar} in the Appendix for more results.\\

\noindent \textbf{Comparison with Baselines.} Figure \ref{fig:comparison_metrics} shows the comparison of the baselines with HACK. We can observe that both HACK and Model Confidence (MC) can discover typical and atypical images. Compared with MC, HACK defines \pt\ as the distance of the sample to other samples which is more aligned with human intuition. Moreover, in addition to \pt, HACK can also be used to organize examples by semantic similarities. Holdout Retraining (HR) is not effective for \pt\ discovery due to the randomness of model training.

\subsection{Application of Prototypicality}

\noindent \textbf{Reducing Sample Complexity.} The proposed HACK can discover prototypical images as well as atypical images. We show that with \emph{atypical} images we can reduce the sample complexity for training the model. Prototypical images are representative of the dataset but lack variations. Atypical examples contain more variations and it is intuitive that models trained on atypical examples should generalize better to the test samples. To verify this hypothesis, we select a subset of samples based on the norm of the features which indicates \pt\ of the examples. We consider using both the most typical and atypical examples for training the model. In particular, typical samples correspond to the samples with smaller norms and atypical samples correspond to the samples with larger norms. The angular layout of the hyperbolic features naturally captures sample diversity, thus for selecting atypical examples, we also consider introducing more diversity by sampling images with large norms along the angular direction. 

\begin{figure}[!t]
   \centering
    \setlength{\tabcolsep}{0pt}
\begin{tabular}{cc}
    \includegraphics[width=0.5\linewidth]{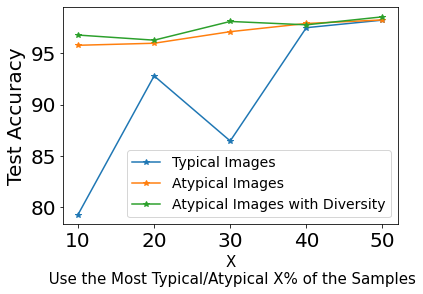} &     \includegraphics[width=0.5\linewidth]{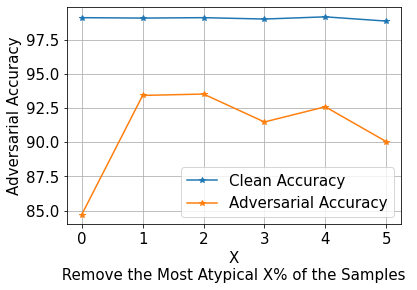} \\
    a) & b)
\end{tabular}
    \caption{a) Training with atypical examples achieves higher accuracy than training with typical examples. b) The adversarial accuracy greatly improves after removing the X\% of most atypical examples.}
        \label{fig:typical_classification}
\end{figure}

We train a LeNet on MNIST for 10 epochs with a learning rate of 0.1. Figure \ref{fig:typical_classification} a) shows that training with atypical images can achieve much higher accuracy than training with typical images. In particular, training with the most atypical 10\% of the images achieves 16.54\% higher accuracy than with the most typical 10\% of the images. Thus, HACK provides an easy solution to reduce sample complexity. The results further verify that HACK can distinguish between prototypical and atypical examples.\\

\noindent \textbf{Increasing Model Robustness.} Training models with atypical examples can lead to a vulnerable model to adversarial attacks \cite{liu2018less,carlini2018prototypical}. Intuitively, atypical examples lead to a less smooth decision boundary thus a small perturbation to examples is likely to change the prediction. With HACK, we can easily identify atypical samples to improve the robustness of the model. We use MNIST as the benchmark and use FGSM \cite{goodfellow2014explaining} to attack the model with an $\epsilon = 0.07$. We identify the atypical examples with HACK and remove the most atypical X\% of the examples. Figure \ref{fig:typical_classification} b) shows that discarding atypically examples greatly improves the robustness of the model: the adversarial accuracy is improved from  84.72\% to 93.42\% by discarding the most atypical 1\% of the examples. It is worth noting that the clean accuracy remains the same after removing a small number of atypical examples.

\section{Summary}
We propose an unsupervised learning method, called HACK, for organizing images with sphere packing in hyperbolic space. HACK optimizes the assignments of the images to a fixed set of uniformly distributed particles by naturally exploring the properties of hyperbolic space. As a result, prototypical and semantic structures emerge naturally due to the feature learning. We apply HACK to synthetic data with known \pt\ and standard image datasets. The discovered \pt\ and atypical examples can be used to reduce sample complexity and increase model robustness. The idea of HACK can also be generalized to learn other geometrical structures from the data by specifying different geometric patterns.

{\small
\bibliographystyle{ieee_fullname}
\bibliography{egbib}
}

\clearpage

\appendix

\section{More Details on K-NN Density Estimation on MNIST}
\label{sec:moco}

\noindent \textbf{Feature Extraction:} We use a LeNet \cite{lecun1998gradient} without classifier as the encoder and follow the scheme of MoCo~\cite{He2019MomentumCF} to train the feature extractor. We run the training for 200 epochs and the initial learning rate is 0.06. We use a cosine learning rate scheduler \cite{loshchilov2016sgdr}.\\

\begin{figure}[!t]
    \centering
    \includegraphics[width=1\linewidth]{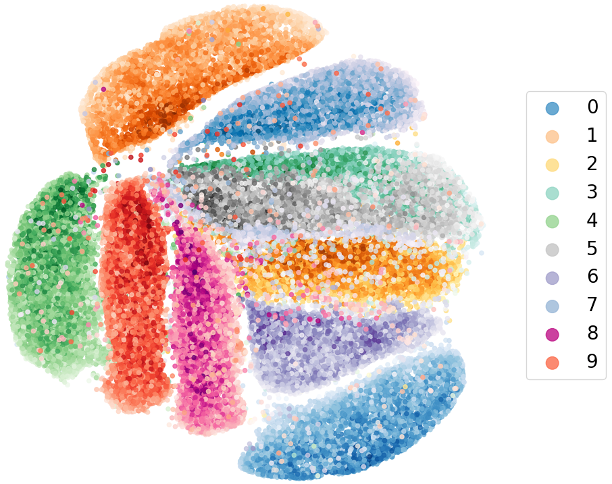}
    \caption{The KNN density estimation on MoCo~\cite{He2019MomentumCF} features of MNIST~\cite{lecun1998mnist}. The shades of color represent the density value: the darker the color, the higher the density.}
        \label{fig:knn_moco}
\end{figure}

\noindent \textbf{Visualization:} Figure \ref{fig:knn_moco} visualize the KNN density estimation on MoCo \cite{He2019MomentumCF} features of MNIST \cite{lecun1998mnist}. The output features have the dimension of $64$. To visualize the features, we use t-SNE \cite{van2008visualizing} with the perplexity of $40$ and $300$ iterations for optimization.

\section{More Details on Hyperbolic Instance Assignment}

A more detailed description of the hyperbolic instance assignment is given. 

\begin{figure*}[!t]
    \centering
    \begin{tabular}{ cc}
\includegraphics[width=0.4\textwidth]{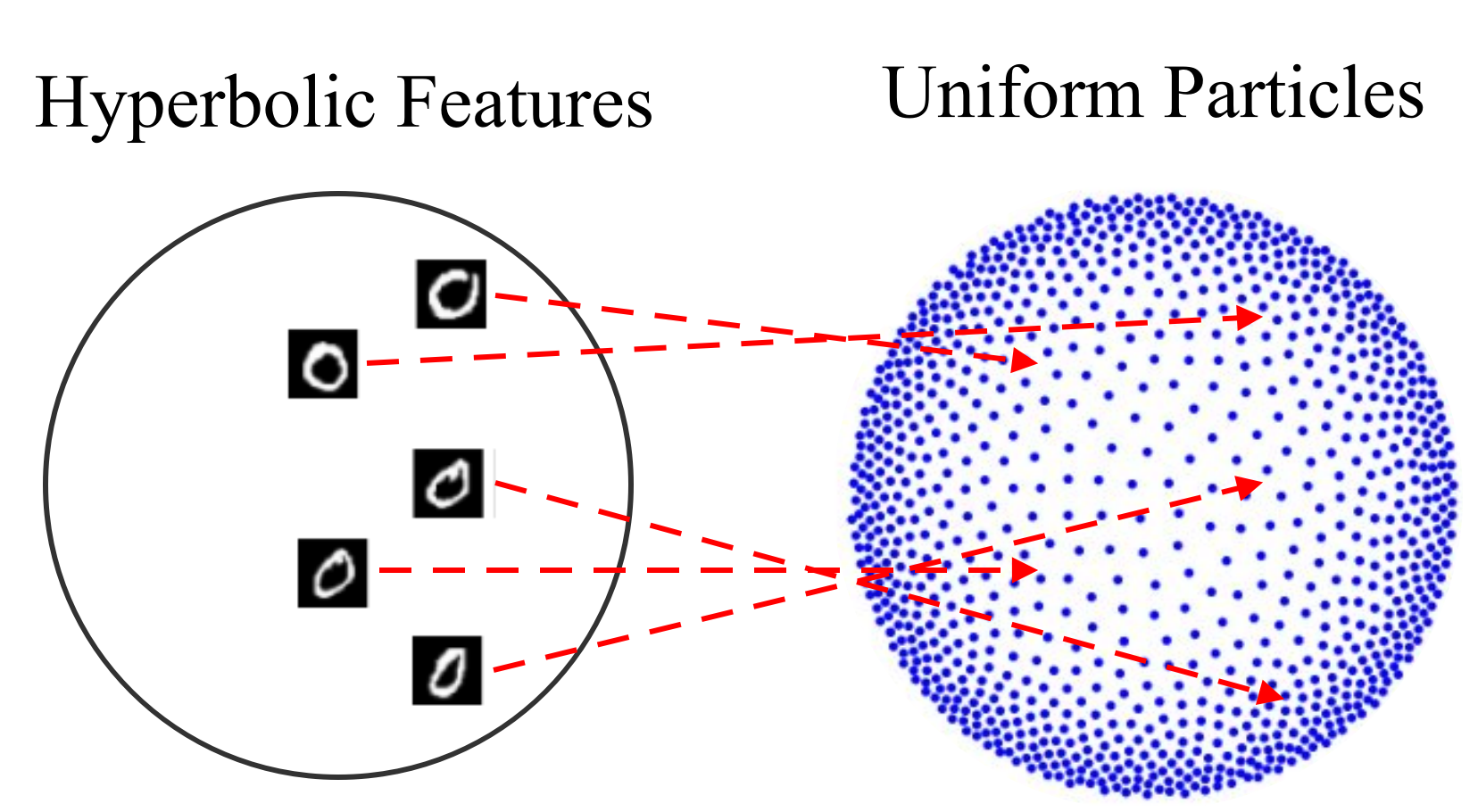}  & \includegraphics[width=0.4\textwidth]{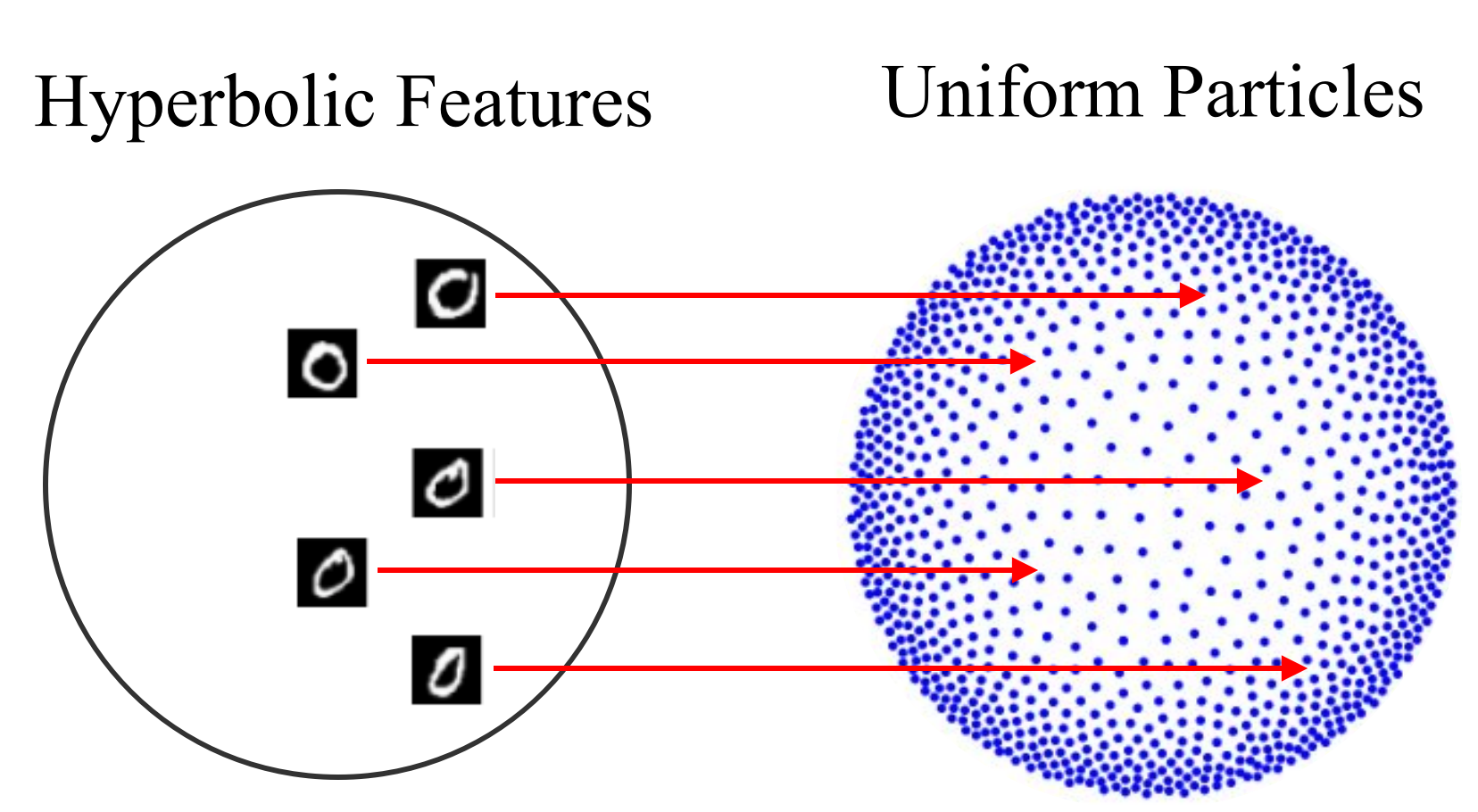}  \\
a)  & b)
    \end{tabular}
    \caption{\textbf{Hyperbolic Instance Assignment minimizes the total hyperbolic distances between the image features and the particles.} a) Initial assignment. b) Optimized assignment.}
    \label{fig:opt_assign}
\end{figure*}

Initially, we randomly assign the particles to the images. Given a batch of samples $\{(\mathbf{x}_1, s_1), (\mathbf{x}_2, s_2), ..., (\mathbf{x}_b, s_b)\}$, where $\mathbf{x}_i$ is an image and $s_i$ is the corresponding particle. Given an encoder $f_\theta$, we generate the hyperbolic feature for each image $\mathbf{x}_i$ as $f_\theta(\mathbf{x}_i) \in \mathbb{B}^2$, where $\mathbb{B}^2$ is a two-dimensional Poincar\'e ball.

we aim to find the minimum cost bipartite matching of the images to the particles. The cost to minimize is the total hyperbolic distance of the hyperbolic features to the particles. We first compute all the pairwise distances between the hyperbolic features and the particles. This is the cost matrix of the bipartite graph. Then we use the Hungarian algorithm to optimize the assignment (Figure \ref{fig:opt_assign}).

Suppose we train the encoder $f_\theta$ for $T$ epochs. We run the hyperbolic instance assignment every other epoch to avoid instability during training. \textbf{We optimize the encoder $f_\theta$ to minimize the hyperbolic distance of the hyperbolic feature to the assigned particle in each batch}.

\section{Details of Baselines}
\label{sec:baseline}

\noindent \textbf{Holdout Retraining:} We consider the Holdout Retraining proposed in \cite{carlini2018prototypical}. The idea is that the distance of features of prototypical examples obtained from models trained on different datasets should be close. In Holdout Retraining, multiple models are trained on the same dataset. The distances of the features of the images obtained from different models are computed and ranked. The prototypical examples are those examples with the closest feature distance.

\noindent \textbf{Model Confidence:} Intuitively, the model should be confident on prototypical examples. Thus, it is natural to use the confidence of the model prediction as the criterion for \pt. Once we train a model on the dataset, we use the confidence of the model to rank the examples. The  prototypical examples are those examples that the model is most 

\section{More Results on Prototypicality Discovery}
\label{sec:more_cifar}
\begin{figure*}[!ht]
    \centering
    \begin{tabular}{ cc}
\includegraphics[width=0.45\textwidth]{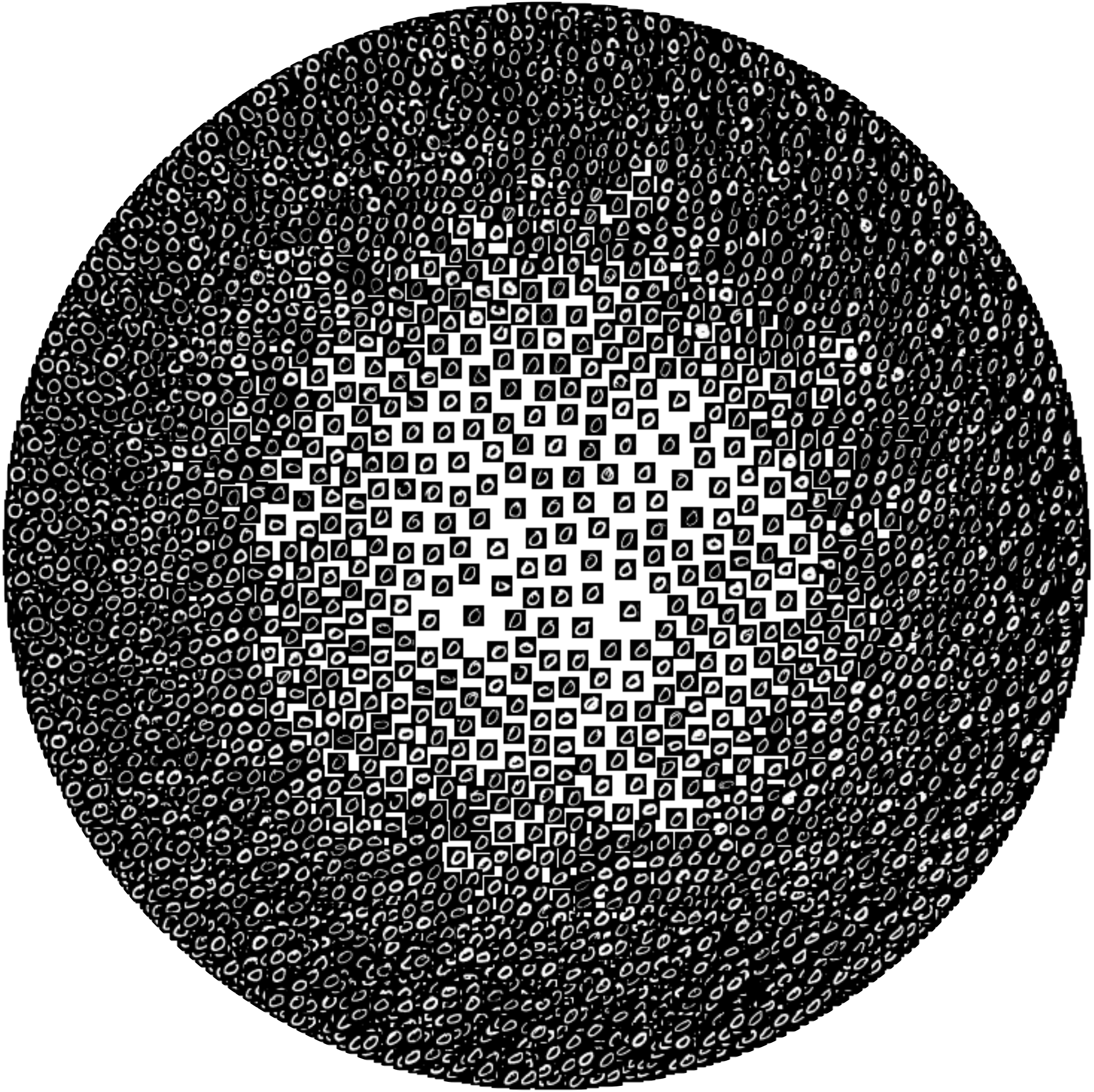}  & \includegraphics[width=0.45\textwidth]{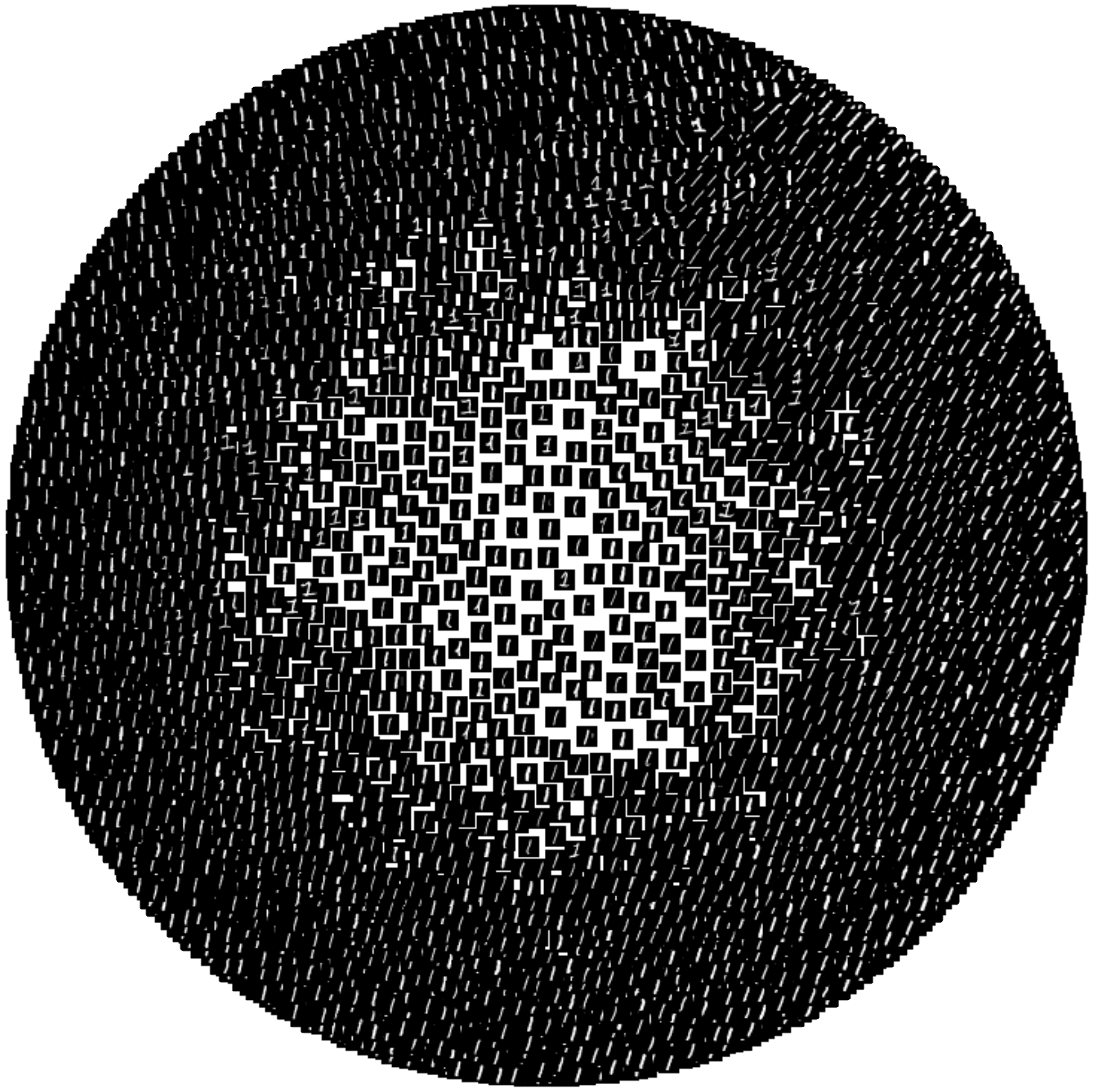}  \\
a)  & b) \\
\includegraphics[width=0.45\textwidth]{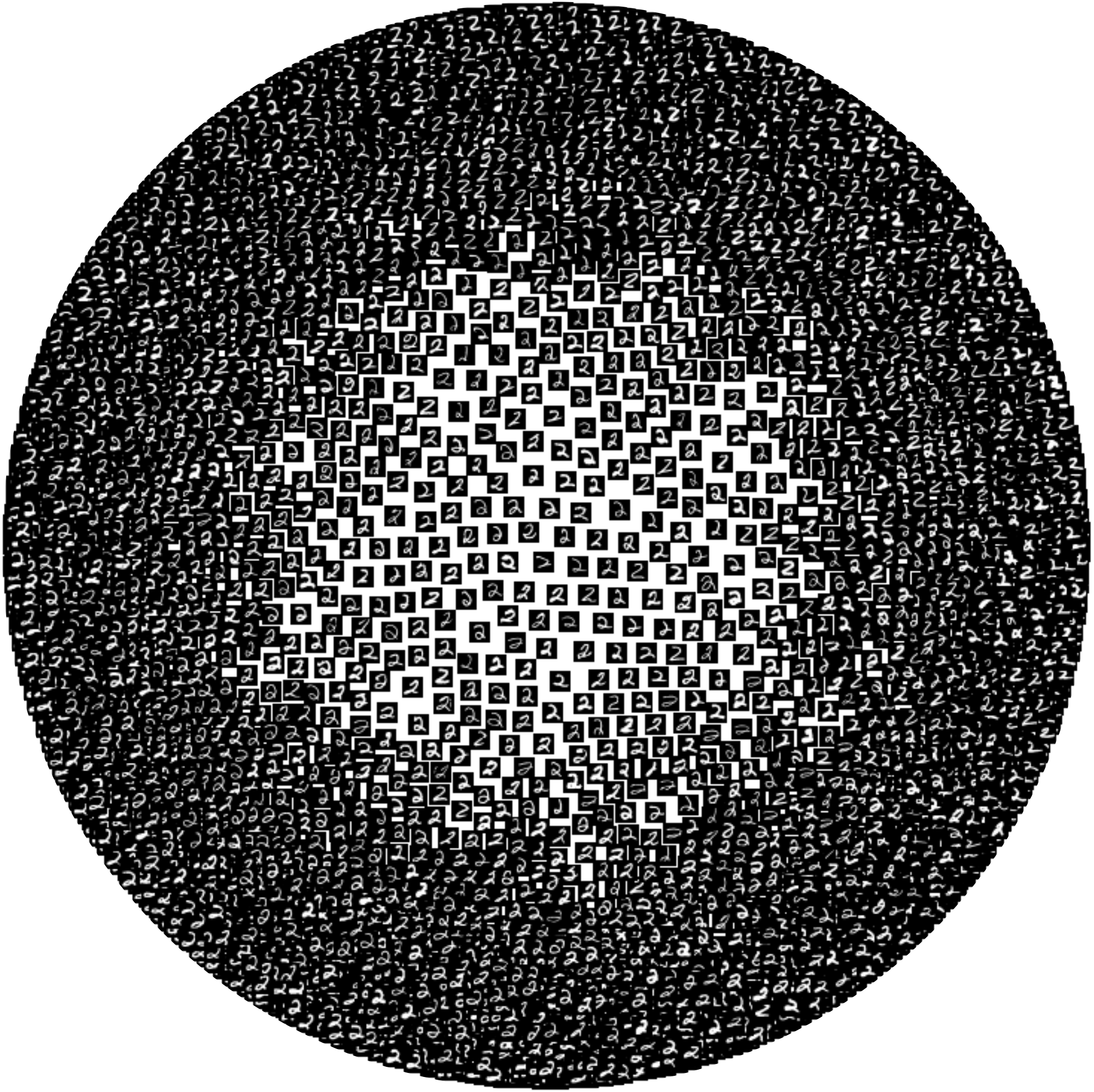}  & \includegraphics[width=0.45\textwidth]{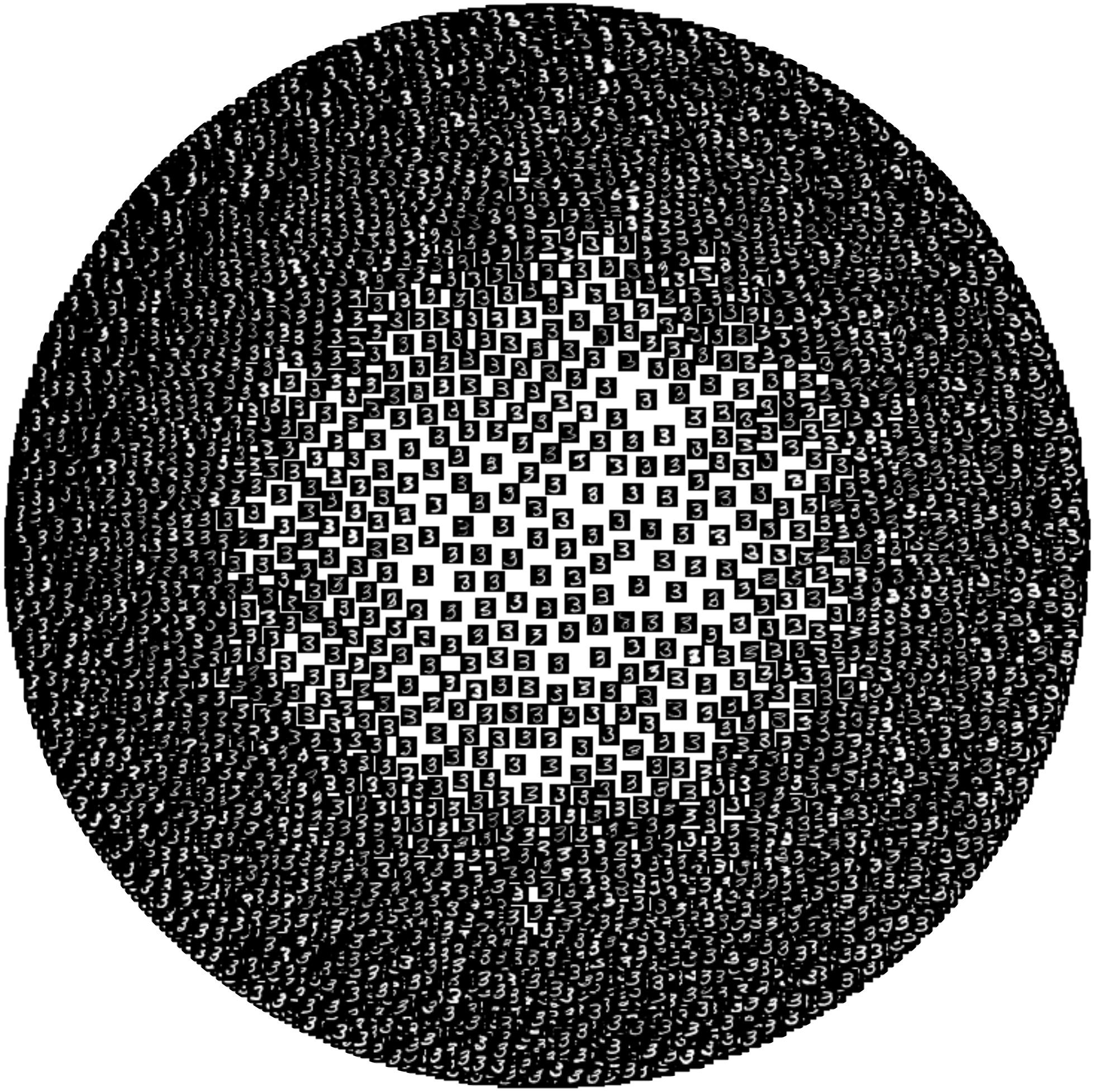}  \\
c)  & d)
    \end{tabular}
    \caption{\textbf{HACK captures \pt\ and semantic similarity on MNIST.} a) Class 0. b) Class 1. c) Class 2. d) Class 3. }
    \label{fig:mnist_all}
\end{figure*}

\begin{figure*}[!ht]
    \centering
    \begin{tabular}{ cc}
\includegraphics[width=0.4\textwidth]{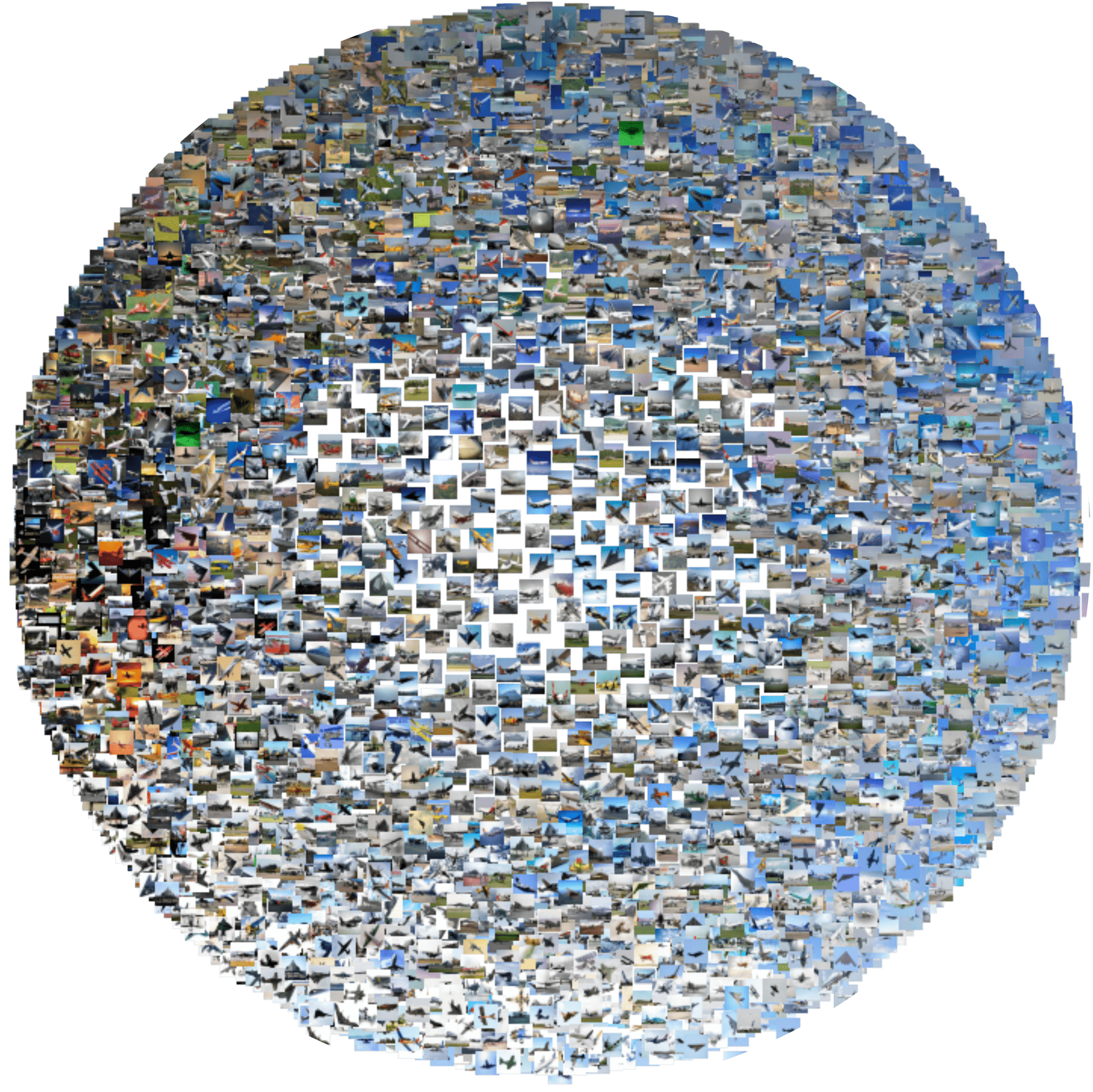}  & \includegraphics[width=0.4\textwidth]{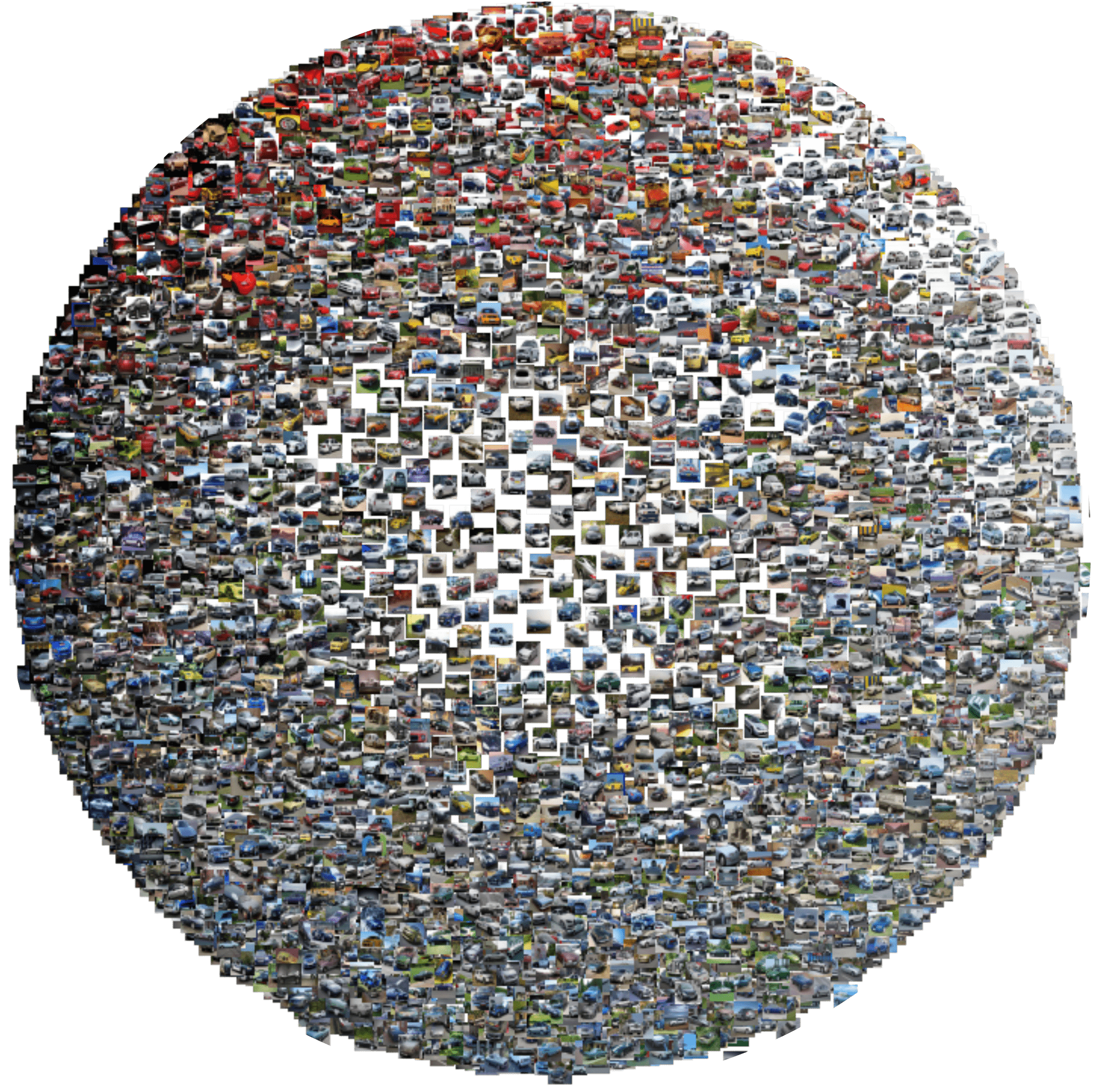}  \\
a)  & b) \\
\includegraphics[width=0.4\textwidth]{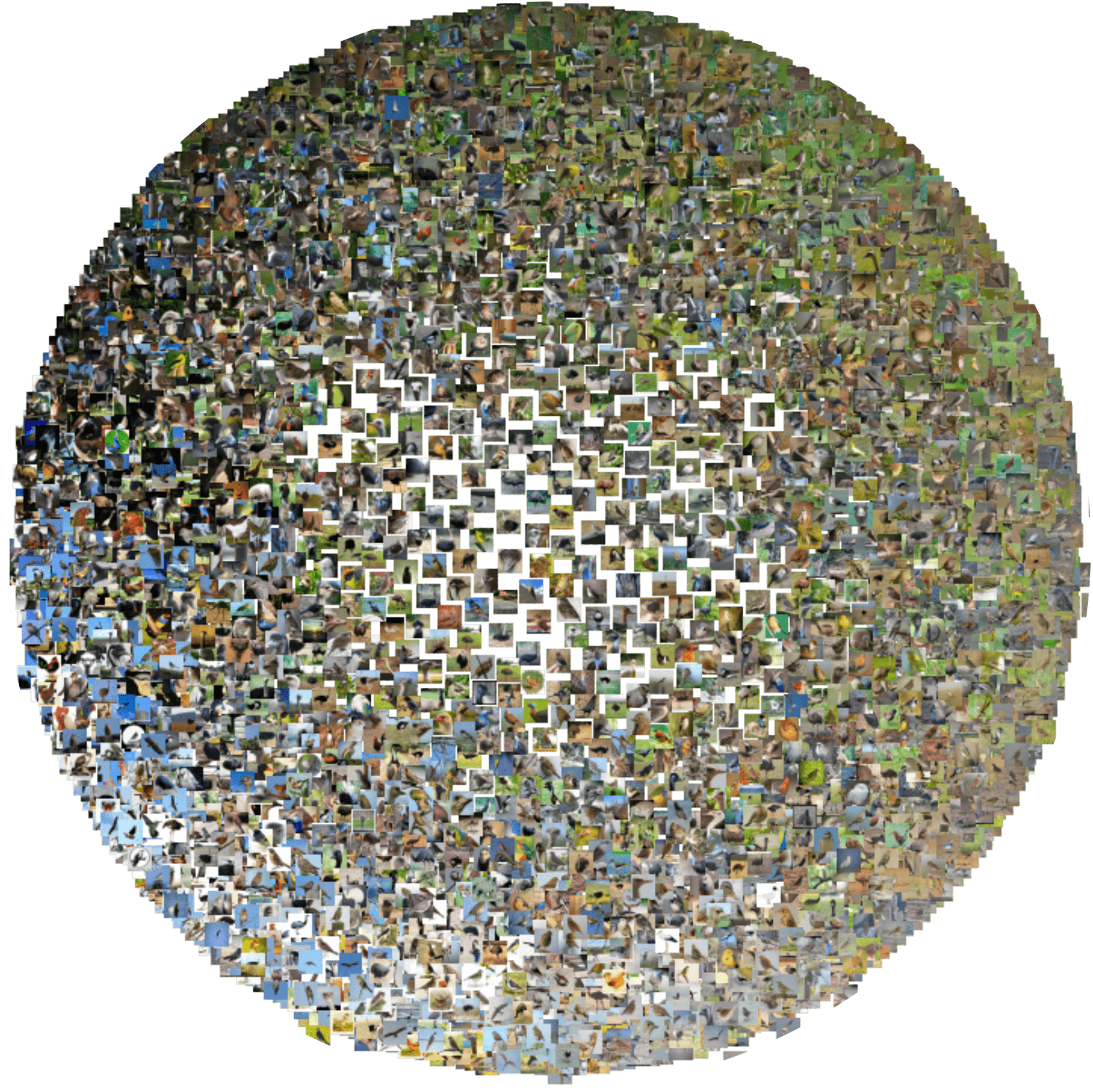}  & \includegraphics[width=0.4\textwidth]{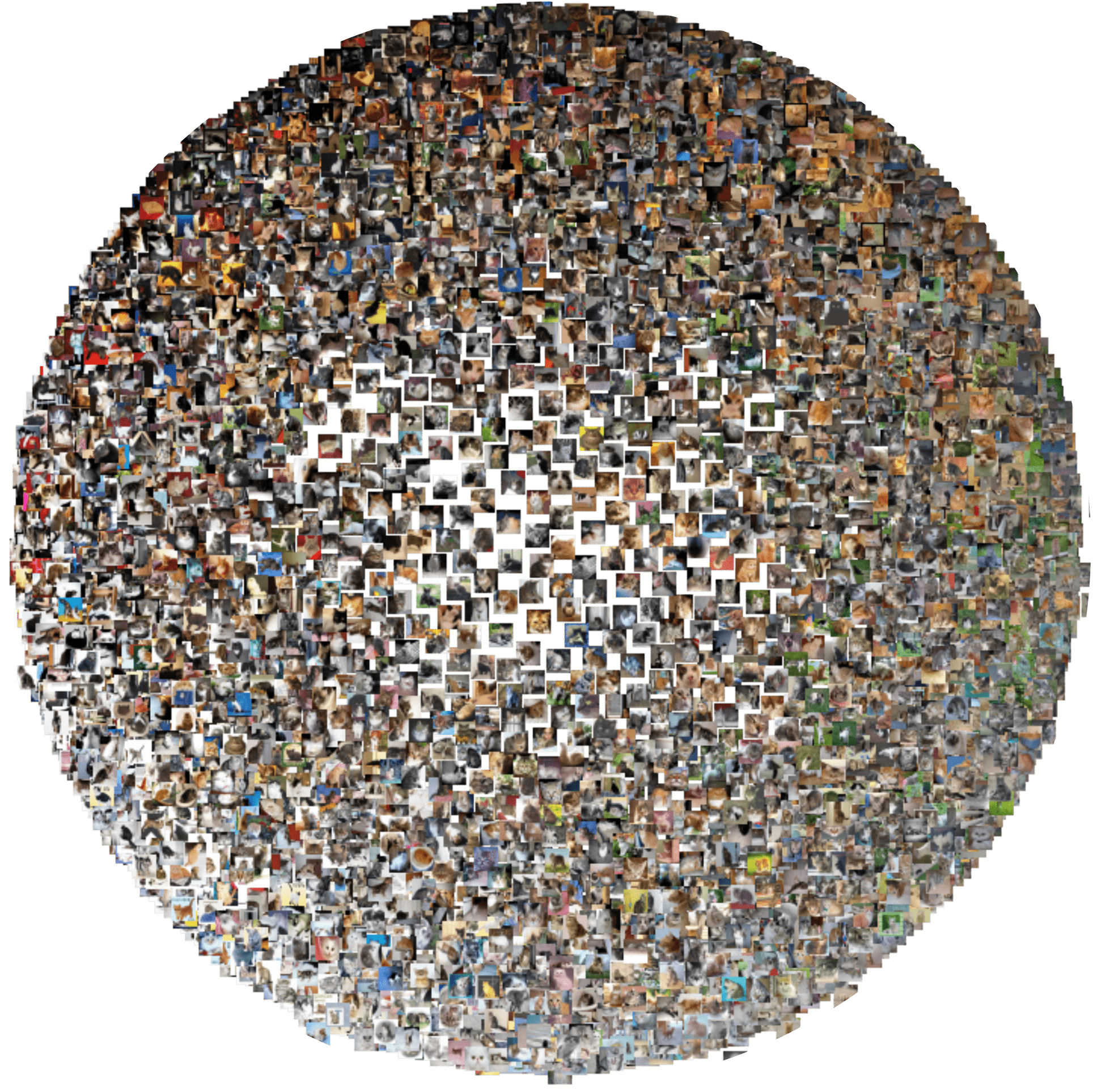}  \\
c)  & d)
    \end{tabular}
    \caption{\textbf{HACK captures \pt\ and semantic similarity on CIFAR10.} a) Class ``airplane". b) Class ``automobile". c) Class ``bird". d) Class ``cat".}
    \label{fig:cifar_all}
\end{figure*}

\begin{figure*}[!t]
    \begin{tabular}{ m{0.2\linewidth} | m{0.7\linewidth} }
   Typical Images: & \includegraphics[width=0.7\textwidth]{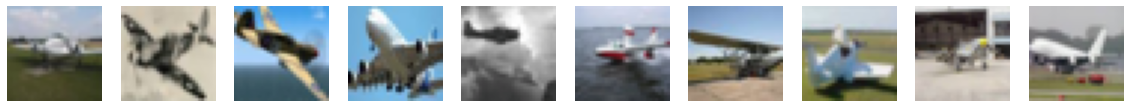}  \\
  Atypical Images: &     \includegraphics[width=0.7\textwidth]{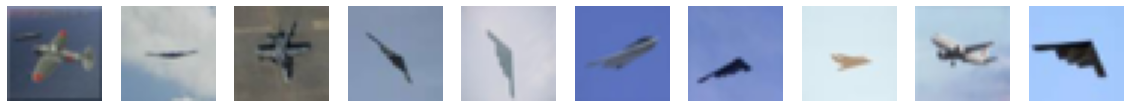} \\
\midrule \\
     Typical Images: & \includegraphics[width=0.7\textwidth]{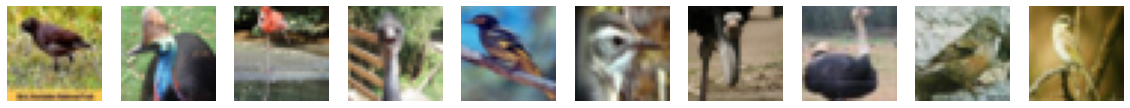}  \\
  Atypical Images: &     \includegraphics[width=0.7\textwidth]{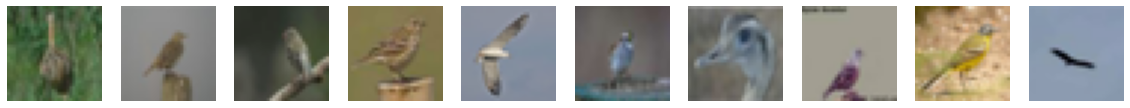}  \\
        \end{tabular}
    \caption{\textbf{Most typical and atypical images extracted by HACK from CIFAR10.}}
    \label{fig:cifar_retrieval}
\end{figure*}

We show the visualization of all the images in Figure \ref{fig:mnist_all} and Figure \ref{fig:cifar_all}. The images are organized naturally based on their \pt\ and semantic similarity. We further conduct retrieval based on the norm of the hyperbolic features to extract the most typical and atypical images on CIAFR10 in Figure \ref{fig:cifar_retrieval}. The hyperbolic features with large norms correspond to atypical images and the hyperbolic features with small norms correspond to typical images. It can be observed that the object in the atypical images is not visible. 

\section{Gradually Adding More Congealed Images}
\label{sec:gradual}
\begin{figure*}[!ht]
    \centering
\setlength{\tabcolsep}{0pt}
    \begin{tabular}{ccccc}
\includegraphics[width=0.2\textwidth]{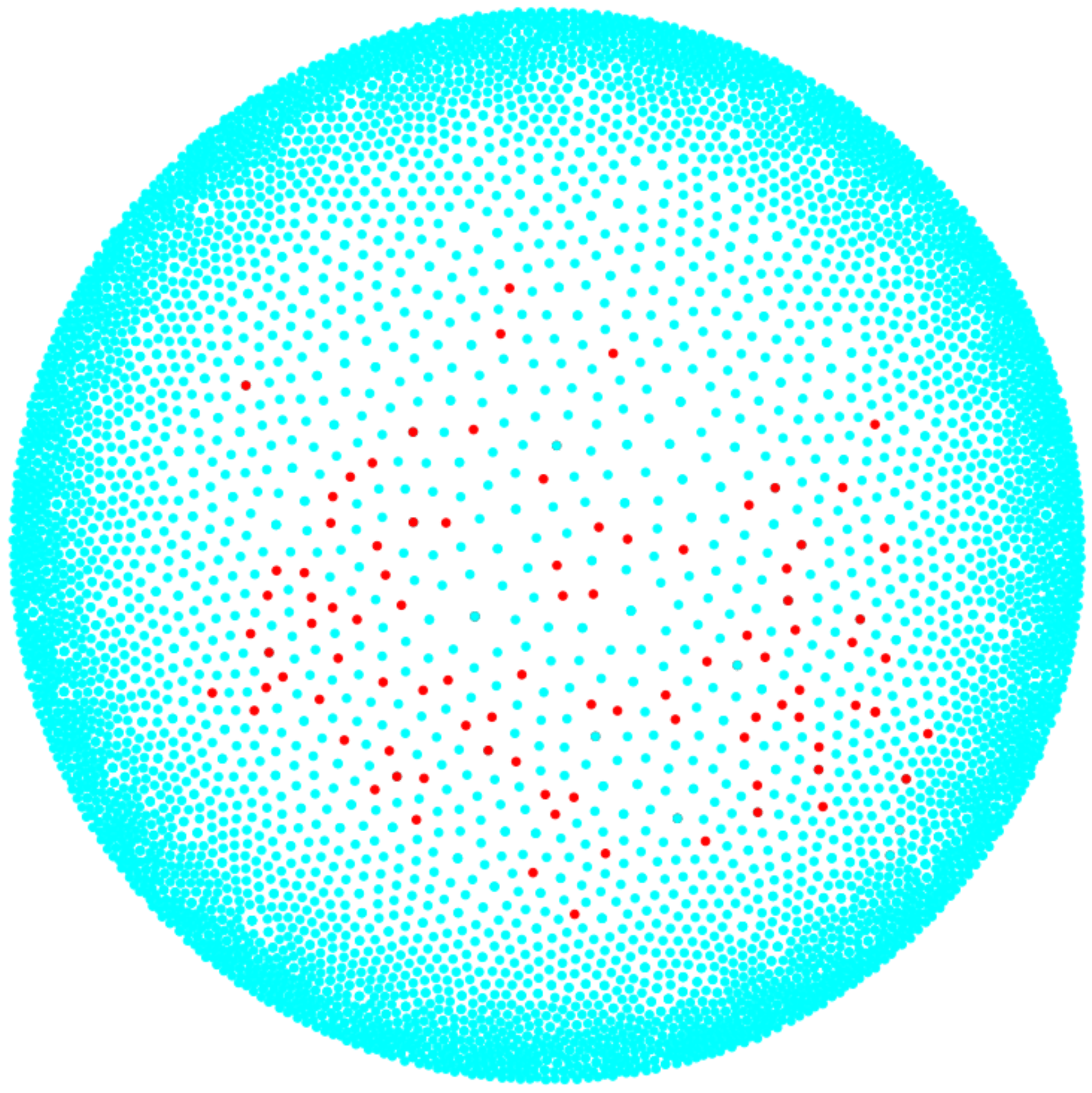} & \includegraphics[width=0.2\textwidth]{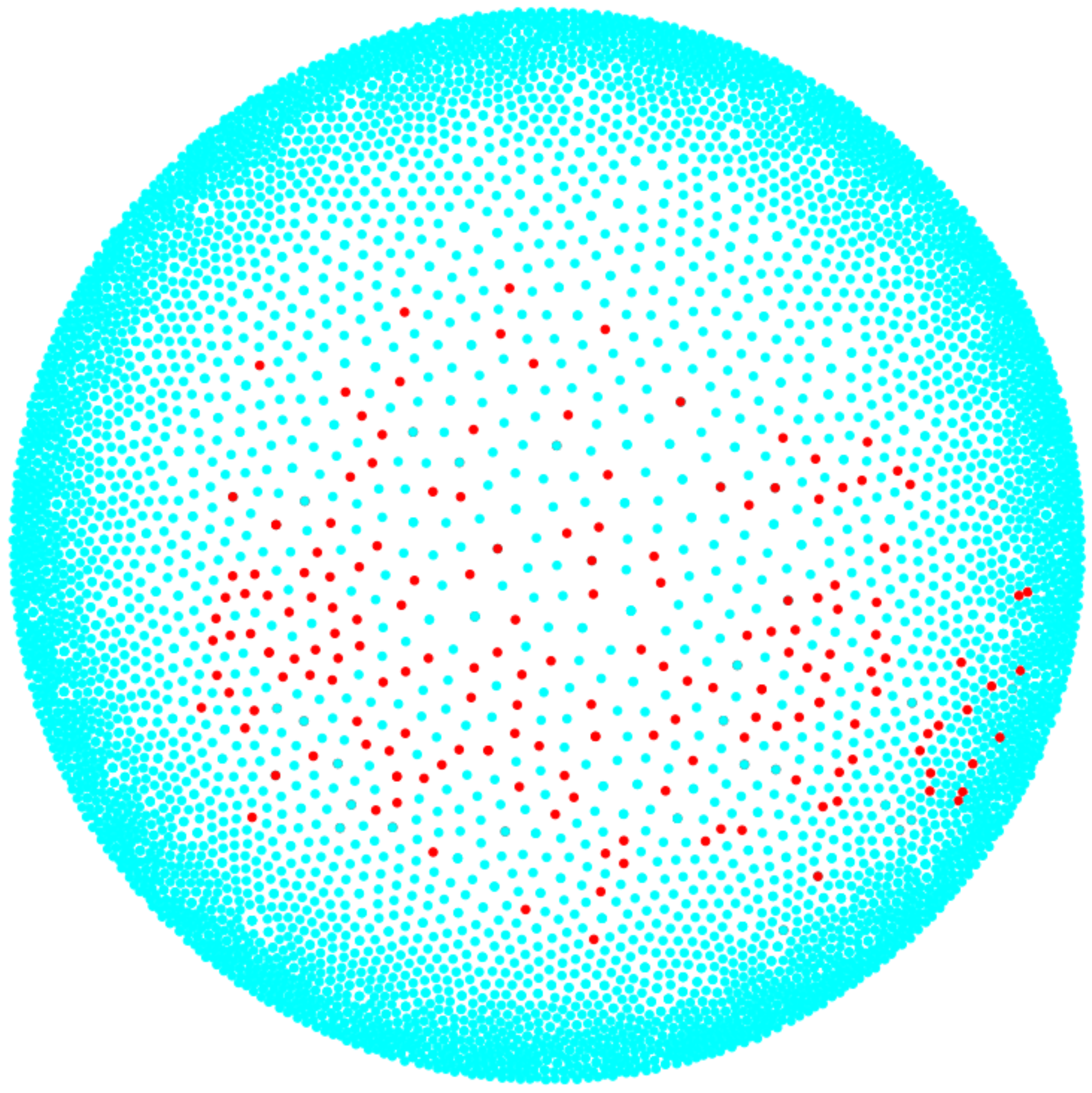} & \includegraphics[width=0.2\textwidth]{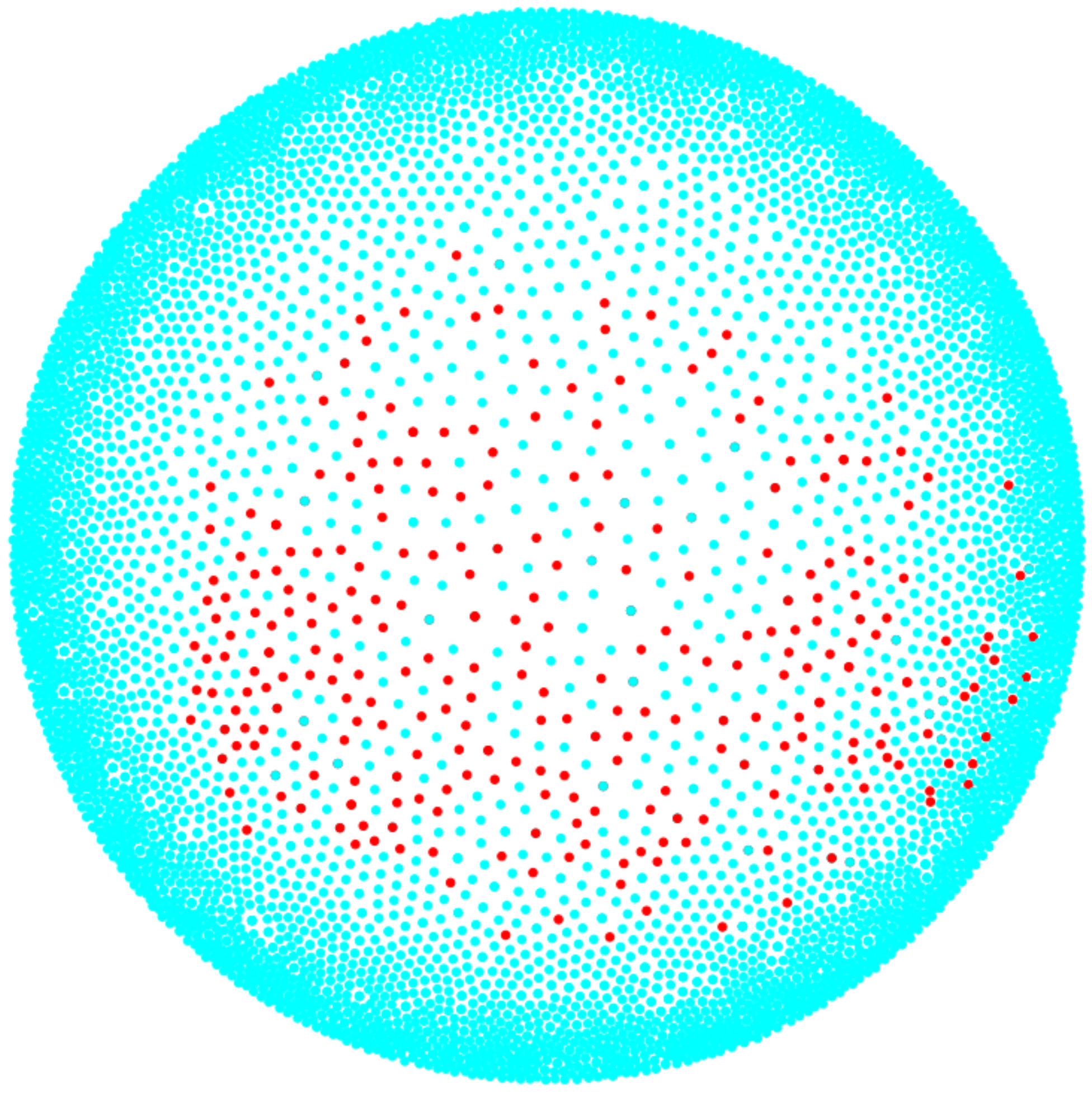}  & 
\includegraphics[width=0.2\textwidth]{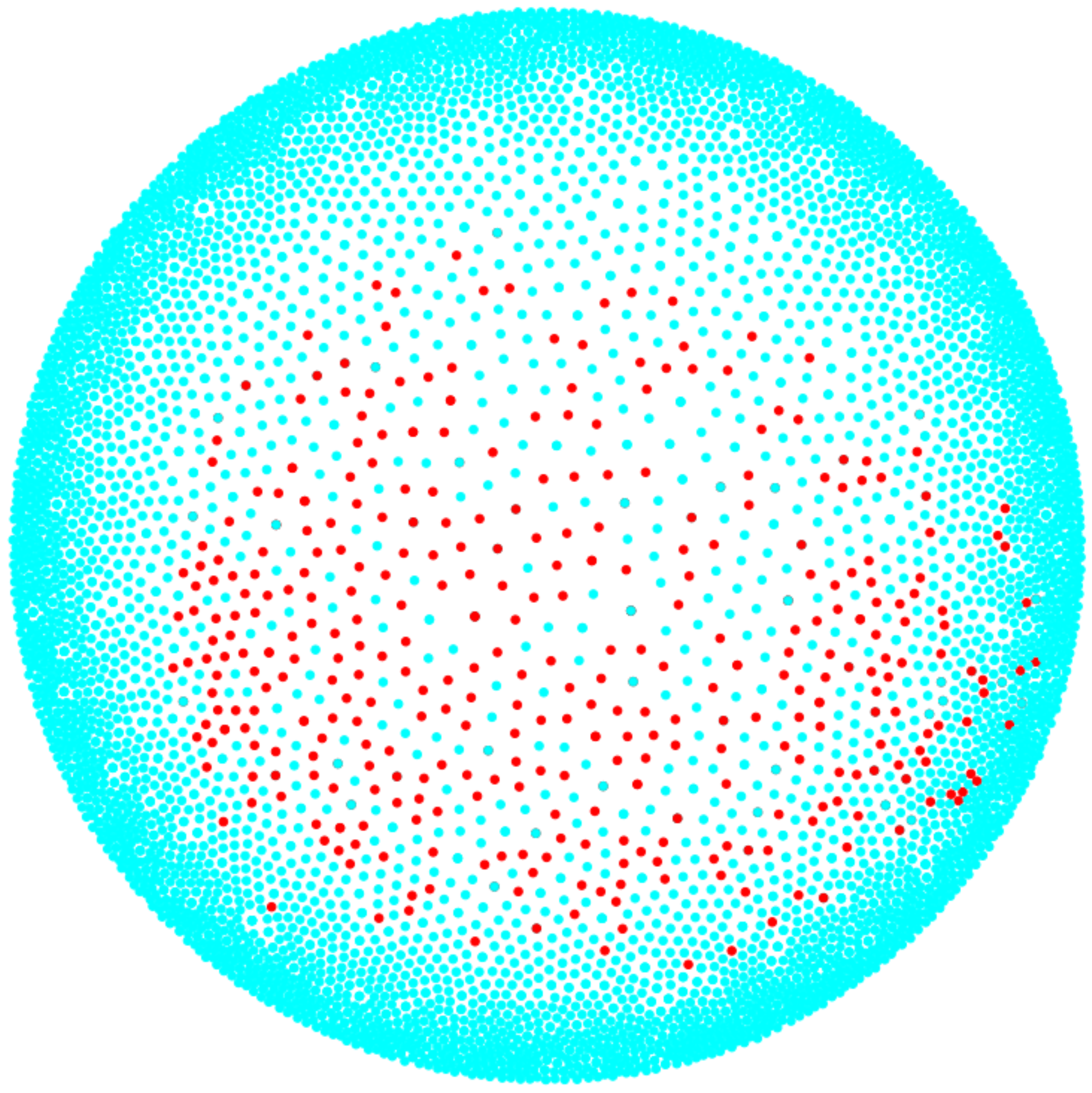}  &
\includegraphics[width=0.2\textwidth]{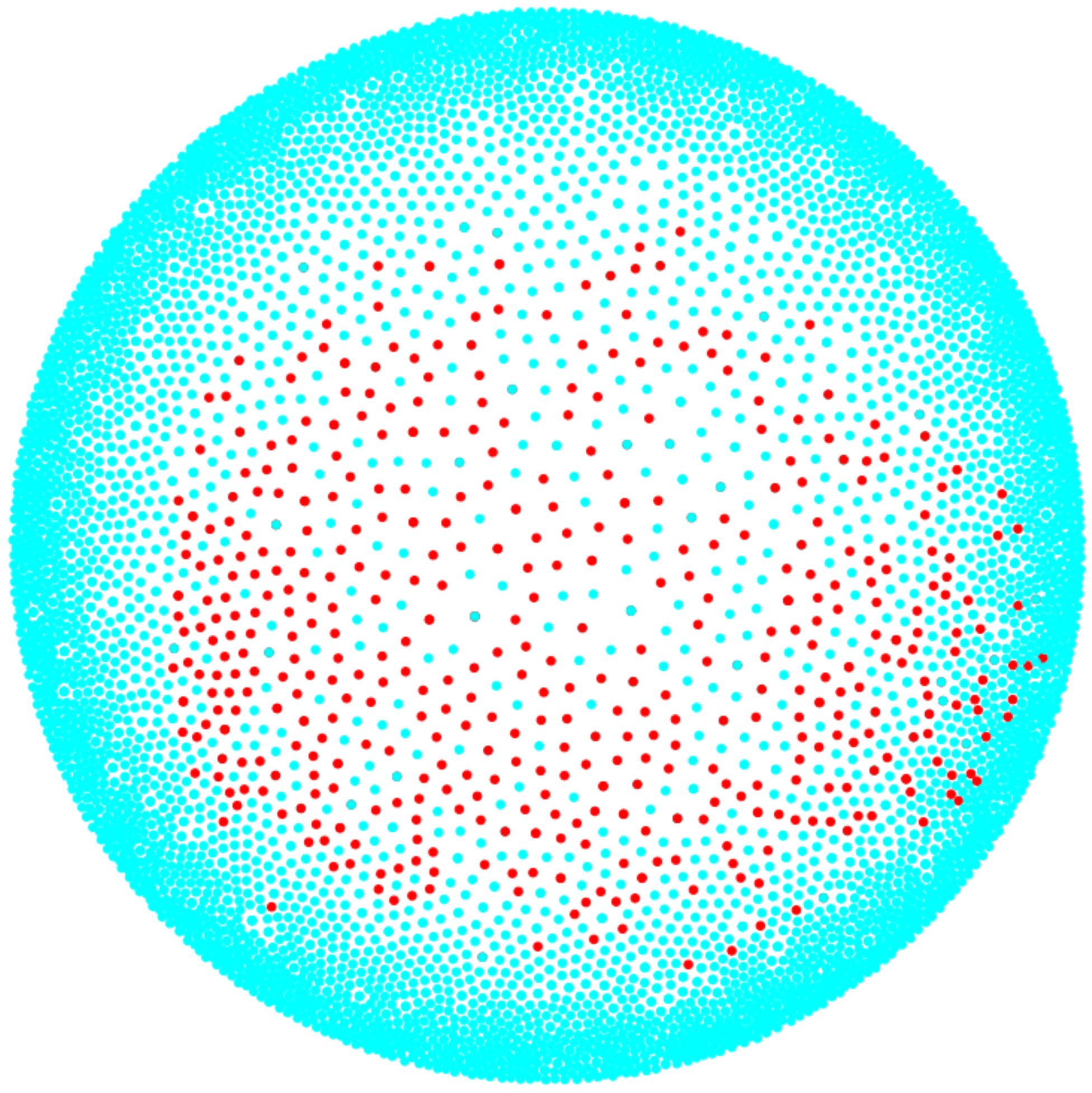}   \\ 
100 & 200  & 300 &  400 & 500
    \end{tabular}
    \caption{\textbf{HACK consistently places congealed images in the center of the Poincar\'e ball}. We gradually increase the number of original images replaced by congealed images from 100 to 500. The congealed images are marked with \textcolor{red}{red} dots and the original images are marked with \textcolor{cyan}{cyan} dots. }
    \label{fig:more_congealing}
\end{figure*}

We gradually increase the number of original images replaced by congealed images from 100 to 500. Still, as shown in Figure \ref{fig:more_congealing}, HACK can learn a representation that captures the concept of prototypicality regardless of the number of congealed images. This again confirms the effectiveness of HACK for discovering  prototypicality.

\section{Different Random Seeds}
\begin{figure*}[!ht]
    \centering
\setlength{\tabcolsep}{0pt}
    \begin{tabular}{ccccc}
\includegraphics[width=0.2\textwidth]{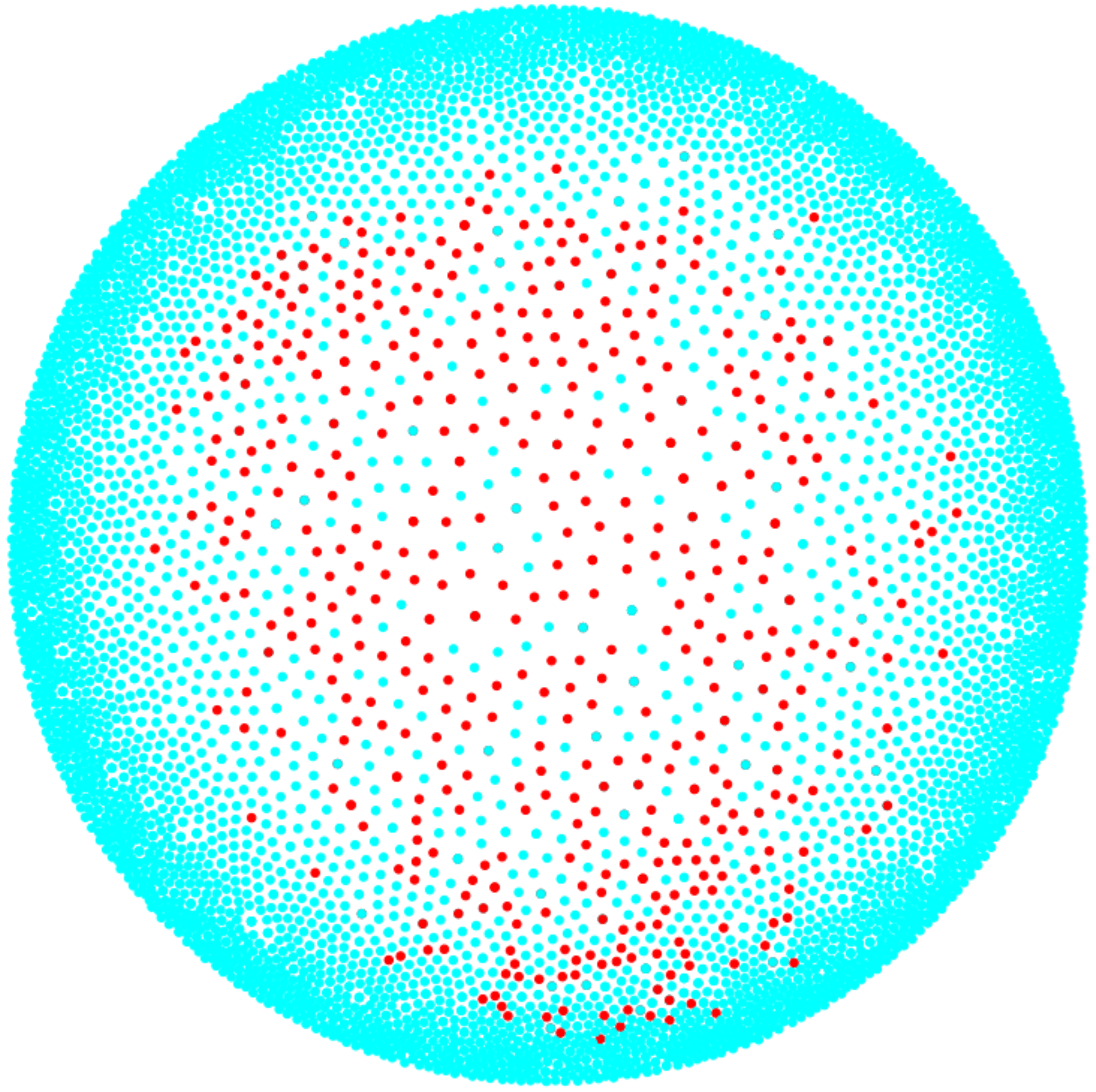} & \includegraphics[width=0.2\textwidth]{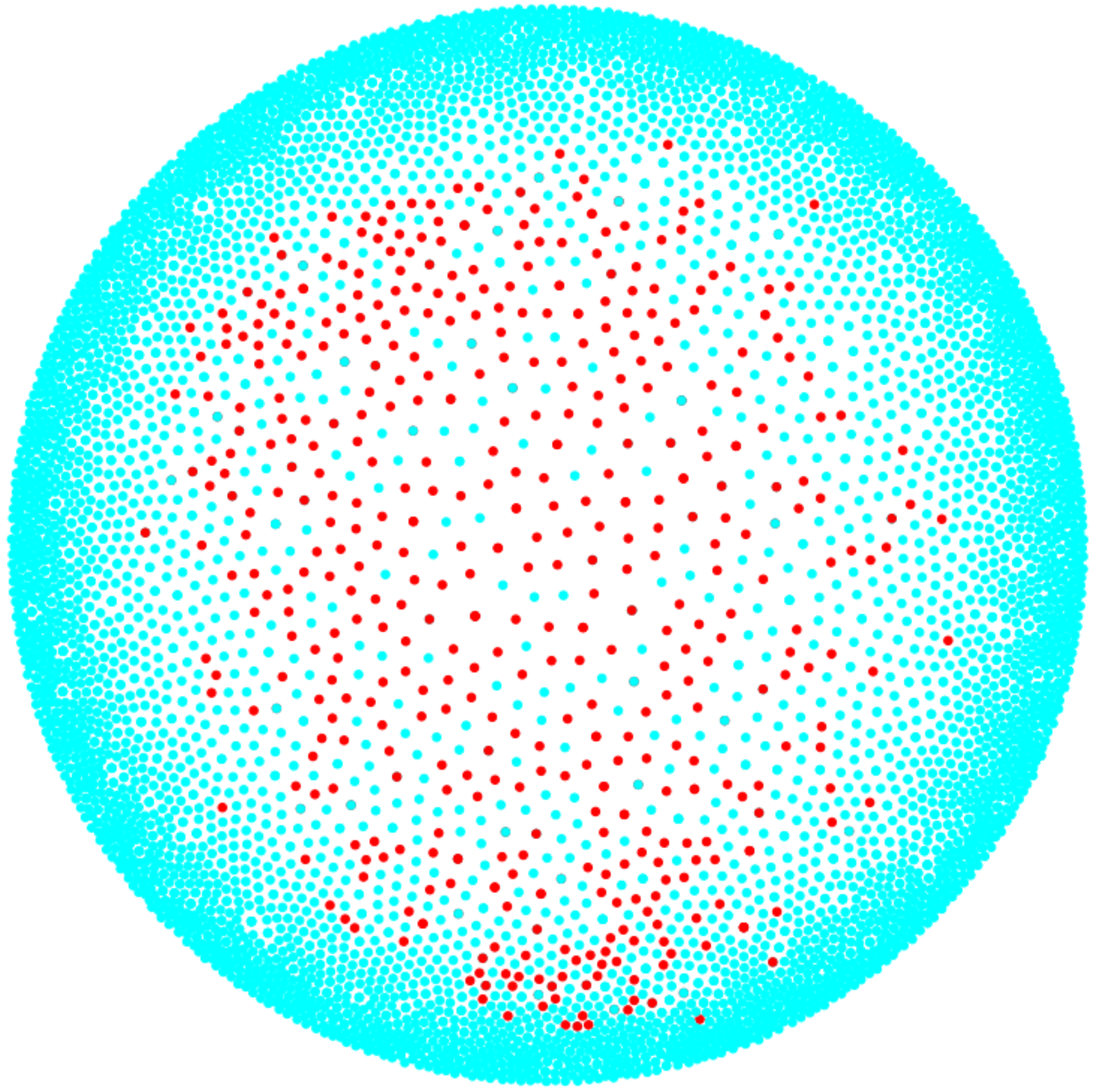} & \includegraphics[width=0.2\textwidth]{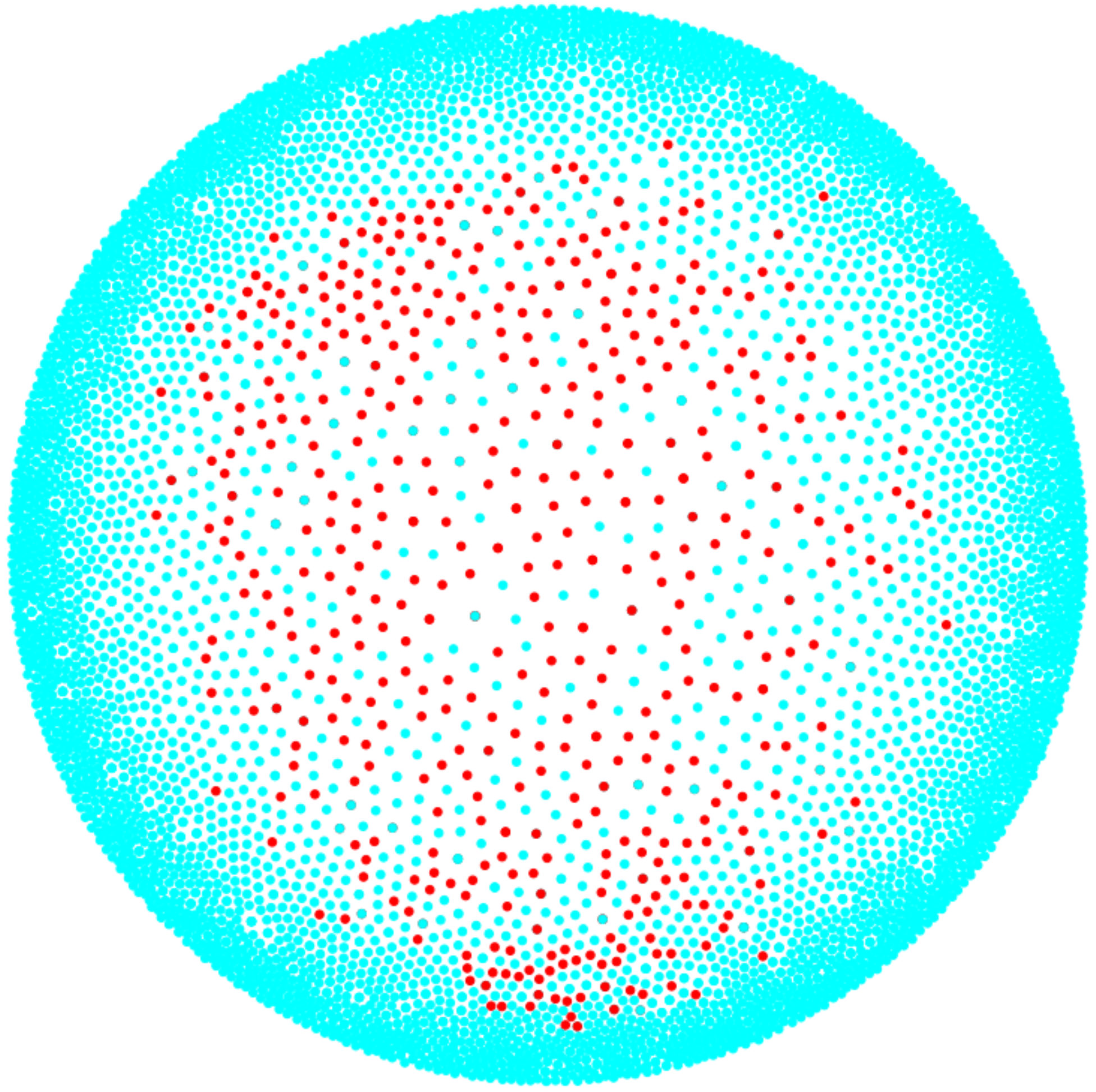}  & 
\includegraphics[width=0.2\textwidth]{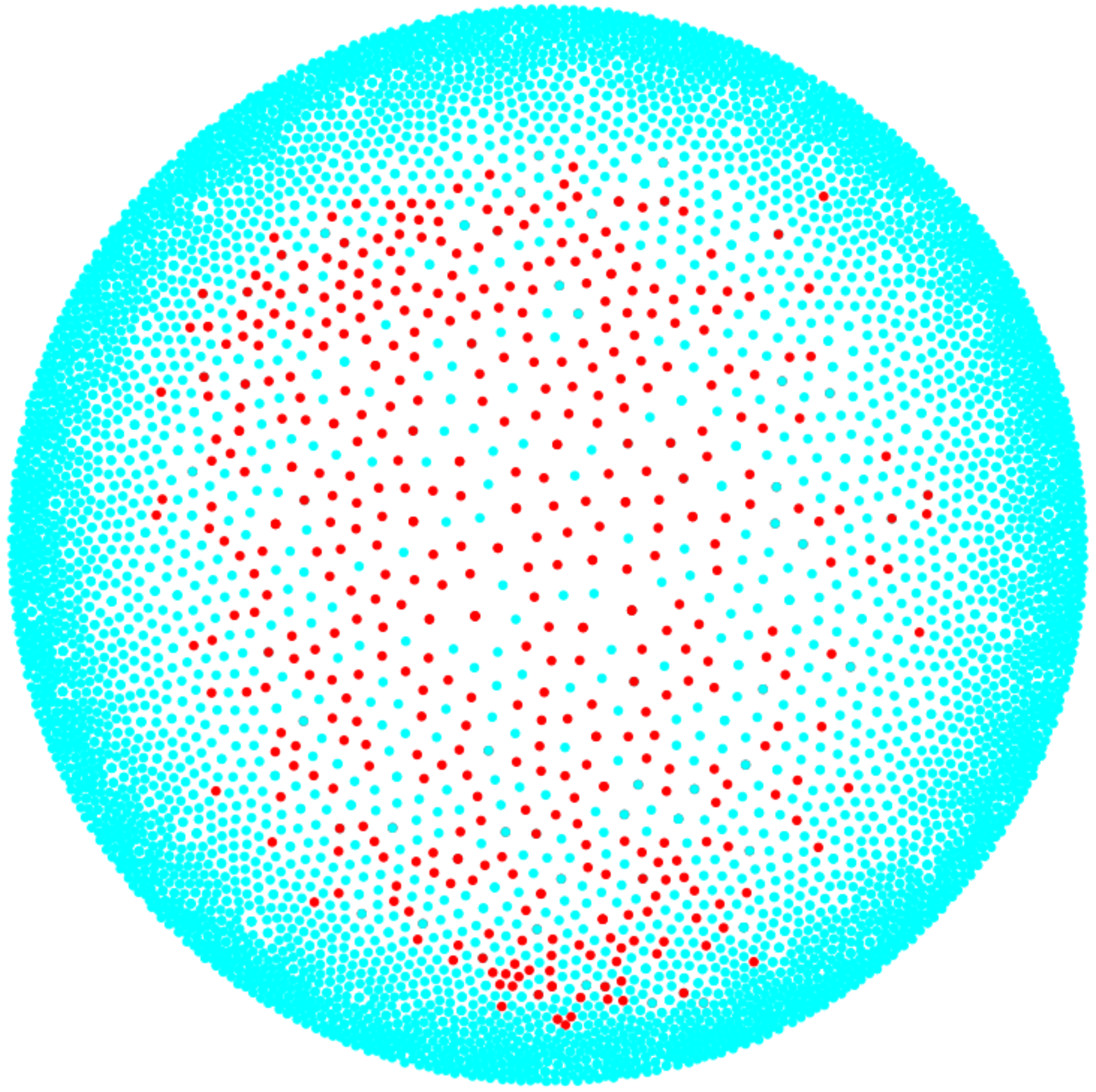}  &
\includegraphics[width=0.2\textwidth]{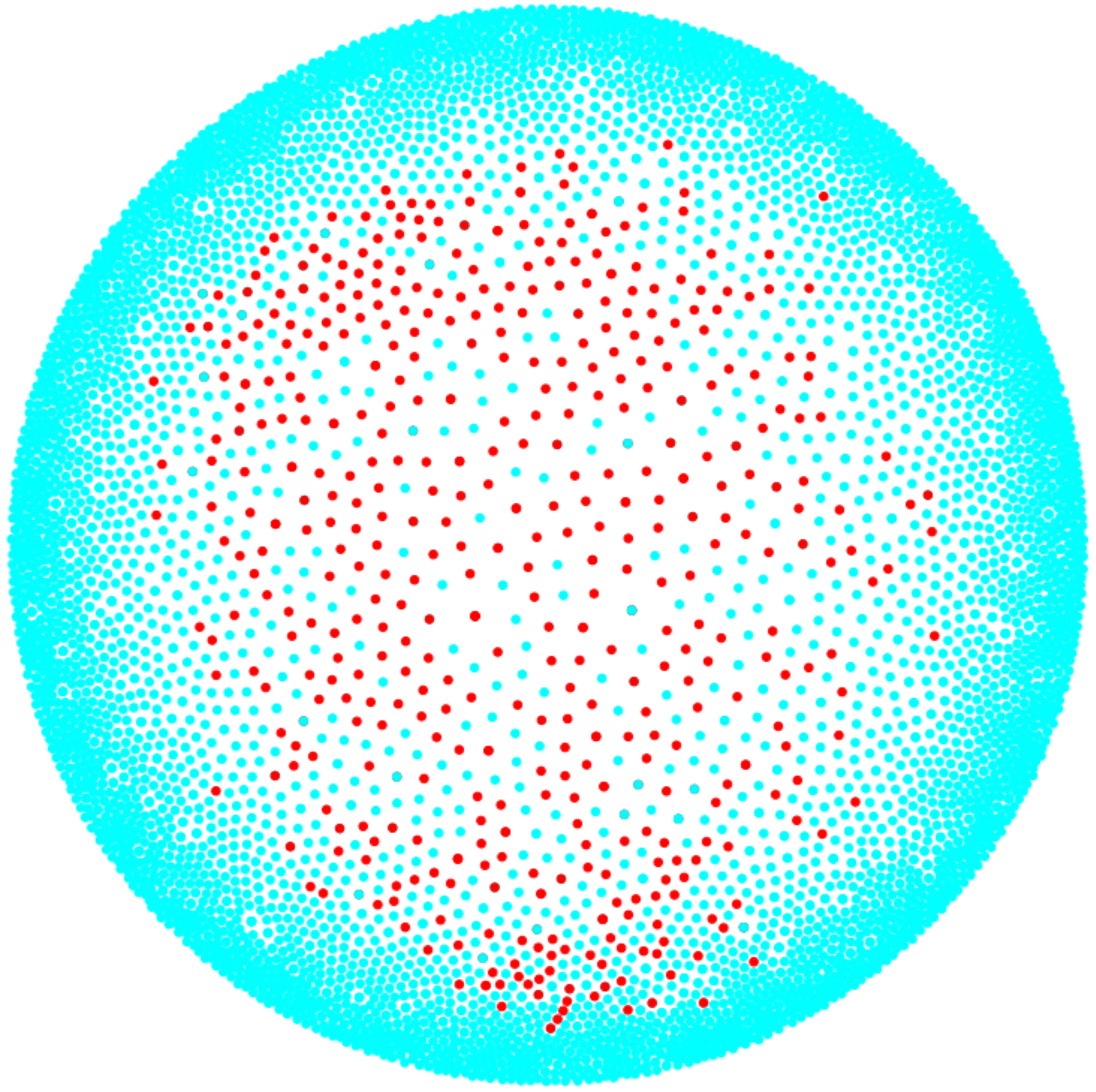}   \\ 
Seed 1 & Seed 2  & Seed 3 &  Seed 4 & Seed 5
    \end{tabular}
    \caption{\textbf{HACK consistently places congealed images in the center of the Poincar\'e ball in multiple runs with different random seeds.}. The congealed images are marked with \textcolor{red}{red} dots and the original images are marked with \textcolor{cyan}{cyan} dots. }
    \label{fig:more_seeds}
\end{figure*}

We further run the assignment 5 times with different random seeds. The results are shown in Figure \ref{fig:more_seeds}. We observe that the algorithm does not suffer from high variance and the congealed images are always assigned to the particles in the center of the Poincar\'e ball. This further confirms the efficacy of the proposed method for discovering prototypicality. 

\section{Emergence of Prototypicality in the Feature Space}

Existing unsupervised learning methods mainly focus on learning features for differentiating different classes or samples \cite{wu2018unsupervised,he2020momentum,chen2020simple}. The learned representations are transferred to various downstream tasks such as segmentation and detection. In contrast, the features learned by HACK aim at capturing  prototypicality within a single class. 

To investigate the effectiveness of HACK in revealing prototypicality, we can include or exclude congealed images in the training process. When the congealed images are included in the training process, we expect the congealed images to be located in the center of the Poincar\'e ball while the original images to be located near the boundary of the Poincar\'e ball. When the congealed images are excluded from the training process, we expect the features of congealed images produced via the trained network to be located in the center of the Poincar\'e ball.

\subsection{Training with congealed images and original images}
\label{sec:with_c_and_o}

\begin{figure*}[!ht]
    \centering
\setlength{\tabcolsep}{0pt}
    \begin{tabular}{ccccc}
\includegraphics[width=0.2\textwidth]{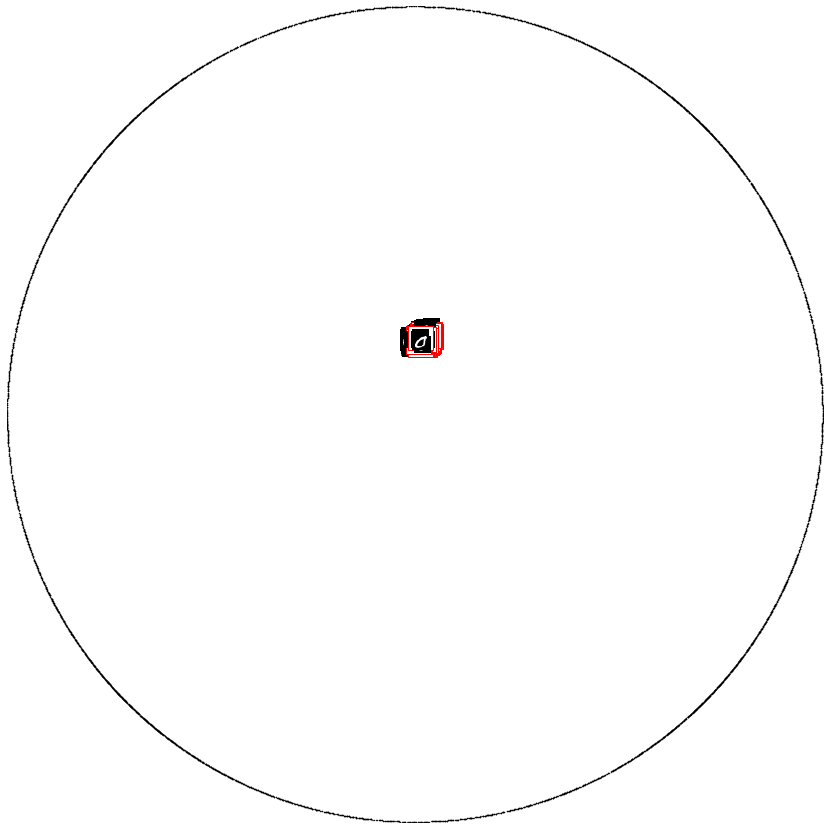} & \includegraphics[width=0.2\textwidth]{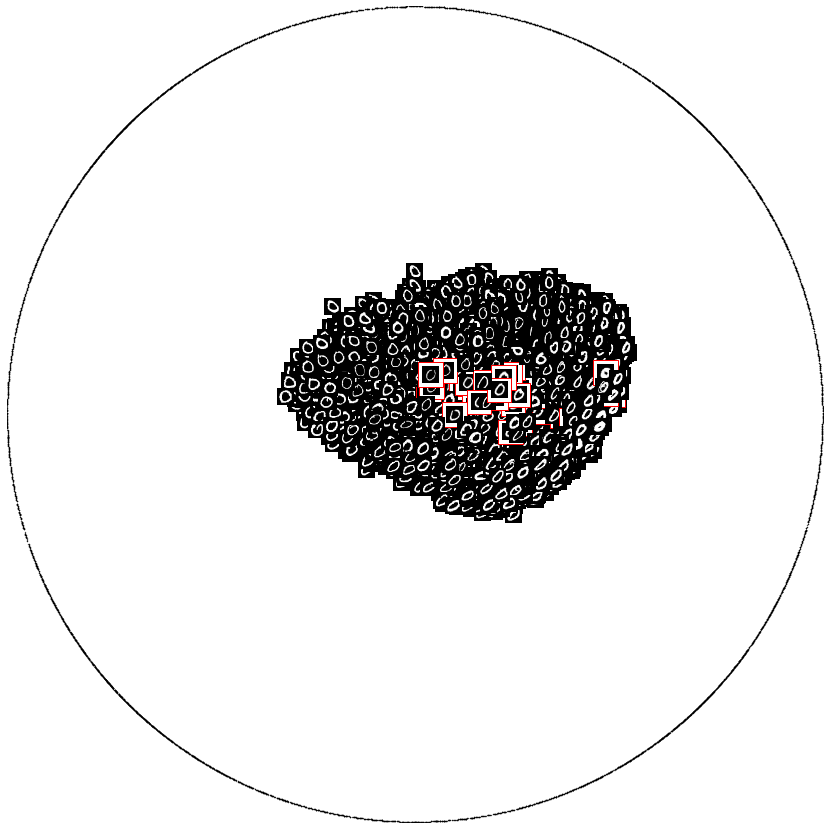} & \includegraphics[width=0.2\textwidth]{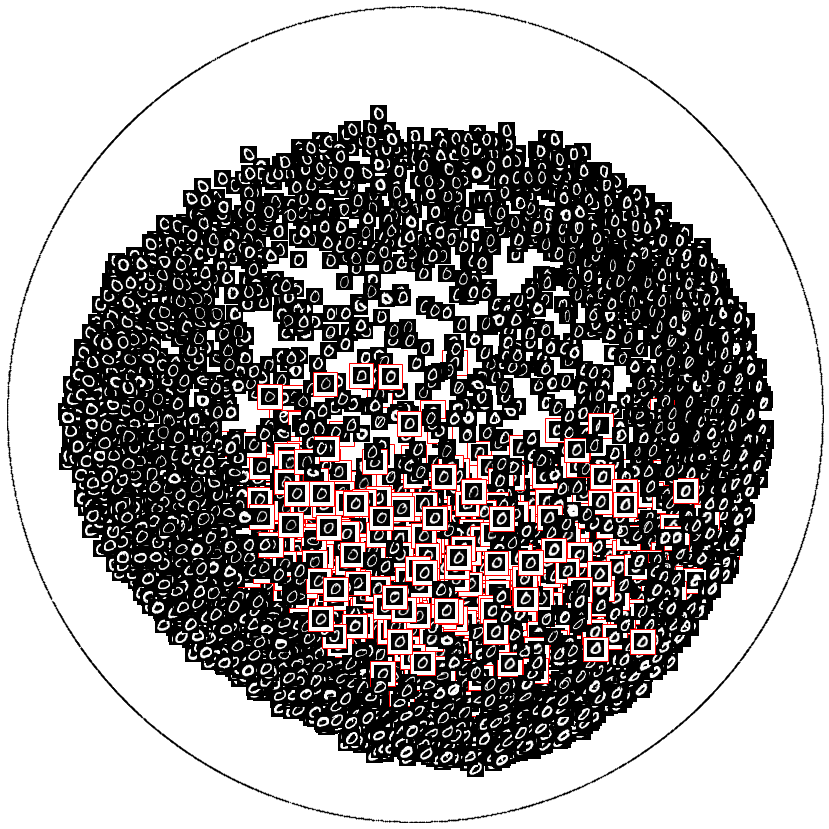}  & 
\includegraphics[width=0.2\textwidth]{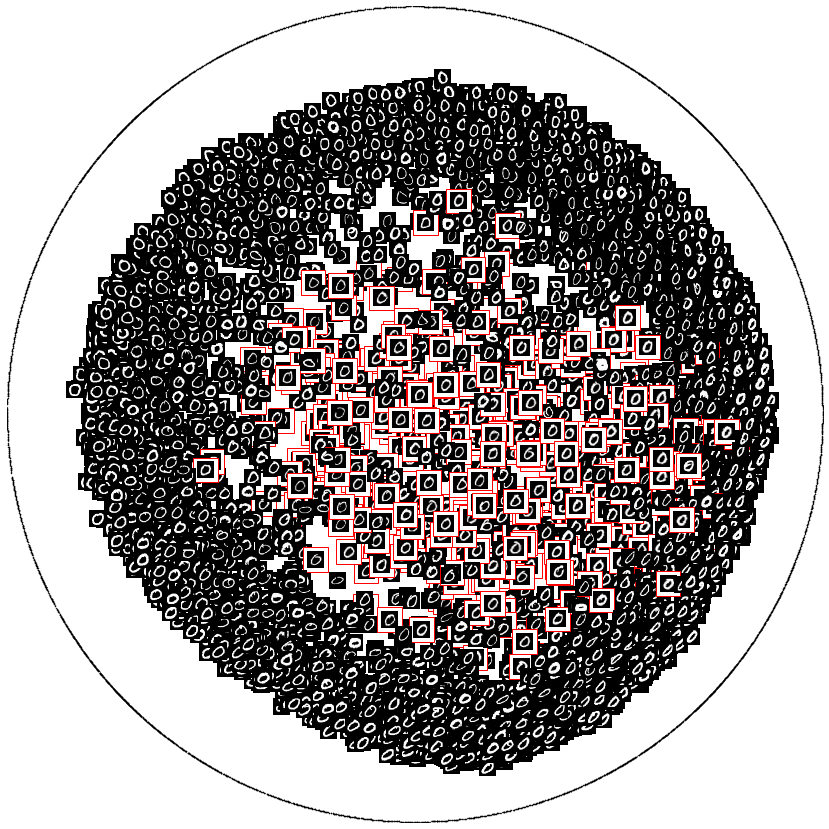}  &
\includegraphics[width=0.2\textwidth]{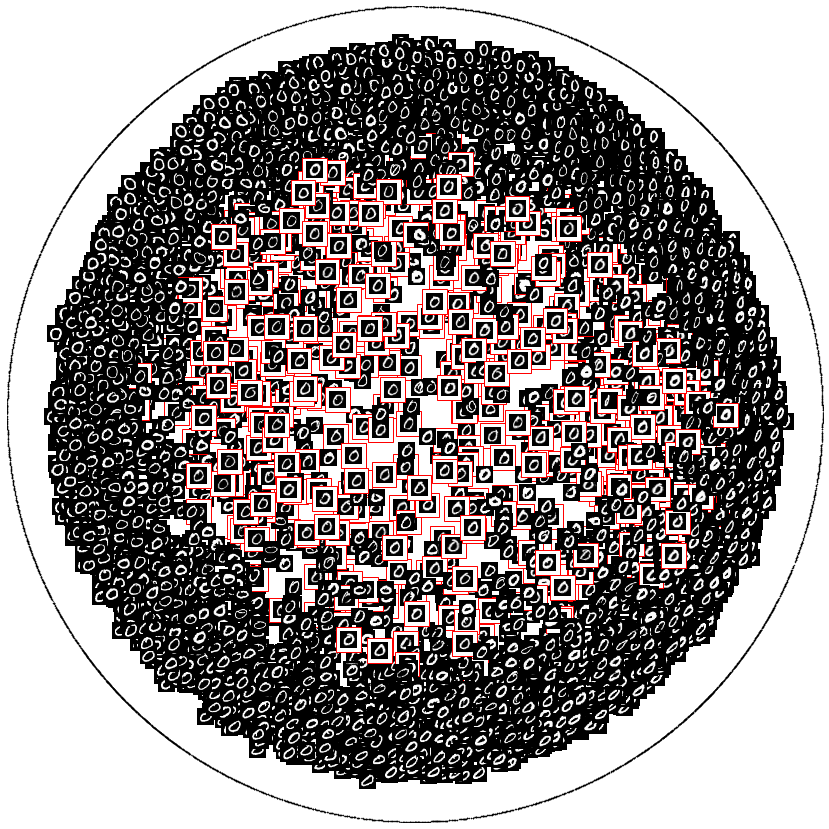}   \\ 
Epoch 1 & Epoch 5  & Epoch 10 &  Epoch 15 & Epoch 200
    \end{tabular}
    \caption{\textbf{Atypical images gradually move to the boundary of the Poincar\'e ball}. This shows that the representations learned by HACK capture prototypicality. Congealed images are in \textcolor{red}{red} boxes which are more typical. The network is trained with \emph{both} the congealed images and original images.}
    \label{fig:with_congealing_and_original}
\end{figure*}

We follow the same setups as in Section 4.3.1 of the main text. Figure \ref{fig:with_congealing_and_original} shows the hyperbolic features of the congealed images and original images in different training epochs. The features of the congealed images stay in the center of the Poincar\'e ball while the features of the original images gradually expand to the boundary.

\subsection{Training only with original images}

\begin{figure*}[!ht]
    \centering
\setlength{\tabcolsep}{0pt}
    \begin{tabular}{ccccc}
\includegraphics[width=0.2\textwidth]{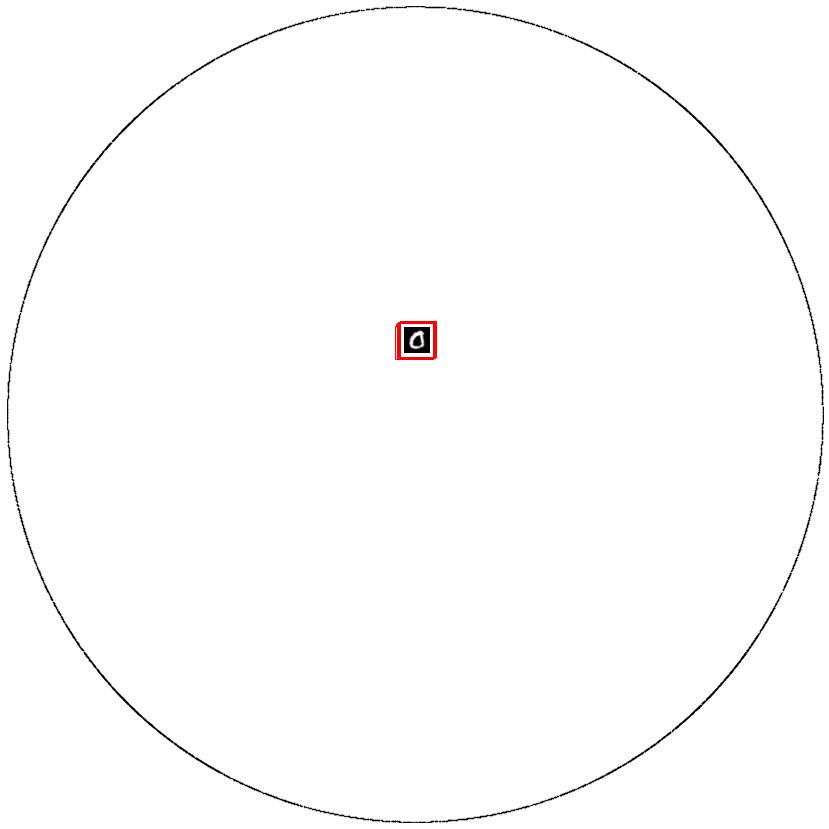} & \includegraphics[width=0.2\textwidth]{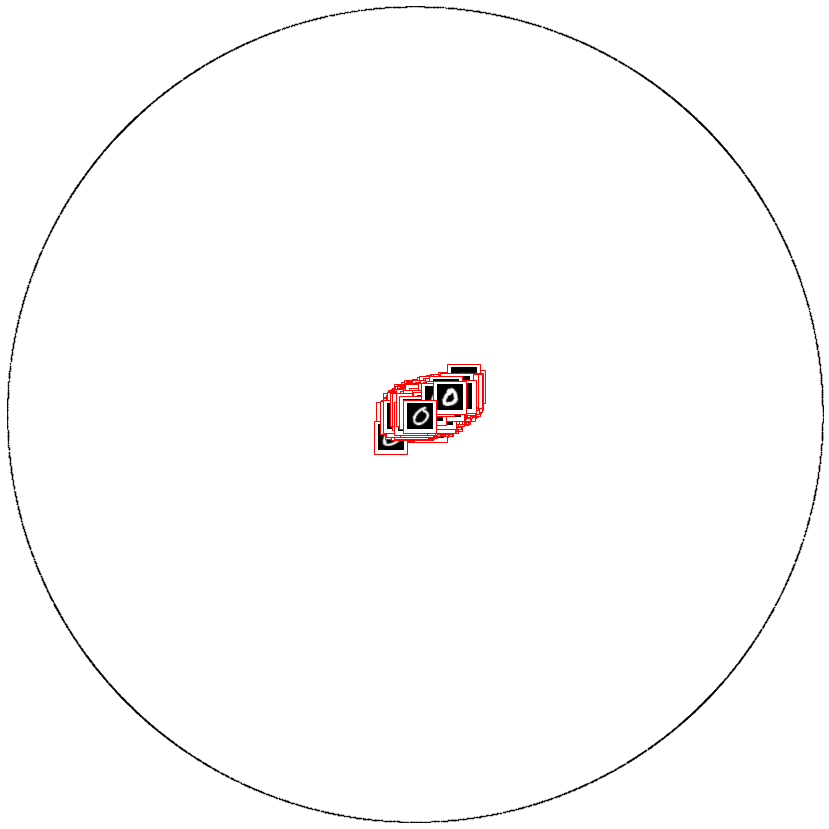} & \includegraphics[width=0.2\textwidth]{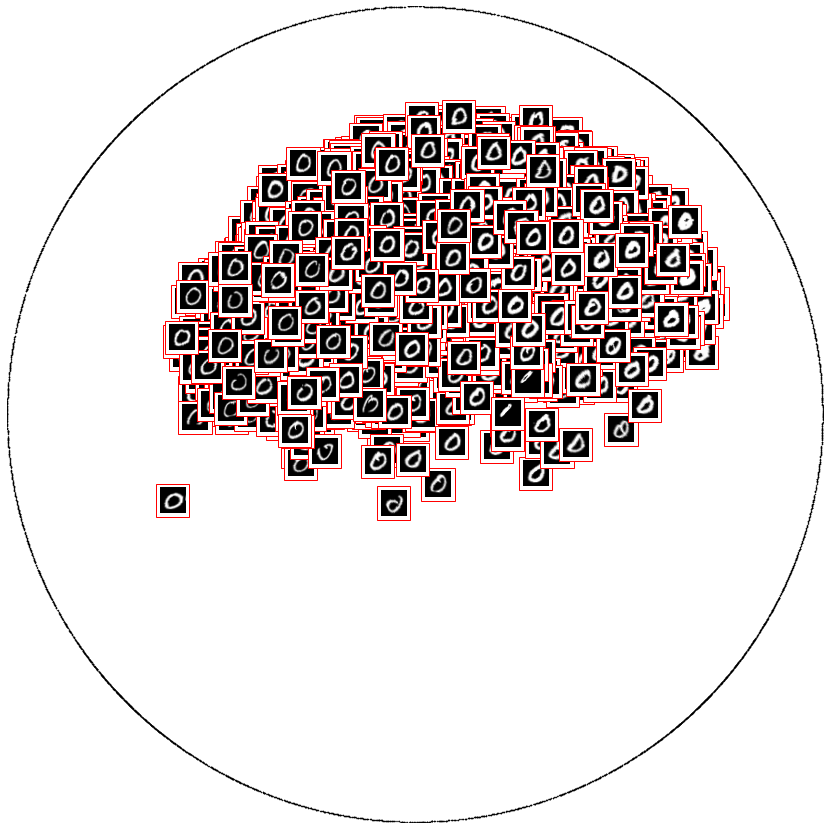}  & 
\includegraphics[width=0.2\textwidth]{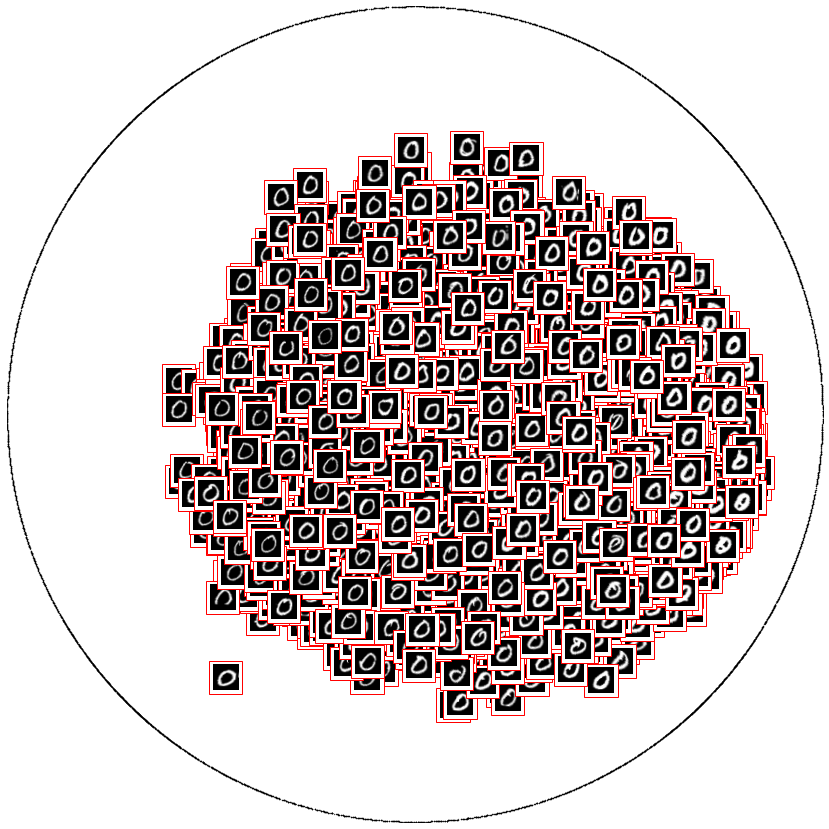}  &
\includegraphics[width=0.2\textwidth]{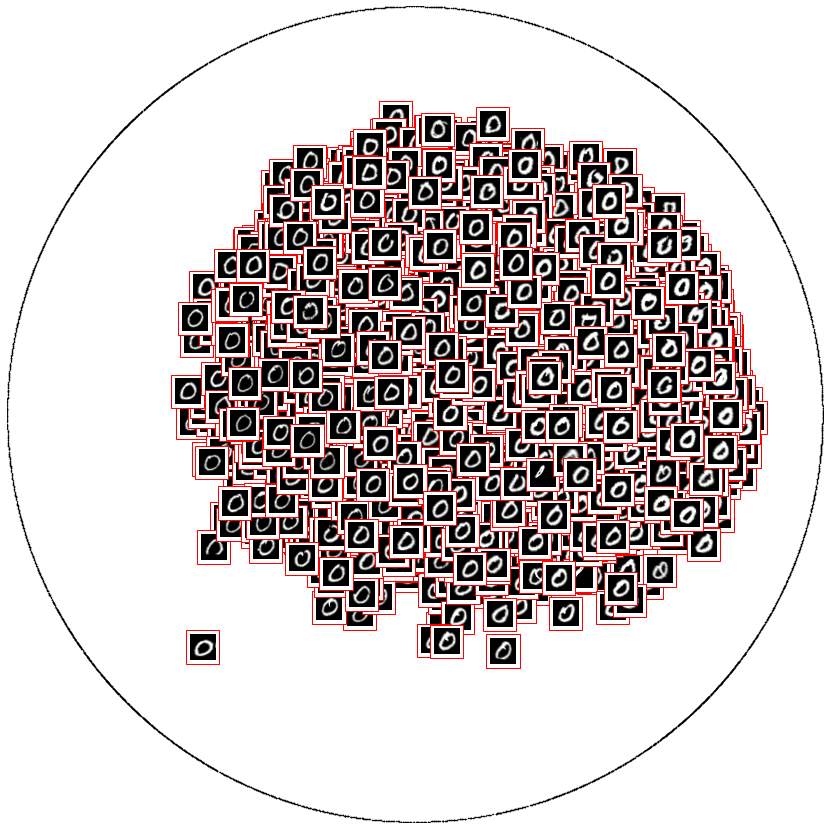}   \\ 
Epoch 1 & Epoch 10  & Epoch 20 &  Epoch 40 & Epoch 200
    \end{tabular}
    \caption{\textbf{The representations learned by HACK gradually capture prototypicality during the training process.} Congealed images are in \textcolor{red}{red} boxes which are more typical. We produce the features of the congealed images with the trained network in different epochs. The network is \emph{only} trained with original images.}
    \label{fig:only_with_congealing}
\end{figure*}

Figure \ref{fig:only_with_congealing} shows the hyperbolic features of the congealed images \textbf{when the model is trained only with original images}. As we have shown before, congealed images are naturally more typical than their corresponding original images since they are aligned with the average image. The features of congealed images are all located close to the center of the Poincar\'e ball. This demonstrates that prototypicality naturally emerges in the feature space.

Without using congealed images during training, we exclude any artifacts and further confirm the effectiveness of HACK for discovering prototypicality. We also observe that the features produced by HACK also capture the fine-grained similarities among the congealing images despite the fact that all the images are aligned with the average image.

\newpage

\newpage

\section{Discussions on Societal Impact and Limitations.}

We address the problem of unsupervised learning in hyperbolic space. We believe the proposed HACK should not raise any ethical considerations. We discuss current limitations below,

\textbf{Applying to the Whole Dataset} Currently, HACK is applied to each class separately. Thus, it would be interesting to apply HACK to all the classes at once without supervision. This is much more challenging since we need to differentiate between examples from different classes as well as the prototypical and semantic structure.

\textbf{Exploring other Geometrical Structures} We consider uniform packing in hyperbolic space to organize the images. It is also possible to extend HACK by specifying other geometrical structures to encourage the corresponding organization to emerge from the dataset.

\end{document}